\def\OSHelixOpenVersion{1}
\documentclass{article}
\usepackage{iclr2027_conference,times}

\usepackage{amsmath,amsfonts,bm}

\def\eqref#1{equation~\ref{#1}}
\def\1{\bm{1}}

\DeclareMathAlphabet{\mathsfit}{\encodingdefault}{\sfdefault}{m}{sl}
\SetMathAlphabet{\mathsfit}{bold}{\encodingdefault}{\sfdefault}{bx}{n}

\usepackage{url}
\usepackage{graphicx}
\usepackage{float}
\usepackage{wrapfig}
\usepackage{cuted}
\usepackage{caption}
\usepackage{booktabs}
\usepackage{tabularx}
\usepackage{multirow}
\usepackage[table]{xcolor}
\usepackage{placeins}
\usepackage[most]{tcolorbox}
\definecolor{linkblue}{RGB}{0,0,128}
\usepackage[colorlinks=true, allcolors=linkblue]{hyperref}
\usepackage{fontawesome5}

\newif\ifoshelixopenversion
\oshelixopenversionfalse
\ifdefined\OSHelixOpenVersion
  \oshelixopenversiontrue
\fi

\definecolor{lightgray}{gray}{0.92}
\definecolor{darkgray}{gray}{0.35}
\definecolor{gainGreen}{RGB}{0,125,75}
\definecolor{lossRed}{RGB}{190,45,45}

\newcommand{\gaincell}[2]{\mbox{#1{\color{gainGreen}\fontsize{5}{5.5}\selectfont(\ensuremath{\uparrow}#2)}}}
\newcommand{\losscell}[2]{\mbox{#1{\color{lossRed}\fontsize{5}{5.5}\selectfont(\ensuremath{\downarrow}#2)}}}

\title{\ours{}: Unifying GUI Interaction and Code Execution in Native Computer-Using Agents}

\author{
\begin{minipage}{0.95\textwidth}
\centering
\textbf{Bowen Yang}\textsuperscript{1,2,*} \enspace
\textbf{Kaiming Jin}\textsuperscript{3,*} \enspace
\textbf{Zichen Ding}\textsuperscript{2,*} \enspace
\textbf{Zhenyu Wu}\textsuperscript{4} \\
\textbf{Zehao Li}\textsuperscript{1} \enspace
\textbf{Zhoumianze Liu}\textsuperscript{5} \enspace
\textbf{Kanzhi Cheng}\textsuperscript{6} \enspace
\textbf{Fangzhi Xu}\textsuperscript{7} \\
\textbf{Yanan Sun}\textsuperscript{2} \enspace
\textbf{Qiushi Sun}\textsuperscript{8,$\dagger$} \enspace
\textbf{Yu Qiao}\textsuperscript{2,$\dagger$} \\
{\normalfont
\textsuperscript{1}University of Science and Technology of China \enspace
\textsuperscript{2}Shanghai AI Laboratory \\
\textsuperscript{3}National University of Singapore \enspace
\textsuperscript{4}Shanghai Jiao Tong University \enspace
\textsuperscript{5}Fudan University \\
\textsuperscript{6}Nanjing University \enspace
\textsuperscript{7}Xi'an Jiaotong University \enspace
\textsuperscript{8}The University of Hong Kong \\
\textsuperscript{*}Equal contribution. \enspace
\textsuperscript{$\dagger$}Corresponding authors.
}
\end{minipage}
}

\makeatletter
\newcommand{\ours}[1]{OS-Helix\if\relax\detokenize{#1}\relax\else-#1\fi}
\makeatother

\ifoshelixopenversion
  \iclrfinalcopy
\fi
\begin{document}

\fancypagestyle{oshelixfirstpage}{%
  \fancyhead{}
  % \fancyfoot[C]{\thepage}%
  \renewcommand{\headrulewidth}{0.4pt}%
}

\maketitle

\ifoshelixopenversion
  \fancyhead{}
  \renewcommand{\headrulewidth}{0pt}
  \thispagestyle{oshelixfirstpage}
  \vspace{-1.2\baselineskip}
\fi

\begin{abstract}
% 真实桌面自动化既不是纯视觉问题，也不是纯程序化问题。智能体必须同时理解用户界面，并高效操纵文件、脚本和系统状态。现有计算机使用智能体大多专注于 GUI 控制；而简单给 GUI 模型添加代码工具反而可能破坏其行为，因为模型并未学会何时在两种动作模态之间切换。我们提出 \ours{}，一个端到端训练的多模态计算机使用智能体，将 GUI 动作与代码执行统一到同一动作空间中。为使这一范式可训练且可复现，我们提出应用--产物组合，通过将扩展的桌面应用集合与扩展的真实文件产物集合配对来构建可执行 Ubuntu 桌面状态；同时设计了解耦的指令/评估器合成流水线，生成扎根于环境、且不依赖求解模态的可验证任务。我们首先在过滤后的 GUI/代码混合轨迹上进行 SFT 冷启动，随后在真实桌面环境中进行在线强化学习。针对强化学习，我们提出 Step-balanced GSPO，只改变 token-level log-importance weight 到样本级 importance ratio 的聚合方式，按动作步骤而非原始 assistant token 聚合，避免长代码动作主导异构动作优化。在 OSWorld-Verified 上，\ours{27B} 与 \ours{35B-A3B} 相较对应 Qwen3.5 混合基线分别提升 35.0\% 和 33.7\%，而经过强化学习的 \ours{27B-RL} 达到 66.44\% 成功率。额外实验表明，性能提升来自学习到的 GUI/代码协同、可扩展的可验证数据以及平衡的异构动作优化，并在 WindowsAgentArena 上展示出跨操作系统泛化性。我们将在未来开源研究资料。
Real desktop automation is neither purely visual nor purely programmatic. Agents must understand user-facing interfaces while efficiently manipulating files, scripts, and system state.
% Existing computer-using agents mainly specialize in GUI control, and naively adding a code tool can even degrade performance because the model has not learned when to route between GUI interaction and code execution. 
However, native GUI/code coordination cannot be obtained by prompt-level tool composition. In our preliminary experiments, naively adding a code tool degrades a GUI agent, revealing that GUI/code routing must be internalized through training on high-quality hybrid data.
We present \ours{}, an end-to-end trained multimodal computer-using agent that unifies GUI actions and code execution in one action space. 
To construct the required supervision, we introduce application--artifact composition, pairing desktop applications with authentic file artifacts to instantiate executable Ubuntu states, and a decoupled instruction/evaluator generation pipeline that produces environment-grounded tasks whose final states can be verified regardless of the solution modality. 
We cold-start the policy with SFT on filtered hybrid data and further improve it through online RL in live desktop environments. 
For RL, we propose Step-balanced GSPO, which aggregates importance weight by action step rather than raw assistant tokens to prevent long code actions from dominating heterogeneous-action optimization. 
On OSWorld-Verified, \ours{27B} and \ours{35B-A3B} achieve relative improvements of 35.0\% and 33.7\% over their matched Qwen3.5 hybrid baselines, while \ours{27B-RL} further improves by 42.9\% to reach 66.44\% success. 
Ablations provide evidence consistent with contributions from learned GUI/code coordination, scalable verifiable supervision, and action step balanced optimization, while demonstrating cross-platform transfer on WindowsAgentArena. 
\ifoshelixopenversion
Our project homepage is at \faHome\,\href{https://os-copilot.github.io/OS-Helix/}{os-copilot.github.io/OS-Helix}.
\else
We will release all research artifacts.
\fi
\end{abstract}

\section{Introduction}
\label{sec:introduction}
\begin{wrapfigure}[15]{r}{0.43\textwidth}
  \vspace{-1.2\baselineskip}
  \centering
  \includegraphics[width=\linewidth]{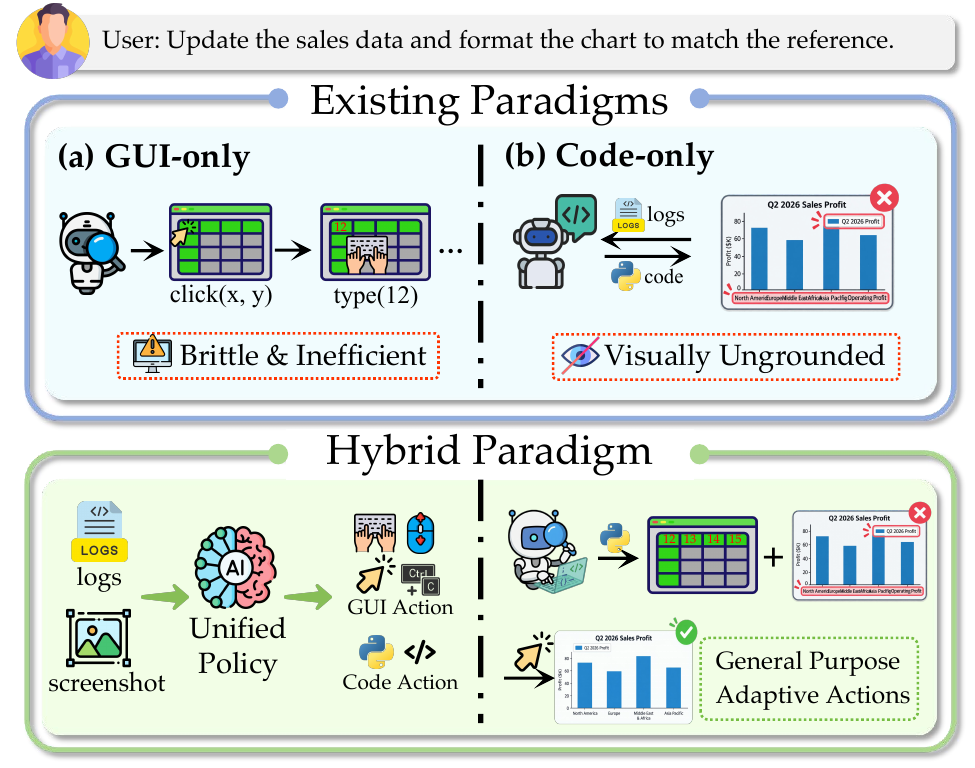}
  \vspace{-1.2\baselineskip}
  % Original caption: Overview of the GUI/code hybrid setting targeted by computer-using agents such as \ours{}. A real desktop agent should combine visual interaction with direct file-system and command-line control.
  \caption{Overview of GUI-only, code-only, and hybrid computer use.}
  \label{fig:teaser}
  \vspace{-0.8\baselineskip}
\end{wrapfigure}

% 背景1:CUA (从GUI Agent引入)
% 随着VLMs的快速发展，大语言模型正经历着从对话到执行任务的转变，而计算机使用是构建通用agent的重要场景。传统的CUAs通过阅读、理解屏幕来模拟人类行为，以此来实现digital automation。
% The rapid progress of vision-language models (VLMs)~\citep{team2026qwen3,wang2025internvl35,team2026kimi,anthropic2025claude4.5,zeng2026glm,singh2025openai,bai2026intern} is shifting foundation models from passive dialogue systems toward agents that can execute open-ended tasks in digital environments. Among these settings, computer use is a particularly important testbed for general-purpose agency: computer-using agents~(CUAs) can perceive the screen, reason over the current interface state, and operate applications in the same environments used by humans. Existing CUAs therefore primarily cast digital automation as a graphical interaction problem, where the agent reads screenshots, localizes interface elements, and emits low-level mouse and keyboard actions.
The rapid progress of vision-language models (VLMs)~\citep{team2026qwen3,wang2025internvl35,team2026kimi,anthropic2025claude4.5,zeng2026glm,openai2025gpt5systemcard,bai2026intern,zou2026intern} has accelerated the development of computer-using agents (CUAs), which aim to execute open-ended tasks in human-facing digital environments. Most existing CUAs formulate digital automation as graphical interaction, where agents perceive screenshots, reason over interface states, and act through mouse and keyboard operations.

However, GUI-centric CUAs~\citep{ScaleCUA,EvoCUA,wang2025opencuaopenfoundationscomputeruse,TongUI,luo2025gui,cheng2024seeclick,feizi2025groundcua} still face practical bottlenecks in realistic \mbox{desktop} workflows. 
Visual grounding can fail on small or ambiguous interface elements, and fine-grained interaction can make structured state manipulation cumbersome even with batched actions.
A complementary interface is programmatic control, through which command-line and code agents can inspect files, access system state, and execute compound operations directly through the shell~\citep{yang2026cli,yang2024swe,wang2025openhands}. 
Yet code execution alone cannot fully address the long tail of desktop task scenarios that depends on visual layout, application-specific UI state, or software behavior exposed only through the interface. As illustrated in Figure~\ref{fig:teaser}, these complementary strengths motivate a single agent that coordinates GUI and code, where GUI provides general visual access to arbitrary applications while code provides efficient and precise manipulation of programmatic state.

Despite this complementarity, building a native hybrid GUI/code CUA remains an open problem. Simply attaching a code tool to a GUI model does not teach it when or why to use either modality. In our preliminary experiments, as shown in Table~\ref{tab:training_model_analysis}, Qwen3-VL-32B-Instruct~\citep{bai2025qwen3} achieves over $\sim$34\% success on OSWorld-Verified~\citep{osworld_verified} with GUI actions alone, but falls to roughly $\sim$32\% when code execution is naively exposed. 
Prompt-level composition can therefore disrupt an existing GUI policy rather than create coordinated behavior, making it essential to internalize hybrid coordination capabilities into the model through training on high-quality hybrid data.
Several existing approaches have explored hybrid execution paradigms that combine GUI interaction with code-like actions. Framework-centric systems~\citep{gonzalez2025unreasonable,CoAct,OS-Symphony} rely on external harnesses to coordinate GUI, code, memory, and recovery; tool-call-based approaches~\citep{jia2025osworldmcp,yang2025ultracua,shi2026cua} expose programmatic operations through predefined application-specific interfaces; and model-fusion approaches~\citep{wang2025uiTARS2,qin2025uitars} combine complementary capabilities through predefined designs. 
Although these systems expand the scope of hybrid computer use, they leave open how a single policy can learn the routing between the GUI and code actions through hybrid training.
To address these challenges, we present \ours{}, an end-to-end trained computer-using agent that unifies GUI interaction and code execution within a single policy.

% 在基础设施层面，我们提出用于可扩展桌面环境的应用--产物组合方案。我们将可执行桌面状态的构建分解为具有版本和初始化配置的应用，以及真实文件产物；随后依据兼容关系在虚拟机中将二者组合起来。文件库覆盖文档、演示文稿、电子表格、图像和代码项目等二十余种常见产物类型，并与 45 个应用关联。由此得到的应用--产物组合能够实例化可复现的初始桌面状态，以支持真实任务合成。我们进一步构建了一条大规模合成与过滤流水线，由探索智能体提出扎根于当前环境、多样、无歧义的指令，再由独立的评估器智能体为通过筛选的指令构建任务特定且以规则为主的评估器，从而支持可靠的轨迹采集和在线强化学习。
% On the infrastructure side, we propose application--artifact composition for scalable desktop environments. We factor executable desktop-state construction into versioned applications with initialization configurations and authentic artifacts, then compose them inside virtual machines according to compatibility. The resulting library covers more than twenty artifact types and 45 applications, supporting reproducible initial desktop states for realistic task synthesis. We further build a synthesis and filtering pipeline in which an \textbf{Exploration Agent} proposes grounded instructions and an independent \textbf{Evaluator Agent} constructs task-specific, primarily rule-based evaluators, enabling reliable trajectory collection and online reinforcement learning.
% 在基础设施层面，我们引入了应用——制品组合方法，将版本化的应用与真实文件制品相结合，以实例化多样化、可复现的桌面状态。\textbf{探索智能体}基于这些状态提出任务指令，而独立的\textbf{评估智能体}则构建特定任务的、以规则为主的评估器。整体而言，该流水线支持对高质量GUI/代码轨迹进行可扩展的合成、筛选与采集，从而为将混合协同能力内化至模型中提供了必要的数据支撑。
On the infrastructure side, we introduce application--artifact composition, which combines versioned applications with authentic artifacts to instantiate diverse and reproducible desktop states. An \textbf{Exploration Agent} proposes instructions grounded in these states through brief exploration of the initialized environment, while an independent \textbf{Evaluator Agent} constructs task-specific, primarily rule-based evaluators. Together, this pipeline supports scalable synthesis, filtering, and collection of high-quality GUI/code trajectories for model training, thereby providing the data foundation necessary to internalize hybrid coordination capabilities into the model.

On the training side, we first use filtered hybrid trajectories to internalize basic action semantics and modality routing, and then optimize the policy through online RL in executable desktop environments. 
In early RL experiments, we observe a steady increase in the proportion of code actions accompanied by a degradation in performance. We attribute this to the token-count asymmetry between short GUI actions and substantially longer code actions, which causes the optimization signal to be dominated by code steps. 
To address this, we propose \textbf{Step-balanced GSPO}, which constructs the sample-level importance ratio over action steps rather than raw assistant-token counts, enabling more balanced optimization of the two modalities.
% 这些训练组件直接回答了单一策略如何通过混合训练原生习得GUI与代码动作协同能力这一开放问题，由此得到的智能体在多样化桌面任务中表现优异且稳定。
These training components directly answer the open question of how a single policy can natively learn to coordinate GUI and code actions through hybrid training, yielding an agent that performs strongly and reliably across diverse desktop tasks.
% 评测结果
% 实验结果：在一致复现harness下，27B和35B-A3B模型都显著超过对应Qwen3.5混合动作baseline，27B经过Step-balanced GSPO进一步提升。
% We conduct extensive experiments on OSWorld-Verified. The results validate the full \ours{} pipeline, from application--artifact environment construction to hybrid-policy training. In particular, \ours{27B} reaches 62.75\% success, \ours{35B-A3B} reaches 57.27\%, and \ours{27B-RL} further improves to 66.44\%, yielding a large $\sim$20\% absolute improvement over matched Qwen3.5-27B GUI+code baselines. We also provide detailed ablations to validate the effectiveness of the proposed method.
% 贡献，列点
% 总结而言，OS-Helix 的贡献如下：
% 1. 范式/思想：第一篇细致探讨 GUI/Code 混合范式(面向数字世界真实可用)的端到端模型训练的工作
% 2. Model：基于 Qwen3.5 系列，我们提供了 OS-Helix-27B 和 OS-Helix-35B-A3B 这两种中等型号的模型以供开源社区研究
% 3. Infra：我们提供并开源了完整的数据合成虚拟机环境，完善的RL管理环境，完备的任务指令合成与任务轨迹收集管线
% 4. Data：我们提供并开源了约9K强模型生成的高质量指令、约84K SFT 样本，以及约31K 精心过滤后处理后的成功SFT样本（轨迹与SFT样本动作空间均GUI/Code混合），以供社区使用与研究

In summary, our contributions are:

1) We formulate native GUI/code hybrid computer use as an end-to-end policy-learning problem and show through controlled experiments that merely exposing a GUI policy to code execution fails to yield reliable routing, motivating the need to internalize coordination through hybrid training.

2) We introduce application--artifact composition and a decoupled instruction/evaluator synthesis pipeline to construct diverse, reproducible, executable desktop states and generate high-quality hybrid data for internalizing GUI and code execution capabilities.

3) We propose Step-balanced GSPO for the agentic RL stage of native hybrid policy training, which prevents long code actions from dominating optimization, and enables a single policy to learn GUI and code coordination.

4) We will release the \ours{} model family and research artifacts, including the desktop data-synthesis and RL environments, task and trajectory collection pipelines, approximately 9K generated instructions, 83K unfiltered SFT samples, and about 31K successful filtered SFT samples in the unified GUI/code action space.

\ifoshelixopenversion
  \section{Related Work}
\label{sec:related_work}

\textbf{Hybrid GUI/Code Computer-Using Agents.}
Traditional GUI-native agents learn screenshot-to-action policies over visual interface states~\citep{qin2025uitars,wang2025opencuaopenfoundationscomputeruse,wu2025atlas,ScaleCUA,wang2025mmbenchgui}, whereas command-line and coding agents operate over files, shells, and programmatic state~\citep{ClaudeCode,OpenClaw,yang2026swe,yang2024swe}. Their complementary strengths motivate hybrid computer use.
Framework-centric systems~\citep{gonzalez2025unreasonable,CoAct,OS-Symphony} coordinate GUI, code, memory, and recovery through external harnesses, often using strong proprietary models. Their performance is coupled to costly inference and hand-engineered orchestration, making the model's own hybrid coordination difficult to isolate. Tool-call-based approaches~\citep{jia2025osworldmcp,yang2025ultracua,shi2026cua} expose programmatic operations through predefined, application-specific interfaces, limiting coverage and transfer to available integrations and tool schemas. Some approaches integrate GUI, code, and other capabilities through model fusion~\citep{wang2025uiTARS2,qin2025uitars}, relying on predefined fusion rather than coordination learned within a single policy.
Qwen-UI-Agent~\citep{zhou2026qwenuiagent} also explores training with mixed GUI and code-like action capabilities, demonstrating the practical potential of native hybrid agents. 
However, existing studies describe data construction, executable environments, supervision, and training configurations at different levels of granularity, making controlled reproduction and comparison of learned coordination mechanisms difficult. 
Our work complements by studying executable application--artifact state construction, decoupled instruction--evaluator synthesis, and more stable optimization for training a single GUI/code policy.

\textbf{Environment Scaling and Task Synthesis for Computer-Using Agents.}
% 真实、长视野的任务场景与交互环境对于CUA的训练至关重要，但人力很贵并且不具有可扩展性，所以如何通过自动化的方式获取高质量的数据成为了一个难题。部分工作通过自动爬取网络教程和文档，并将其分解为任务指令。还有一些从待交互的环境出发，根据环境来推出任务。例如Explorer根据交互的网页首页提取任务intent，并不断迭代完善；而OS-Genesis和GUI-ReWalk则是通过与环境的探索式交互反向合成任务。这类早期工作的特点是合成任务来收集高质量轨迹进行sft，并没有涉及环境scaling的部分，导致泛化性有限。
Realistic long-horizon tasks and interactive environments are essential for training CUAs, yet manual task design and trajectory annotation are expensive and difficult to scale. This makes automated acquisition of high-quality instructions and demonstrations a central problem. One line of work crawls web tutorials and documentation and decomposes them into executable task instructions~\citep{AgentTrek,TongUI}. Another line starts from the target environment itself and infers tasks from what the agent can observe or manipulate: Explorer~\citep{Explorer} extracts and refines task intents from web entry pages, while OS-Genesis~\citep{OS-Genesis}, GUI-ReWalk~\citep{GUI-ReWalk}, TreeCUA~\citep{TreeCUA}, and AUTO PLAY~\citep{ramrakhya2026scaling} synthesize tasks through exploratory interaction with the environment. These methods primarily scale task and trajectory synthesis, but they do not substantially scale the underlying interactive worlds, which constrains the diversity of behaviors learned by the agent.
% 与 Web 和 Mobile 等相对受限的交互域不同，Ubuntu 和 Windows 等 PC 环境依托系统级代码执行、成熟的应用生态、持久文件系统和包管理器，为可扩展的 CUA 训练提供了天然基础，并支持以具体系统状态为依据的可验证任务。近期工作已经开始使用虚拟机扩展桌面环境并合成任务，但这些系统通常更侧重扩大预定义桌面配置中的任务数量，而桌面真实性还取决于应用、文件类型、真实文件内容和持久系统状态的联合分布。
% 如表~\ref{tab:task_synthesis_comparison} 所总结，现有工作仍缺少一种统一基础设施，能够在扩展可执行桌面状态空间的同时，基于这些状态合成扎根于环境且可程序验证的任务。
Unlike more constrained interaction domains such as web and mobile~\citep{GUI-Genesis,InfiniteWeb,Autowebworld,PhoneWorld,OpenMobile,yan2026openmobile2}, PC environments, such as Ubuntu, Windows, and macOS, provide a natural substrate for scalable CUA training through system-level code execution, mature application ecosystems, persistent file systems, and package managers, enabling verifiable tasks grounded in concrete system state~\citep{sun2026osreward,fu2026macagentbench}. Recent work has begun to use virtual machines to expand desktop environments and synthesize tasks~\citep{CUA-Gym,lv2026scalecua,huang2026evocua}, exploring different parts of desktop-state construction and task generation. Table~\ref{tab:task_synthesis_comparison} summarizes the reported coverage of existing task-synthesis pipelines. Existing efforts differ in the extent to which they scale applications, authentic artifacts, executable initial states, and verifiable terminal conditions. 
% Our pipeline combines these components in a unified setting, while cross-application synthesis remains a preliminary component.
Our pipeline integrates application--artifact composition, reproducible desktop-state construction, and environment-grounded instruction and evaluator synthesis into a unified framework for scalable, verifiable desktop-task generation.

\textbf{Training Paradigms for Computer-Using Agents.}
% CUA训练范式大体经历了从SFT到离线偏好优化，再到在线Agentic RL的演进。SFT负责学习动作格式、视觉grounding和基础任务执行；离线偏好优化通过DPO、IPO等方法利用偏好数据改进策略，但仍主要受限于静态轨迹；在线RL进一步分为低层视觉定位的Grounding RL，以及在真实可交互环境中优化完整任务轨迹的Agentic RL。Agentic RL相关工作也开始围绕CUA场景提出或采用不同算法，例如ComputerRL的Entropulse、CUA-Gym的GSPO、ScaleCUA的GRPO-style训练和EvoCUA-1.5的STEPO。
CUA training has broadly evolved from imitation-style supervised fine-tuning toward reinforcement learning in interactive environments. SFT-based systems learn action formats, visual grounding, and task execution from human demonstrations, synthetic trajectories, web tutorials, or teacher-model rollouts~\citep{cheng2024seeclick,qin2025uitars,wang2025opencuaopenfoundationscomputeruse,TongUI,OS-Genesis,ScaleCUA}. Between SFT and online interaction, offline preference-optimization methods such as DPO~\citep{gao2026solving} and offline GRPO~\citep{yang2026gui} improve policies from static preference data, but remain limited by the coverage of collected trajectories and cannot directly optimize recovery under environment feedback. Online RL methods for CUAs are further split into grounding RL and agentic RL. Grounding RL optimizes low-level GUI localization or action selection with specialized visual rewards~\citep{lu2025ui,yuan2025enhancing,zhou2025guig1,tang2025gui,feizi2025groundcua}. Agentic RL instead optimizes complete trajectories in executable environments with rule-based or verifiable rewards. Recent CUA work has begun to tailor the optimization algorithm to this setting: ComputerRL~\citep{lai2025computerrl} introduces Entropulse, which alternates online RL with SFT to mitigate entropy collapse; CUA-Gym~\citep{CUA-Gym} trains CUA checkpoints with GSPO~\citep{zheng2025gspo}; and ScaleCUA~\citep{lv2026scalecua} adopts GRPO-style RL~\citep{shao2024deepseekmath,yu2026dapo} together with frontier sampling and visual-context segmentation. EvoCUA-1.5~\citep{huang2026evocua} proposes Step-Level Policy Optimization for decomposed multi-turn samples.
% 这些面向 CUA 场景的训练方法表明，核心挑战并不只是套用通用的 LLM 强化学习目标，而是如何使在线优化适应长程多模态上下文、稀疏的轨迹级奖励、缓慢的环境反馈，以及异质化的动作粒度。
% These CUA-specific training recipes highlight that the central challenge is not merely applying a generic LLM RL objective, but adapting online optimization to long multimodal contexts, sparse trajectory-level rewards, slow environment feedback, and heterogeneous action granularities.
% 然而，现有 CUA 训练研究主要优化以 GUI 为中心的动作空间，或通过外部系统编排加入代码工具；如何在单一策略中系统训练原生 GUI 与代码动作之间的路由，并在异质动作粒度下保持稳定优化，仍缺乏充分研究。
Nevertheless, how to systematically train a single policy to route between native GUI and code actions while maintaining stable optimization across heterogeneous action granularities remains underexplored. 
Our work addresses this setting through hybrid SFT and online agentic RL with Step-balanced GSPO, mitigating optimization imbalance and enabling more effective native coordination.

\fi
\begin{figure}[t]
  \centering
  \includegraphics[width=1.0\textwidth]{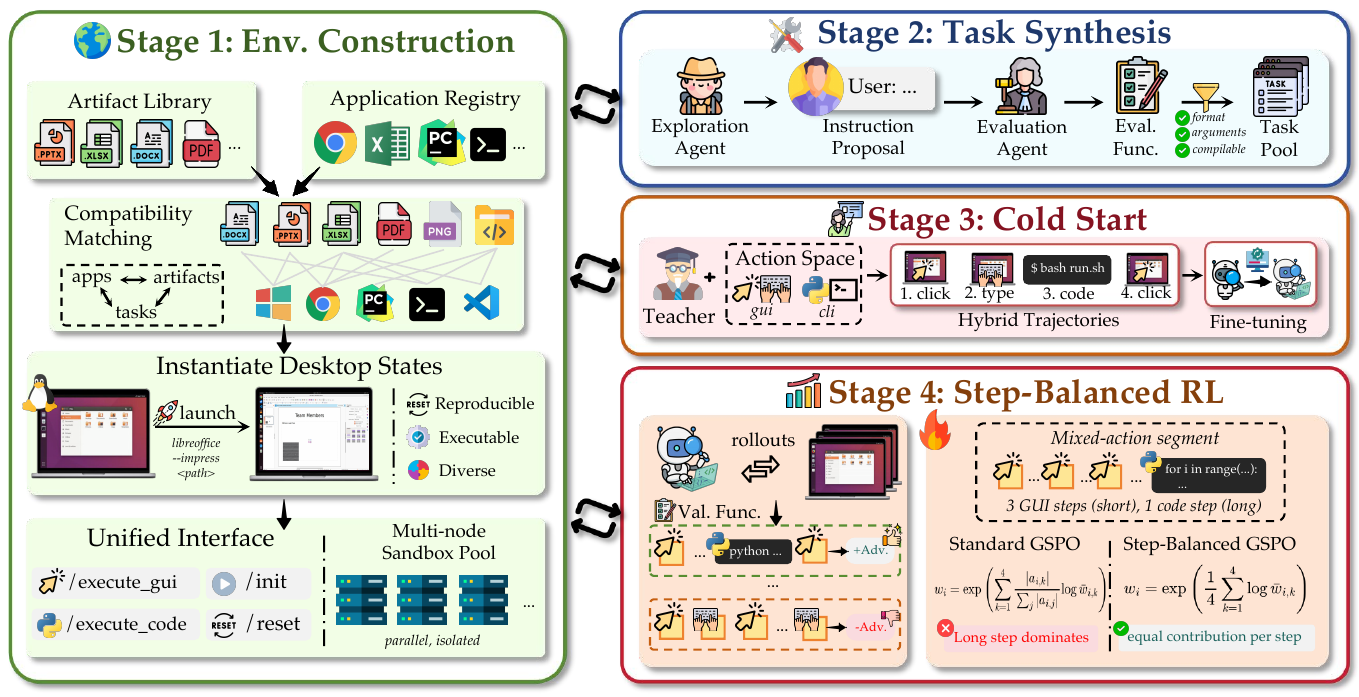}
  % \vspace{1mm}
  % Original caption: The \ours{} pipeline: application--artifact composition for scalable desktop environments, application-centered task synthesis, hybrid trajectory collection, and SFT+RL training for GUI/code collaboration. Appendix~\ref{app:prototype_multiapp} details a graph-guided extension for more complex multi-application tasks.
  \captionsetup{skip=0pt, position=bottom}
  \caption{Overview of the \ours{} pipeline. \textbf{Stage 1} constructs executable desktop environments by composing applications and artifacts. \textbf{Stage 2} explores each environment to synthesize instructions and evaluators, yielding verifiable tasks. \textbf{Stage 3} uses teacher-generated GUI/code trajectories for SFT cold start. \textbf{Stage 4} performs online agentic RL in live sandboxes.}
  \vspace{-10pt}
  \label{fig:method} 
\end{figure}

\section{\ours{}}
\label{sec:method}

\subsection{Overview}
\label{subsec:overview}

% \ours{} 建立在一个简单的前提之上：真正的计算机使用智能体应当将 GUI 交互与代码执行视为改变同一数字状态的两种互补方式。我们并非在推理阶段为 GUI 模型简单外挂一个 shell 工具，而是构建了一条统一流水线，使环境、指令、轨迹与强化学习信号均面向 GUI/代码混合执行进行设计。
An overview of the \ours{} pipeline is shown in Figure~\ref{fig:method}. \ours{} is built around a simple premise: a real computer-using agent should treat GUI interaction and code execution as two complementary ways of changing the same digital state. Rather than attaching a shell tool to a GUI model at inference time, we construct a unified pipeline in which environments, instructions, trajectories, and reinforcement signals are all designed for hybrid GUI/code execution.

% 形式化地，令 \(q\) 表示用户指令，\(h_i\) 表示第 \(i\) 步之前的历史上下文，\(o_i\) 表示当前屏幕观测。策略从统一的混合动作空间中选择动作，其中 \(\mathcal{A}_{\mathrm{meta}}\) 表示任务终止等 episode-level meta-action。在此基础上，我们加入代码动作，使智能体能够直接在操作系统中执行 Python 或 Bash。代码执行产生的标准输出与标准错误会作为工具响应追加到当前轮上下文，并供后续决策使用。
Formally, let \(q\) denote the user instruction, \(h_i\) the history context before step \(i\), and \(o_i\) the current screen observation. The policy selects an action from a unified hybrid action space:
\[
    a_i \sim \pi_\theta(\cdot \mid q,h_i,o_i),
    \qquad
    a_i \in \mathcal{A}_{\mathrm{hybrid}}
    = \underbrace{\mathcal{A}_{\mathrm{GUI}} \cup \mathcal{A}_{\mathrm{meta}}}_{\mathcal{A}_{\mathrm{native}}}
    \cup \mathcal{A}_{\mathrm{code}}.
\]
Here, $\mathcal{A}_{\mathrm{GUI}}$ denotes GUI actions such as clicking and typing; \(\mathcal{A}_{\mathrm{meta}}\) denotes episode-level meta-actions such as task termination; actions in \(\mathcal{A}_{\mathrm{code}}\) execute either Python or Bash directly in the operating system. 
Each code action runs in a fresh process. Its stdout and stderr are appended to the current-turn context as a tool response for subsequent decision-making.
\subsection{Data Infrastructure}
\label{sec:data_infra}

% 计算机使用智能体训练的规模，最终受限于可产生有意义任务的可执行状态的多样性。这一点在桌面环境中尤其重要，因为桌面工作流是由应用、用户产物以及持久化的本地状态共同塑造的。我们将真实产物视为扩展这一状态空间的自然维度，其中，文件类型提供了与能够操作这些文件的应用之间的直接关联。基于这一原则，我们构建多样化的可执行桌面环境，并从这些环境状态中合成有依据、可验证的任务。
The scale of computer-using agent training is ultimately bounded by the diversity of executable states from which meaningful tasks can arise. This is particularly important on desktops, where workflows are shaped jointly by applications, user artifacts, and persistent local state. We view authentic artifacts as a natural axis for expanding this state space, with artifact types and file formats providing a direct link to the applications that can operate on them. Based on this principle, we construct diverse executable desktop environments and synthesize grounded, verifiable tasks from the resulting states.

% Avoid stretching inter-paragraph spacing on this shortened page.
\raggedbottom
\subsubsection{Application--Artifact Composition for Desktop Environments}
\label{sec:file_infra}
% 真实世界中的桌面工作流通常建立在已有产物之上，这些产物会通过应用被读取、编辑或转换。我们因此将可扩展桌面环境的构建分解为两类可复用组件：具有固定版本、启动方式和多种初始化配置的应用，以及携带真实内容与持久状态的产物。二者依据兼容关系组合为可执行的桌面状态，现实任务也由此产生。
Real-world desktop workflows are commonly grounded in existing artifacts that are read, edited, or transformed through applications. Thus, we factor scalable desktop-environment construction into two reusable components: applications with fixed versions, launch procedures, and diverse initialization configurations, and artifacts carrying authentic content and persistent state. Their compatibility-aware composition produces executable desktop states from which realistic tasks can arise.

% 稍微举几个例子介绍一下 file type 和应用的对应关系
% 为了大规模实例化这类状态，我们将产物和兼容应用组织成一个 \emph{应用--产物兼容图}，其中边由产物的文件类型决定。底层的应用配置记录了每个应用支持的文件类型，以及实现可复现执行所需的版本、启动命令和初始化设置。例如，一个 \texttt{.pptx} 产物可以连接到 LibreOffice Impress 和 WPS PowerPoint，相应的应用配置会指定如何使用文件启动各个应用，以及所需的应用特定初始化设置。由此得到的状态会在可复现的 Ubuntu 虚拟机中实例化，用于任务合成和轨迹收集。
To instantiate such states at scale, we organize artifacts and applications into an \emph{application--artifact compatibility graph}, whose edges are determined by the artifact types and file formats that applications can consume. The underlying application configuration records these supported input types, together with the version, launch commands, and initialization settings required for reproducible execution. For example, a \texttt{.pptx} artifact can be connected to LibreOffice Impress and WPS Presentation directly, whose application configurations specify how each application is launched with the file and which application-specific initialization settings are required. The resulting states are instantiated on reproducible Ubuntu virtual machines for task synthesis and trajectory collection.

\subsubsection{Environment-Grounded Task Synthesis}
\label{sec:task_synthesis}

% 在构建上述桌面环境之后，我们进一步将环境多样性转化为训练任务多样性。每个合成 episode 首先采样应用，然后通过 \emph{应用--产物兼容图} 选择兼容产物，从而实例化一个具体的任务上下文。
% Building upon the diverse desktop environments constructed above, we next translate environment-level diversity into a broad and structured distribution of training tasks. Specifically, each task-synthesis episode begins by sampling a target application, after which the \emph{application--artifact compatibility graph} is used to identify and sample artifacts that are semantically and functionally compatible with the selected application. The sampled application-artifact combination then instantiates a concrete task context, providing the foundation for synthesizing executable and environment-grounded training tasks.
Given the diverse desktop environments constructed above, we next translate environment diversity into training-task diversity. Each synthesis episode first samples a target application and then selects compatible artifacts via the \emph{application--artifact compatibility graph}, thereby instantiating a coherent and executable task context.

% 金哥我来了
% 是否应该写成公式（agent的input、output），需要讨论
% ybw: 解决多样性不足方案 = 提前探索 + 动态维护覆盖功能
% ybw: 静态应用教程 + 动态的当前应用覆盖功能（全局记忆）
% 主要依赖生成模型的先验知识容易产生泛化且重复的指令模式，并且难以充分利用大规模实例化环境中的状态多样性。因此，我们通过主动探索当前环境来生成指令。为避免反复覆盖相同功能，我们维护一份全局应用记忆，记录各应用此前已经涉及的功能，并在后续探索时提供这些历史信息，从而逐步扩展功能覆盖。探索智能体接收该记忆以及应用元数据、精确文件路径、启动配置、教程和初始截图，并执行少量仅限打开、点击、滚动和终止的短暂非破坏性探索。随后，它根据实际观察到的产物内容与应用状态提出真实的用户指令，并给出初始化配置和可观察的成功条件；指令不预设 GUI 或代码解法，也不泄露与答案相关的信息。
Relying primarily on model priors tends to produce generic, repetitive instruction patterns and underutilize the state diversity of instantiated environments at scale. We therefore ground instruction generation in active exploration of the current environment. To avoid repeatedly covering familiar functionality, we maintain a global application memory that records functions previously exercised for each application and exposes this history during subsequent exploration, encouraging broader coverage over time. The \textbf{Exploration Agent} receives this memory together with application metadata, exact file paths, launch configuration, tutorial, and initial screenshot, and performs a short non-destructive exploration restricted to opening, clicking, scrolling, and terminating. Based on the observed artifact content and application state, it proposes a realistic user instruction together with its initialization configuration and observable success conditions, without prescribing a GUI/code solution or revealing answer-critical information.

% ybw: 以文件为中心，构建文件—应用图；先采样应用，再采样文件
% 为了使合成任务能够被一致地验证，我们采用一种 \emph{一切皆文件} 的抽象，其中每个任务的终止状态都由虚拟机内部的一个或多个持久化产物捕获。这些产物既包括面向用户的文件，也包括由文件承载的应用或系统状态，例如配置和缓存。该抽象将异构的桌面目标映射到一种通用表示，使其通常可以通过文件内容、结构、元数据或配置进行程序化检查。基于文件的验证会指定期望的终止状态，而不约束达到该状态的方式，因此同一任务可以通过 GUI 交互、代码执行，或二者结合来完成。
To make synthesized tasks consistently verifiable, we adopt an \emph{everything-is-a-file} abstraction in which terminal conditions are mapped, whenever possible, to inspectable file representations, and evaluator synthesis selects rule-based, VLM-based, or hybrid checks according to their observability. Verification specifies the desired terminal state without constraining how it is reached, allowing the same task to be completed through GUI interaction, code execution, or their combination.

% 应用统一称为 Application / App，指令 + 评估器 = 任务
% 中文：我们将指令提议与评估器合成解耦。一旦指令提议通过筛选，其指令和成功条件即被冻结，并传递给独立的评估器智能体，由其编译相应的验证逻辑。可通过程序观测的条件主要由基于规则的检查评估虚拟机文件和命令行状态；无法通过程序可靠恢复的视觉条件则使用基于 VLM 的评估。
We decouple the instruction proposal from evaluator synthesis. Once an instruction proposal is accepted, its instruction and success conditions are frozen and passed to an independent \textbf{Evaluator Agent}, which compiles the corresponding verification logic. Programmatically observable conditions are evaluated primarily with rule-based checks over VM files and command-line state, while visual conditions that cannot be reliably recovered programmatically are handled with VLM-based evaluation.
% 评估器必须能够编译，并且需要对初始状态给出零奖励；同时，指令会从可达性、无歧义性、真实性以及验证条件是否充分等方面进行检查。无效的评估器会在可能时被修复，语义上近似重复的任务会在过滤后被移除。环境驱动的任务合成与基于文件的验证结合起来，产生了一个可扩展的可验证任务池，供 SFT 轨迹收集和在线 RL 共同使用。
% Evaluators must compile and assign zero reward to the initial state, while instructions are checked for reachability, unambiguity, realism, and sufficient verification conditions. Invalid evaluators are repaired when possible, and semantically near-duplicate tasks are removed after filtering. Together, environment-grounded task synthesis and file-backed verification produce a scalable pool of verifiable tasks shared by SFT trajectory collection and online RL.
% 中文：指令与评估器在流水线中分别接受验证。指令需要满足可达性、无歧义性、真实性以及验证条件充分性；评估器则必须能够编译并对初始状态给出零奖励，无效评估器会在可能时进行修复。通过验证的指令与评估器共同构成任务，随后再移除语义上近似重复的任务。环境驱动的合成与基于文件的验证由此形成一个可扩展的可验证任务池，供 SFT 轨迹收集和在线强化学习共同使用。
Instructions and evaluators are validated separately in the pipeline. Instructions are checked for reachability, unambiguity, realism, and sufficient verification conditions, while evaluators must compile and assign zero reward to the initial state. Invalid evaluators are repaired when possible. Each validated instruction and evaluator then constitutes a task, after which semantically near-duplicate tasks are removed. Together, environment-grounded synthesis and file-backed verification produce a scalable pool of verifiable tasks shared by SFT trajectory collection and online RL. Data infrastructure details are provided in Appendix~\ref{app:data_details}.

% ybw: 图2 Explore Agent + Evaluator Agent(全文同步说辞)，去掉 Judge Agent（简单用一个漏斗之类的表达下就可以）; SFT 部分精简再精简；RL 部分，去掉 N Rollout 这种所有XXPO都有的流程表述；看看如何抽象表达 GUI Code Token 不平衡的问题 + 我们的应对方法(可以是公式对比 / 类似讨论图里的曲线对比，可以fake)
% ==================================

\enlargethispage{\baselineskip}
\subsection{Training Paradigm}
\label{subsec:training_paradigm}

% 给定可验证任务池，我们分两个阶段训练 \ours{}，且两个阶段承担不同作用。
% SFT 通过模仿高质量混合轨迹，使原本以 GUI 操作为主的开源模型获得可用的统一动作接口和初步模态路由能力；在线强化学习则不再学习“有哪些动作”，而是依据可验证的桌面环境回报，优化长时程过程中“何时选择何种动作以及如何从错误中恢复”。为适配异构 GUI/代码轨迹，我们在强化学习阶段采用 Step-balanced GSPO。
% Given the task pool, we train \ours{} in two stages. 
% SFT imitates high-quality hybrid trajectories to equip originally GUI-centric open-source models with a usable unified action interface and an initial modality-routing policy. 
% Online RL no longer learns merely \emph{which actions exist}; instead, it optimizes \emph{when to choose each action and how to recover from errors} over long horizons using verifiable desktop returns. We use Step-balanced GSPO in this RL stage to accommodate heterogeneous GUI/code trajectories.

\subsubsection{SFT Cold Start}
\label{subsec:sft}

% 大多数作为基础模型的开源 CUA 此前主要围绕纯 GUI 交互进行训练：它们可能具备视觉理解、坐标定位和低层操作能力，却没有学习把代码执行作为同等地位动作之后所产生的新决策边界。仅仅加入代码工具会扩大动作空间，却不会自动提供何时保留 GUI 操作、何时切换到代码以及何时返回界面验证的路由先验，甚至可能扰乱模型已经形成的 GUI 行为。因此，SFT 在本文中承担冷启动作用：它建立统一工具模式的动作语义，并通过教师轨迹赋予模型粗粒度但可用的 GUI/代码路由策略。
Most open-source CUAs used as base models have previously been trained primarily for GUI-only interaction. They may possess visual understanding, coordinate grounding, and low-level control, but have not learned the new decision boundary created when code execution becomes an equally valid action. Merely adding a code tool expands the action space without supplying a routing prior for when to remain in the GUI, when to switch to code, and when to return to the interface for verification; it can even disrupt already learned GUI behavior. SFT therefore serves as a cold start in our pipeline: it establishes the action semantics of the unified tool schema and distills from teacher trajectories a coarse but usable GUI/code routing policy. 
% SFT 的目标是让混合策略首先变得可执行，而不是直接优化长时程任务回报；后者由在线强化学习在路由能力稳定之后，根据真实环境结果继续完成。
% Its role is to make the hybrid policy operational, not to directly optimize long-horizon task return. The latter is addressed by online RL after routing has been stabilized, using outcomes from the live environment.
% 我们在第~\ref{subsec:data_infrastructure} 节的数据基础设施所生成的任务上采集 SFT 轨迹。为降低模型对单一教师特有行为的过拟合，我们使用多种前沿模型，包括 Claude 4.5 系列~\citep{anthropic2025claude4.5}、Claude 4.6 系列~\citep{anthropic2026claude4.6} 和 Gemini 3.1 Pro~\citep{gemini31proeval}。这些模型原生的计算机操作会通过后处理被规范化到 \ours{} 的动作模式中。每条轨迹都与任务提议阶段之后独立合成的对应评估器配对，并以评估器得分作为拒绝采样的主要信号。随后，我们使用过滤流水线移除格式错误的工具调用、与模式不匹配的动作、低质量轨迹以及未通过评估器检查的轨迹，最终得到一组高质量 SFT 样本。训练超参数和数据混合细节见附录~\ref{app:sft_config}。
% 我们在第~\ref{subsec:data_infrastructure} 节的数据基础设施所生成的任务上使用教师模型Claude-Sonnet-4.5同时配置ComputerUse和BashTool采集 SFT 轨迹。随后轨迹会通过后处理被规范化到 \ours{} 的动作模式中。每条轨迹都被评估器评分(若涉及视觉评分则采用教师模型自身)，并作为拒绝采样的主要信号。随后，我们使用过滤流水线移除格式错误的工具调用、与模式不匹配的动作、低质量以及未通过评估器检查的轨迹，最终得到一组高质量 SFT 样本。训练超参数和数据混合细节见附录~\ref{app:sft_config}。
We collect SFT trajectories on tasks produced by the data infrastructure described in Section~\ref{sec:data_infra}, using Claude Sonnet 4.6~\citep{anthropic2026claude4.6} as the teacher model with both Computer Use and Bash Tool enabled. The resulting trajectories are then postprocessed and normalized into the \ours{} action schema. 
Each trajectory is scored by its evaluator, using the teacher model itself when visual judgment is required, and the resulting score serves as the primary signal for rejection sampling. We further apply a filtering pipeline to remove malformed tool calls, schema-mismatched actions, and trajectories that fail evaluator or quality checks, yielding a set of high-quality SFT samples. Training hyperparameters and data-mixture details are provided in Appendix~\ref{app:sft_config}.

\flushbottom
\subsubsection{Agentic RL with Step-Balanced GSPO}
\label{subsec:gspo}
% 逐句 Check
% 我们在经过谨慎筛选的仅包含rule-base评估器的生成任务上执行在线强化学习，以在与 SFT 和评测相同的 GUI/代码混合执行框架中提升长时程任务成功率。每条轨迹最多与实时桌面环境交互 50 步，配套环境评估器的得分构成唯一奖励信号。如果某个 assistant 轮次包含无法解析的工具调用、不属于既定模式的动作或其他无效动作格式错误，则该 assistant 响应会被单独分离为无效样本并被赋予奖励 \(0\)，即使轨迹其余部分可能获得正的评估器得分。有效片段继承取值位于 \([0,1]\) 的评估器得分。
% We perform online RL on carefully filtered generated tasks with only rule-based evaluators to improve long-horizon success in the same harness used for SFT and evaluation. Each rollout interacts with a live desktop environment for up to 50 steps, and the paired environment evaluator, provides the sole reward signal. If an assistant turn contains an unparsable tool call, a non-schema action, or another invalid action-format failure, that assistant response is isolated as an invalid sample and assigned reward \(0\), even when the remaining trajectory might receive a positive evaluator score. Valid segments inherit evaluator scores in \([0,1]\).
% 我们首先构建了一套跨节点环境管理工具，允许监控管理多台单机的百余个Ubuntu环境sandbox，对训练端暴露统一端口，在内部分发请求。接着，我们在谨慎筛选的仅包含rule-base评估器的生成任务上执行在线强化学习以进一步优化模型性能。每条轨迹与实时桌面环境交互，配套评估器的得分构成唯一奖励(outcome reward)信号。如果某个 assistant 轮次包含无法解析的工具调用、不属于既定模式的动作或其他无效动作格式错误，则该 assistant 响应会被单独分离为无效样本并被赋予奖励 \(0\)，即使轨迹其余部分可能获得正的评估器得分。
We first build a cross-node environment management system that monitors and manages over one hundred Ubuntu sandbox environments across multiple standalone machines. The system exposes a unified port to the training side and internally dispatches requests to the environments. We then perform online RL on carefully filtered generated tasks with rule-based evaluators only, further improving model performance. Each rollout interacts with a live desktop environment, and the paired evaluator score serves as the sole outcome reward signal. If an assistant turn contains an unparsable tool call or any other invalid action-format error, that assistant response is isolated as an invalid sample and assigned a reward of \(0\), even if the remaining trajectory could receive a positive evaluator score.

% 三项效率机制使高成本桌面交互下的在线强化学习变得切实可行。第一，我们采用异步强化学习流水线，并允许策略版本最多滞后一个版本。将轨迹采集与策略优化解耦后，环境交互可以与模型更新并行重叠，从而显著减少流水线中的空闲气泡。第二，我们维护逐任务的成功率估计，并动态采样位于当前能力边界附近的任务：已经过于简单或当前仍过于困难的任务会被降低权重，而成功率居中的任务会被更频繁地采样。
We use three efficiency mechanisms to make online RL feasible under costly desktop interaction. 
First, we adopt an asynchronous RL pipeline that allows the rollout policy to lag behind the latest policy by at most one version, enabling environment interaction and policy optimization to run in parallel. 
Second, we maintain per-task success-rate estimates and bias sampling toward tasks near the current capability frontier, down-weighting tasks that are already too easy or remain too difficult. 
% 第三，我们采用视觉片段训练。保留全部历史截图会产生难以承受的长上下文，而为每个动作步骤分别生成训练样本又会使样本数量随轨迹长度线性增长。因此，我们每隔固定步数将轨迹切分为保留有界图像历史的片段，每个片段包含多个参与损失计算的 assistant 动作，同时在rollout推理上保持相同逻辑，以达到训推一致。
% 第三，我们对视觉上下文进行分段处理，以控制上下文长度和样本增长。Rollout被切分为若干片段，每段历史截图数量限定在5至10张，每个片段可包含多个参与损失计算的助手动作。我们在rollout推理阶段采用相同的分段逻辑，以保持训练与推理之间的一致性。
Third, we apply bounded sliding-window visual-context segmentation to control context length and sample growth. Rollouts are split into segments and each segment can contain multiple assistant actions that contribute to loss. 
We apply the same sliding-window segmentation logic during rollout inference to maintain consistency between training and inference. 
% Sampling and segmentation details are provided in Appendix~\ref{app:rl_sampling_segmentation}.

% 中文：我们使用标准 GSPO 的初步强化学习实验呈现出系统性地向代码密集行为偏移的现象，如图~\ref{fig:step_balance_code_ratio} 所示。我们将其归因于序列级策略优化与异构动作空间之间的固有失配。在对视觉上下文切分后得到的 GUI/代码混合训练片段中，代码动作所包含的 assistant token 往往远多于 GUI 点击或按键动作。标准 GSPO 会对片段中所有有效 assistant token 的 token-level log-importance weight 求平均；因此，即使一条长代码动作只是多个 GUI 与代码步骤中的一步，它仍可能主导片段级 importance ratio 和裁剪决策。这种 token 长度偏置会使策略越来越偏好代码，其原因并非代码始终是正确模态，而是代码密集步骤对优化施加了不成比例的影响。
% - Our preliminary RL experiments with standard GSPO exhibited a systematic drift toward code-intensive behavior, as shown in Panel~(a) of Figure~\ref{fig:training_behavior}.
% - We attribute this phenomenon to an inherent mismatch between sequence-level policy optimization and heterogeneous action spaces.
% - Within a mixed GUI/code training sample, a code action typically contains substantially more assistant tokens than a GUI click or keystroke.
% - Standard GSPO averages token-level log-importance weights over all valid assistant tokens in the sample. Consequently, even when a long code action constitutes only one of several GUI and code steps, it can dominate both the sample-level importance ratio and the clipping decision. This token-length bias increasingly steers the policy toward code, not because code is invariably the appropriate modality, but because code-intensive steps exert disproportionate influence on optimization.
In our preliminary RL experiments, standard GSPO exhibited a systematic drift toward code-intensive behavior as shown in Figure~\ref{fig:training_behavior}. Specifically, as training progresses, the model increasingly favors code-related actions, ultimately resulting in a slight performance degradation relative to the initial SFT checkpoint. 
Within a mixed GUI/code training sample, a code action typically contains substantially more assistant tokens than a GUI click or keystroke.
Standard GSPO averages token-level log-importance weights over all valid assistant tokens in the sample. Thus, even when a long code action constitutes only one of several GUI and code steps, it can dominate both the sample-level importance ratio and the clipping decision. 
This motivates an aggregation in which the weight assigned to each action step in the sample-level log-ratio is no longer determined by its token count.

% Step-balanced GSPO 仅修改样本级比率的聚合方式，而不改变 token-level importance weight 本身。对于训练样本 \(i\)，令 \(a_{i,k}\) 表示动作步骤 \(k\) 中有效 assistant token 位置的集合，\(\tau_{i,k}\) 表示该步骤的动作类型，并定义 token-level importance weight \(w_{i,t}(\theta)=\pi_\theta(y_{i,t}\mid h_{i,t})/\pi_{\mathrm{old}}(y_{i,t}\mid h_{i,t})\) 及其对数 \(\delta_{i,t}=\log w_{i,t}(\theta)\)。我们首先计算步骤级平均 log-importance weight，随后据此定义步骤平衡的样本级 importance ratio。在主要设置中，GUI 步骤与代码步骤使用相同权重，因此无论 token 长度如何，每个动作步骤都对样本级比率作出等量贡献。
Step-balanced GSPO modifies only the aggregation of the sample-level ratio. For training sample \(i\), let \(a_{i,k}\) be the set of assistant-token positions that contribute to the loss for action step \(k\), and let \(\tau_{i,k}\) denote its action type. Define the token-level importance weight \(w_{i,t}(\theta)=\pi_\theta(y_{i,t}\mid h_{i,t})/\pi_{\mathrm{old}}(y_{i,t}\mid h_{i,t})\) and its log-importance weight \(\delta_{i,t}=\log w_{i,t}(\theta)\). We first compute the mean token-level log-importance weight for each action step
\[
    \bar{\delta}_{i,k}
    =
    \frac{1}{|a_{i,k}|}\sum_{t\in a_{i,k}}\delta_{i,t},
\]
and then define the step-balanced sample-level ratio as
\[
    w_i^{\mathrm{step}}(\theta)
    =
    \exp\left(
    \frac{\sum_{k=1}^{K_i}\alpha_{\tau_{i,k}}\bar{\delta}_{i,k}}
         {\sum_{k=1}^{K_i}\alpha_{\tau_{i,k}}}
    \right).
\]
In our setting, GUI and code steps use equal weights, so every action step contributes equally to the sample-level log-ratio regardless of its token count.

% 随后，策略目标沿用标准 GSPO 的计算公式，仅改变用于形成样本级 importance ratio 的聚合方式，将原始序列级聚合替换为 \(w_i^{\mathrm{step}}(\theta)\)。这样可以消除由 token 长度决定的隐式动作权重，避免长代码动作仅因包含更多 token 而对优化产生不成比例的影响。完整推导见附录~\ref{app:step_balanced_gspo}。
The policy objective follows the standard GSPO formulation and changes only the aggregation used to form the sample-level importance ratio, replacing the original sequence-level aggregation with \(w_i^{\mathrm{step}}(\theta)\). This substitution does not, in general, equalize full parameter-gradient norms or variances across action lengths. Appendix~\ref{app:step_balanced_gspo} provides the full derivation.

\section{Experiments}
\label{sec:experiments}
\begin{table}[t]
\vspace{-0.8\baselineskip}
\centering
\footnotesize
\setlength{\tabcolsep}{2.2pt}
\renewcommand{\arraystretch}{0.92}

\begin{tabularx}{\textwidth}{l c *{6}{>{\centering\arraybackslash}X}}
\toprule
\multirow{2}{*}{\textbf{Method}} & \multirow{2}{*}{\textbf{Setting}}
& \multicolumn{6}{c}{\textbf{Success Rate (\%)}} \\
\cmidrule(lr){3-8}
& & \textbf{OS} & \textbf{Office} & \textbf{Daily} & \textbf{Prof.} & \textbf{Work.} & \textbf{Avg.} \\

\midrule
\noalign{\vskip -0.7ex}
\rowcolor{lightgray}
\multicolumn{8}{c}{\textcolor{darkgray}{\textbf{Agentic Frameworks}}} \\[-0.7ex]
\midrule
UiPath w/ GPT-5~\citeyearpar{cristescu2025ui} & GUI & 73.91 & 49.52 & 62.12 & 71.43 & 37.30 & 53.69 \\
CoAct-1 w/ GPT-5~\citeyearpar{song2025coact} & Hybrid & 75.00 & 62.93 & 57.94 & 71.43 & 47.87 & 59.93 \\
GTA1 w/ GPT-5~\citeyearpar{yang2025gta1} & GUI & 79.17 & 63.91 & 62.56 & 79.59 & 50.91 & 63.41 \\
Agent S3 w/ GPT-5~\citeyearpar{gonzalez2025unreasonable} & Hybrid & 77.50 & 66.46 & 61.23 & 69.80 & 51.37 & 62.63 \\
OS-Symphony w/ GPT-5~\citeyearpar{OS-Symphony} & Hybrid & 79.17 & 65.73 & 67.76 & 69.23 & 57.98 & 65.84 \\

\midrule
\noalign{\vskip -0.7ex}
\rowcolor{lightgray}
\multicolumn{8}{c}{\textcolor{darkgray}{\textbf{General Models \& Specialist Models}}} \\[-0.7ex]
\midrule
Claude Sonnet 4.5~\citeyearpar{anthropic2025claude4.5} & GUI & 70.83 & 72.59 & 61.35 & 63.27 & 49.54 & 62.88 \\
OpenCUA-72B~\citeyearpar{wang2025opencuaopenfoundationscomputeruse} & GUI & 61.13 & 44.73 & 49.95 & 72.58 & 22.16 & 44.91 \\
UI-TARS~\citeyearpar{qin2025uitars} & GUI & 41.67 & 50.42 & 55.69 & 51.02 & 14.66 & 41.85 \\
UI-TARS-2~\citeyearpar{wang2025uiTARS2} & GUI & 41.67 & 61.11 & 62.12 & 61.22 & 34.13 & 53.10 \\
DeepMiner-Mano-72B~\citeyearpar{fu2025mano} & GUI & 66.67 & 63.22 & 52.51 & 83.67 & 24.41 & 53.91 \\
EvoCUA-32B~\citeyearpar{EvoCUA} & GUI & 78.26 & 59.79 & 64.55 & 81.63 & 27.89 & 56.73 \\
EvoCUA-1.5-32B~\citeyearpar{huang2026evocua} & GUI & -- & -- & -- & -- & -- & 63.20 \\
Kimi K2.5~\citeyearpar{kimi2026k25} & GUI & 73.91 & 69.20 & 66.42 & 73.47 & 46.11 & 63.34 \\
Kimi K2.6~\citeyearpar{kimi2026k26} & GUI & 79.17 & 80.03 & 77.11 & 81.63 & 54.97 & 73.07 \\
Qwen3.7-Plus~\citeyearpar{qwen2026qwen37} & GUI & 79.17 & 82.91 & 74.03 & 73.47 & 59.25 & 73.33 \\

\midrule
\noalign{\vskip -0.7ex}
\rowcolor{lightgray}
\multicolumn{8}{c}{\textcolor{darkgray}{\textbf{Backbone Baselines}}} \\[-0.7ex]
\midrule
Qwen3.5-27B~\citeyearpar{qwen2026qwen35}$^{\clubsuit}$ & GUI & 66.67 & 39.60 & 57.50 & 52.00 & 36.06 & 46.61 \\
Qwen3.5-27B~\citeyearpar{qwen2026qwen35}$^{\clubsuit}$ & Hybrid & 66.67 & 39.26 & 58.79 & 49.96 & 38.29 & 46.50 \\
Qwen3.5-35B-A3B~\citeyearpar{qwen2026qwen35}$^{\clubsuit}$ & GUI & 37.50 & 46.92 & 44.57 & 38.77 & 25.65 & 39.20 \\
Qwen3.5-35B-A3B~\citeyearpar{qwen2026qwen35}$^{\clubsuit}$ & Hybrid & 69.57 & 40.94 & 46.21 & 52.22 & 31.01 & 42.84 \\
\midrule
\ours{35B-A3B} & Hybrid & \losscell{60.87}{8.7} & \gaincell{70.00}{29.1} & \gaincell{52.19}{6.0} & \gaincell{56.31}{4.1} & \gaincell{45.14}{14.1} & \gaincell{57.27}{14.4} \\
\ours{27B} & Hybrid & \gaincell{75.00}{8.3} & \gaincell{72.60}{33.3} & \losscell{58.32}{0.5} & \gaincell{56.49}{6.5} & \gaincell{54.28}{16.0} & \gaincell{62.75}{16.3} \\
\ours{27B-RL} & Hybrid & \gaincell{82.61}{15.9} & \gaincell{69.14}{29.9} & \gaincell{65.01}{6.2} & \gaincell{79.59}{29.6} & \gaincell{53.32}{15.0} & \gaincell{66.44}{19.9} \\
\bottomrule
\end{tabularx}

\captionsetup{skip=2pt,position=bottom,font=footnotesize}
\caption{Main results of \ours{} on OSWorld. $^{\clubsuit}$ marks results reproduced by us; other baseline results are sourced from the official leaderboard. OS contains system-level OS tasks; Office contains LibreOffice Calc, Impress, and Writer tasks; Daily contains Chrome, Thunderbird, and VLC tasks; Prof. contains GIMP and VS Code tasks; and Work. denotes Workflow tasks involving multiple applications.}
\label{tab:main_results}
\vspace{-1\baselineskip}
\end{table}

\subsection{Experimental Setup}
\label{subsec:evaluation_setup}
\label{subsec:baselines}

\noindent\textbf{Evaluation Benchmarks.}
% 同 OS-Symphony-1
Our evaluation centers on desktop environments, employing OSWorld\footnote{In this paper, ``OSWorld'' refers to ``OSWorld-Verified''.}~\citep{xie2024osworld,osworld_verified} as our primary benchmark, which comprises 369 real-world tasks across five domains in an Ubuntu environment. Following common practice~\citep{xie2024osworld}, we exclude the 8 Google Drive tasks, yielding a final set of 361 tasks. We report mean success rate as the primary metric.
% 此外，为进一步验证模型在不同操作系统环境中的泛化性，我们在附录中报告 WindowsAgentArena 结果。正文主实验仍以 OSWorld 为核心，以保证主要结论建立在同一评测环境中的受控比较上。
Beyond the Ubuntu environment, we additionally report results on WindowsAgentArena~\citep{bonatti2024windows} to evaluate cross-platform transfer to Windows.

\noindent\textbf{Models and Baselines.}
% 我们基于 Qwen3.5-27B 与 Qwen3.5-35B-A3B 训练我们的模型，由于我们是在较早且尺寸较小的Qwen3.5系列上训练，因此我们更关注于相同主干下的受控比较，而不是与最新的基础模型进行直接比较。
We instantiate \ours{} from Qwen3.5-27B and Qwen3.5-35B-A3B~\citep{qwen2026qwen35}.
% This matched-backbone design isolates the contribution of our data and training pipeline from gains attributable to scaling or subsequent advances in base models; direct comparisons with newer or substantially larger foundation models are therefore treated as contextual rather than controlled evidence.
% 我们将三类agent列入主表~\ref{}，分为Agentic Framework，专有与通用模型 与 Baselines，涵盖不同尺寸，并特别标注其动作空间。
For broader context, we also report representative agentic frameworks, general-purpose models, specialist models, and matched backbone baselines. 

\noindent\textbf{Implementation Details.}
For task synthesis, we use Claude Sonnet 4.6~\citep{anthropic2026claude4.6} for the Exploration and Evaluator Agents, and GPT-5.5~\citep{openai2026gpt55} for an independent quality review agent.
% 我们的 SFT 模型在约 31K 条由本文流水线合成并过滤后的成功桌面样本与 40K 条涵盖代码、推理等领域的通用多模态样本上训练一个 epoch。这些桌面轨迹采集自包含 8,880 条合成指令的任务池；完整后处理前的桌面样本池约为 84K。RL 模型从 \ours{27B} 检查点开始，从任务池中选择约 1K 个任务，并以评估器得分作为奖励信号。我们报告平均成功率作为主要指标，并在第~\ref{sec:discussion} 节进一步分析动作空间分布与执行效率。完整的数据细节见附录~\ref{app:data_details}，优化和评测执行框架细节见附录~\ref{app:training_evaluation_details}。
For SFT, we train for one epoch on 31K successful filtered desktop samples synthesized by our pipeline and 40K general-purpose multimodal samples spanning code, reasoning, and related domains. 
% These desktop trajectories are collected from a task pool with 9K synthesized instructions; before filtering, the desktop SFT sample pool contains about 84K samples. 
For RL, we initialize from the dense \ours{27B} checkpoint and conduct a 1,000-step training run using rule-based evaluators as reward signals.
Complete data details are provided in Appendix~\ref{app:data_details}, while training details are provided in Appendix~\ref{app:training_details}.
% We report mean success rate as the primary metric and further analyze action-space distributions and execution efficiency in Section~\ref{sec:discussion}. 

% 未经特殊说明时，测试最大步数设置为50，最大历史图片数为5，保留历史上下文全部文本，温度设置为0.3；baseline models使用完全相同的harness测试。
Unless otherwise specified, each evaluation episode is capped at 50 interaction steps and retains up to 8 historical screenshots (resolution 1504x832) together with the full textual interaction history. All reproduced baseline results are evaluated using the same harness and environment configuration.

\subsection{Main Results}
\label{subsec:main_results}
\begin{wraptable}{r}{0.40\textwidth}
\vspace{-0.8\baselineskip}
\centering
\footnotesize
\setlength{\tabcolsep}{0pt}
\renewcommand{\arraystretch}{0.9}
\begin{tabular*}{\linewidth}{@{\extracolsep{\fill}}lc@{}}
\toprule
\textbf{Model} & \textbf{Succ. (\%)} \\
\midrule
Agent S3 w/ GPT-5~\citeyearpar{gonzalez2025unreasonable} & 54.10 \\
OS-Symphony w/ GPT-5~\citeyearpar{OS-Symphony} & 63.45 \\
\midrule
Qwen3.5-27B~\citeyearpar{qwen2026qwen35}$^{\clubsuit}$ & 56.41 \\
\ours{27B} & 57.95 \\
\ours{27B-RL} & 63.87 \\
\bottomrule
\end{tabular*}
\captionsetup{skip=2pt, position=bottom,font=footnotesize}
\caption{Main results of \ours{} on WindowsAgentArena. $^{\clubsuit}$ marks the baseline reproduced by us. All entries use a 50-step budget and a hybrid action space.} % In our reproduced Hybrid evaluations, Code actions are restricted to Python only; commands that would otherwise use Cmd/PowerShell are issued through Python system calls.
\label{tab:waa_main_results}
\vspace{-0.9\baselineskip}
\end{wraptable}

% 表~\ref{tab:main_results} 表明，性能提升来源于对混合策略的有效学习，而非仅仅为基础模型开放额外的代码工具。在同主干受控比较下，\ours{27B} 与 \ours{35B-A3B} 分别比对应的 Hybrid baseline 提升 16.25 和 14.43 个百分点。增益主要集中在 Office 与 Workflow，这两类任务普遍涉及结构化产物编辑、精确状态变换以及跨应用协调，与本文通过应用--产物组合构建任务状态、并统一学习 GUI/code 轨迹的设计相吻合。
% Table~\ref{tab:main_results} shows that the gains arise from learning a hybrid policy rather than merely exposing the backbone to a code tool. In matched-backbone comparisons, \ours{27B} and \ours{35B-A3B} improve over their corresponding Hybrid baselines by 16.25 and 14.43 absolute points (35.0\% and 33.7\% relative), respectively. 
% The strongest gains appear in Office and Workflow, where tasks often require structured artifact editing and cross-application coordination. 
Table~\ref{tab:main_results} shows that hybrid-policy training, rather than code-tool availability alone, yields substantial gains under matched-backbone comparisons. \ours{27B} and \ours{35B-A3B} improve over their corresponding Hybrid baselines by 16.25 and 14.43 absolute points, respectively. The largest gains occur in Office (+33.3 and +29.1 points) and Work. (+16.0 and +14.1 points). Office covers Calc, Impress, and Writer, where tasks often combine visually grounded interaction with rendered document states and precise transformations of structured artifacts. The Work. category comprises multi-application tasks that require task-relevant information to be handed off across applications and action modalities. 
These interpretations are consistent with the agent behavior study and representative trajectory audit in Appendix~\ref{app:case_study}.
% Daily 与 OS 上较为有限或存在波动的收益则表明，这种训练并非对所有桌面交互进行均匀强化，而是主要改善需要在视觉操作与程序化状态变换之间进行路由的工作流。
% Daily and OS show mixed rather than uniformly positive changes, suggesting that the training mainly improves workflows that benefit from routing between visual manipulation and programmatic state transformation.

% 在 SFT 策略基础上，Step-balanced GSPO 将 \ours{27B-RL} 进一步提升至 66.44%。在线 RL 并不针对预先定义的某一类任务，而是借助评估器反馈，在 SFT 策略附近扩展所探索的轨迹集合，并可能调整其动作路由与状态变换决策。Professional、OS 与 Daily 上的提升最为明显，而 Office 与 Workflow 基本维持在 SFT 水平附近。这种不均匀模式说明 RL 改变了策略的决策行为，而非均匀地强化某一种任务属性。总体而言，SFT 提供稳定的混合动作语义与路由先验，而评估器引导的 RL 则基于任务级反馈扩展并细化策略行为。
% 从 SFT 策略出发，步骤均衡 GSPO 达到了 66.44%。如表~\ref{tab:paired_bootstrap_ci} 所示，配对自举法证实了相对于匹配混合基线在 SFT 和 RL 上均取得了稳健的提升，而 RL 相对于 SFT 的增量差异在统计上仍不具有确定性。 按类别细分的结果提供了更进一步的解析：最大增益出现在 Professional、OS 和 Daily 类别，而 Office 和 Workflow 类别与 SFT 水平相比变化不大。这表明 RL 的边际贡献集中于特定任务领域，而非均匀地带来整体提升。
% Starting from the SFT policy, Step-balanced GSPO reaches 66.44\%. As shown in Table~\ref{tab:paired_bootstrap_ci}, paired bootstrap confirms robust gains over the matched Hybrid baseline for both SFT and RL, while the incremental RL--SFT difference remains statistically inconclusive. 
% The category breakdown offers further resolution, with the largest gains appearing in Professional, OS, and Daily, while Office and Workflow remain close to their SFT levels, suggesting that RL's contribution concentrates in specific task domains rather than improving uniformly.

Starting from the SFT policy, \ours{27B-RL} further raises the average success rate from 62.75\% to 66.44\%. Its incremental gains are concentrated in Professional (+23.10 points), OS (+7.61 points), and Daily (+6.69 points), whereas Office and Work. do not improve further over their already strong SFT results. This profile shows that RL provides additional gains in specific task domains rather than a uniform increase across the benchmark. The paired-bootstrap analysis of the aggregate difference is reported in Table~\ref{tab:paired_bootstrap_ci}.

% 在非受控的广泛比较中，\ours{27B-RL} 与表中的 GPT-5 agentic frameworks 具有竞争力，但仍落后于更新且规模更大的通用模型。因此，本文的主要结论建立在稳定的同主干增益之上，以排除模型规模和代际差异带来的影响。
In broader, non-controlled comparisons, \ours{27B-RL} is competitive with GPT-5-based agentic frameworks such as Agent-S3~\citep{gonzalez2025unreasonable} and OS-Symphony~\citep{OS-Symphony}, while newer and substantially larger general models remain stronger overall. 
Table~\ref{tab:waa_main_results} further shows that the gains transfer to WindowsAgentArena, where \ours{27B-RL} improves over the Qwen3.5-27B GUI baseline by 23.05 points. This provides evidence of cross-platform transfer on WindowsAgentArena, with detailed results deferred to Appendix~\ref{app:waa_generalization}. Notably, the RL checkpoint also exhibits a larger improvement over the SFT checkpoint on WAA, suggesting that our online RL training further enhances the model's generalization capability.

\subsection{Ablation Study}
\label{subsec:ablations}

% 接下来进行三组消融实验。数据消融检验 SFT 增益能否在不同能力水平的轨迹教师下保持，以及该增益是否依赖与评测基准相关的应用域数据。训练阶段与算法消融一方面分离 SFT 和 RL 各自及其组合的贡献，另一方面在其余训练配置完全一致的条件下比较原始 GSPO 与 Step-balanced GSPO。范式消融则检验性能究竟来自统一 GUI/code 策略的学习，还是仅仅来自为 GUI 智能体开放额外的代码工具。除非另有说明，所有变体均采用 Qwen3.5-27B，并在 OSWorld-Verified 上使用与主实验相同的评测框架。
We conduct three groups of ablations: data robustness, training-stage and RL-objective attribution, and the unified GUI/code paradigm. 
Unless otherwise specified, all variants use Qwen3.5-27B and are evaluated on OSWorld with the same harness configuration as the main experiments.

% 数据鲁棒性消融。表~\ref{tab:data_composition_ablation} 检验 SFT 增益是否依赖强轨迹教师或来自 OSWorld 相关应用域的数据。在子表 (a) 中，我们固定相同的 800 个合成任务，分别使用相对较弱的教师 Kimi K2.5~\citep{kimi2026k25} 与更强的教师 Claude Sonnet 4.6~\citep{anthropic2026claude4.6} 独立采集并过滤轨迹，然后使用相同的主干、通用数据混合与优化设置训练两个变体。子表 (b) 采用应用级控制，删除九个 OSWorld 应用域中的全部合成样本，使桌面 SFT 数据从约 31K 减少至 17K，同时保持其余训练配置不变。
\textbf{Data robustness ablation.}
Table~\ref{tab:data_composition_ablation} tests whether the SFT gains require a strong trajectory teacher or data from OSWorld-related application domains. Panel (a) compares Kimi K2.5~\citep{kimi2026k25} and Claude Sonnet 4.6~\citep{anthropic2026claude4.6} trajectories collected from the same 800 synthesized tasks. Panel (b) removes all samples from the nine OSWorld application domains, reducing the desktop SFT set from approximately 31K to 17K samples.
% 两组教师数据均显著优于未经适配的 Hybrid 主干，且更强教师取得更高结果，说明教师能力影响监督质量，但流水线的有效性并不依赖单一教师。
Both teacher-derived datasets outperform the unadapted Hybrid backbone, while the stronger teacher performs better. 
% 更严格的应用级排除进一步表明，即使移除九个相关应用域并大幅缩减桌面数据，性能仍能comparable。该行为展示出模型对数据规模和应用组成变化表现出的稳健性，说明训练所得能力能够跨应用迁移(重点)，而非依赖特定的benchmark领域任务模板。
% Removing the nine related application domains shifts success from 62.75\% to 62.31\%, a comparable result despite the substantially reduced desktop data. This robustness to both data scale and application composition suggests that the learned capabilities transfer across applications rather than relying on benchmark-specific task templates.
Removing the nine related application domains shifts success from 62.75\% to 62.31\%. The retained performance after domain removal is consistent with capabilities that transfer beyond the evaluated application domains.

\begin{table}[htbp]
\vspace{-0.1\baselineskip}
\centering
\footnotesize
\setlength{\tabcolsep}{1.5pt}
\renewcommand{\arraystretch}{0.9}
\begin{minipage}{0.49\linewidth}
\centering
\textbf{(a) Trajectory teacher.}\\[-0.2ex]
\begin{tabular*}{\linewidth}{@{\extracolsep{\fill}}lcc@{}}
\toprule
Teacher & Success (\%) & $\Delta$ vs. base \\
\midrule
Kimi K2.5 & 54.03 & +7.53 \\
Claude Sonnet 4.6 & \textbf{58.77} & \textbf{+12.27} \\
\bottomrule
\end{tabular*}
\end{minipage}%
\hfill%
\begin{minipage}{0.49\linewidth}
\centering
\textbf{(b) OSWorld-domain removal.}\\[-0.2ex]
\begin{tabular*}{\linewidth}{@{\extracolsep{\fill}}lcc@{}}
\toprule
Desktop data & Samples & Succ. (\%) \\
\midrule
Full synthesized pool & $\sim$31K & \textbf{62.75} \\
Nine domains removed & $\sim$17K & 62.31 \\
\bottomrule
\end{tabular*}
\end{minipage}
\captionsetup{skip=2pt, position=bottom,font=footnotesize}
% Original caption: Data robustness ablation. Panel (a) changes the teacher model over the same 800 tasks; deltas are measured against the 46.50\% unadapted Hybrid backbone. Panel (b) removes all nine OSWorld application domain tasks; its delta is measured against the full synthesized pool.
\caption{Data robustness ablations.}
\label{tab:data_composition_ablation}
\vspace{0pt}
\end{table}

% 训练阶段与算法归因。表~\ref{tab:training_stage_ablation} 对训练流水线提供两组互补控制。子表 (a) 固定主干与动作空间，仅改变是否使用 SFT 和 RL。SFT 贡献了相对于未经适配 Hybrid 策略的主要增益，而仅使用 RL 只产生有限变化；在 SFT 冷启动后继续进行 RL 则取得最佳结果。这表明在线探索难以独立建立可靠的 GUI/code 动作语义与模态路由，而在 SFT 初始化出可用的混合策略后才能有效发挥作用。
\begin{table}[htbp]
\vspace{-0.4\baselineskip}
\centering
\footnotesize
\setlength{\tabcolsep}{1.5pt}
\renewcommand{\arraystretch}{0.9}
\begin{minipage}[t]{0.49\linewidth}
\centering
\textbf{(a) Training stages.}\\[-0.2ex]
\begin{tabular*}{\linewidth}{@{\extracolsep{\fill}}lccc@{}}
\toprule
Variant & SFT & RL & Succ. (\%) \\
\midrule
Hybrid Backbone & No & No & 46.50 \\
RL only & No & Yes & 48.20 \\
SFT only & Yes & No & 62.75 \\
SFT $\rightarrow$ RL & Yes & Yes & \textbf{66.44} \\
\bottomrule
\end{tabular*}
\end{minipage}%
\hfill%
\begin{minipage}[t]{0.49\linewidth}
\centering
\textbf{(b) RL algorithm.}\\[-0.2ex]
\begin{tabular*}{\linewidth}{@{\extracolsep{\fill}}lccc@{}}
\toprule
RL objective & Succ. (\%) & $\Delta$ vs. SFT & \shortstack{Act.(\%)} \\
\midrule
Standard GSPO & 60.47 & -2.28 & 40.0 \\
Step-balanced GSPO & \textbf{66.44} & \textbf{+3.69} & 26.1 \\
\bottomrule
\end{tabular*}
\end{minipage}
\captionsetup{skip=2pt, position=bottom,font=footnotesize}
\caption{Training-stage and RL algorithm ablations. \textit{Act.} reports the fraction of actions assigned to code execution.}
\label{tab:training_stage_ablation}
\vspace{0pt}
\end{table}

\textbf{Training-stage and algorithm ablation.}
Table~\ref{tab:training_stage_ablation} separates SFT, RL, and the RL objective. Panel (a) shows that SFT provides the main improvement over the unadapted Hybrid policy, RL alone changes little (+1.70 points), and applying RL after SFT yields the highest success rate of 66.44\%.
% 子表 (b) 隔离 RL 目标的作用：标准 GSPO 与 Step-balanced GSPO 使用相同的 SFT 初始化、训练数据、采样策略、奖励信号和优化预算，仅 token-level log-importance weight 到样本级 importance ratio 的聚合方式不同。标准 GSPO 相对于 SFT 初始化出现退化，而 Step-balanced GSPO 在相同起点上取得提升。这一反转直接表明，按 token 数量加权的聚合会扭曲异构 GUI/code 动作空间中的优化，而按动作步骤平衡贡献是将同一环境反馈稳定转化为策略提升的关键。
Panel (b) isolates the RL objective under the same SFT initialization, data, sampling strategy, reward signal, and optimization budget. Standard GSPO lowers success to 60.47\% and increases the code-action rate from 27.9\% under the SFT policy to 40.0\%, whereas Step-balanced GSPO raises success while maintaining a 26.1\% code-action rate close to the SFT policy. 
Together, these results suggest that action-step balancing improves evaluated performance while keeping the GUI/code action balance close to the SFT policy rather than drifting toward code-heavy behavior.

% 统一 GUI/code 范式消融。\ours{} 的核心主张要求区分代码工具是否可用与模型是否学会协调代码执行和 GUI 交互。图~\ref{fig:paradigm_ablation} 提供两组正交控制。子图 (a) 在开启和关闭代码动作的条件下评测同一个 \ours{27B} checkpoint，从而隔离工具的执行价值，并检验混合训练是否保留 GUI 能力。子图 (b) 固定主干、SFT 配置、10K 样本预算与 Hybrid 测试框架，仅改变 SFT 轨迹是 GUI-only 还是 GUI/code 混合；“Code act.” 表示分配给代码执行的工具调用比例。
\begin{table}[H]
\vspace{-0.4\baselineskip}
\centering
\footnotesize
\setlength{\tabcolsep}{1.5pt}
\renewcommand{\arraystretch}{0.9}
\begin{minipage}{0.49\linewidth}
\centering
\textbf{(a) Code access at inference.}\\[-0.2ex]
\begin{tabular*}{\linewidth}{@{\extracolsep{\fill}}lccc@{}}
\toprule
Policy & Succ. (\%) & \shortstack{Acc.} & \shortstack{Act.(\%)} \\
\midrule
Qwen3.5-27B & 42.59 & No & -- \\
\ours{27B} & 52.13 & No & -- \\
\ours{27B} & 62.75 & Yes & 27.9 \\
\bottomrule
\end{tabular*}
\end{minipage}%
\hfill%
\begin{minipage}{0.49\linewidth}
\centering
\textbf{(b) SFT modality at fixed scale.}\\[-0.2ex]
\begin{tabular*}{\linewidth}{@{\extracolsep{\fill}}lcccc@{}}
\toprule
SFT data & Samp. & Succ. (\%) & \shortstack{Acc.} & \shortstack{Act.(\%)} \\
\midrule
GUI only & 10K & 57.56 & Yes & 11.7 \\
Hybrid GUI/code & 10K & 60.61 & Yes & 37.1 \\
\bottomrule
\end{tabular*}
\end{minipage}
\captionsetup{skip=2pt, position=bottom,font=footnotesize}
\caption{Unified GUI/Code Paradigm ablations. \textit{Acc.}\ denotes whether code execution access is enabled at inference; \textit{Act.}\ reports the fraction of actions assigned to code execution.}
% ``Code Act.'' reports the fraction of tool calls assigned to code execution.
\label{tab:paradigm_ablation}
\vspace{-0.8\baselineskip}
\end{table}

\textbf{Unified GUI/code paradigm ablation.}
Table~\ref{tab:paradigm_ablation} disentangles code-tool availability from learned coordination. Panel (a) evaluates a fixed \ours{27B} checkpoint with and without code access. Panel (b) holds the backbone, SFT configuration, 10K-sample budget, and Hybrid evaluation harness fixed, changing only whether the SFT trajectories are GUI-only or hybrid GUI/code. 
Disabling code for the same policy still yields 52.13\% success; enabling it adds 10.62 points, demonstrating retained GUI competence and code complementarity. 
With inference-time code access and the 10K-sample budget fixed, hybrid SFT adds 3.05 points over GUI-only SFT, indicating that training on hybrid trajectories enables a more effective use of shared GUI/code actions rather than relying on GUI-only supervision to generalize to code actions at inference.

\section{Discussion}
\label{sec:discussion}
\begin{figure}[H]
\vspace{-0.4\baselineskip}
\centering
\includegraphics[width=0.98\linewidth]{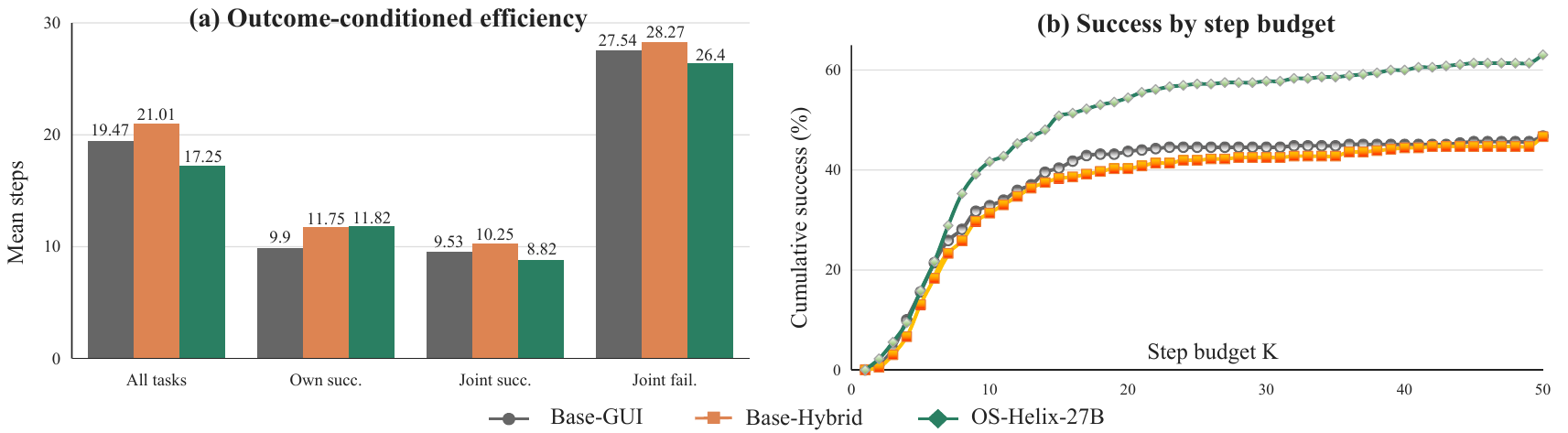}
\captionsetup{skip=3pt, position=bottom,font=footnotesize}
% 中文：OSWorld 上的执行效率。Base-GUI 和 Base-Hybrid 分别表示不提供和提供代码动作的基础模型。评估器得分大于 0.5 的任务记为成功，小于 0.5 的任务记为失败。子图 (a) 报告平均交互步数：All tasks 与 Own successes 分别基于每次运行自身的全部任务与成功任务；Joint successes 与 Joint failures 则仅保留三个策略具有相同二值结果的共享任务上。子图 (b) 报告累计 Success@$K$，并按各次运行实际完成的 OSWorld 任务归一化。
% Original caption: Execution efficiency on OSWorld. Base-GUI and Base-Hybrid denote the backbone evaluated without and with code access, respectively. Tasks with evaluator scores above 0.5 are counted as successes and those below 0.5 as failures. Panel (a) reports mean interaction steps: All tasks and Own successes use each run's full and successful task sets, respectively, whereas Joint successes and Joint failures restrict evaluation to shared tasks on which all three policies have the same binary outcome. Panel (b) reports cumulative Success@$K$, normalized by the OSWorld tasks completed in each run.
\caption{Execution efficiency on OSWorld.}
\label{fig:efficiency_analysis} 
\vspace{-1.0\baselineskip}
\end{figure}

% 准确率与效率。图~\ref{fig:efficiency_analysis} 中，Base-GUI 与 Base-Hybrid 分别表示在不提供和提供代码动作时评测的未经适配主干。评估器得分大于 0.5 的任务记为成功，小于 0.5 的任务记为失败。子图 (a) 报告平均交互步数：All tasks 与 Own successes 分别基于每次运行自身的全部任务与成功任务；Joint successes 与 Joint failures 则仅保留三个策略具有相同二值结果的共享任务，从而控制任务组成差异。子图 (b) 报告累计 Success@$K$，并按各次运行实际完成的 OSWorld 任务归一化。在整体、成功任务和结果匹配的子集上，\ours{27B} 均以更少交互完成任务，同时在较小步数预算下覆盖更多成功案例。这说明混合训练学到的模态路由同时改善执行效率与任务覆盖率，而仅在推理时开放代码工具并不足以产生同样的收益。
\textbf{Accuracy versus efficiency.}
% In Figure~\ref{fig:efficiency_analysis}, Base-GUI and Base-Hybrid denote the unadapted backbone evaluated without and with code actions, respectively. Evaluator scores above 0.5 are counted as successes and scores below 0.5 as failures. Panel (a) reports mean interaction steps: All tasks and Own successes use each run's complete and successful task sets, whereas Joint successes and Joint failures retain only shared tasks on which all three policies have the same binary outcome, controlling for task-composition differences. Panel (b) reports cumulative Success@$K$, normalized by the OSWorld tasks completed in each run. Across the full set, successful episodes, and outcome-matched subsets, \ours{27B} uses fewer interactions while covering more successes under small step budgets. Learned modality routing therefore improves both execution efficiency and task coverage, an effect not reproduced by enabling code only at inference time.
In Figure~\ref{fig:efficiency_analysis}, evaluator scores above 0.5 are counted as successes and scores below 0.5 as failures. Panel (a) reports mean interaction steps. All tasks and Own successes use each run's complete and successful task sets, respectively, whereas Joint successes and Joint failures retain only shared tasks on which all three policies have the same binary outcome, controlling for task-composition differences. Panel (b) reports cumulative Success@$K$ on OSWorld. Across the full set, successful episodes, and outcome-matched subsets, \ours{27B} uses fewer interactions while covering more successes under small step budgets. Learned modality routing therefore improves both execution efficiency and task coverage, an effect not reproduced by enabling code only at inference time.

% 步骤平衡的重要性。图~\ref{fig:training_behavior} 的子图 (a) 将每次在线 RL 更新中调用代码执行的动作比例作为行为诊断指标。标准 GSPO 在训练过程中逐渐提高代码动作占比，而 Step-balanced GSPO 在共同观测区间内保持更低且更稳定的水平。该趋势与方法部分提出的机制一致：按 token 聚合片段级比率会使较长的代码动作获得隐式高权重，而按动作步骤聚合能够缓解由此产生的模态偏移。结合表~\ref{tab:training_stage_ablation} 中控制其余配置后的性能反转，这一行为证据表明，步骤平衡既改善动作路由的稳定性，也使环境奖励更可靠地转化为任务表现。
% step balance 抑制 modality drift
% ↓
% 训练稳定 → 有效探索
% ↓
% pass@k gap 拉大 → 解空间扩展的直接证据
% Panel (a, bottom) further shows that this stabilized exploration also broadens the trajectory distribution: the gap between pass@3 and pass@1 widens from X.X to X.X percentage points after RL, whereas a purely confidence-sharpening update would leave this gap unchanged or narrowed. Step balancing is therefore doubly important—it suppresses modality drift during training and, as a consequence, enables the policy to discover genuinely diverse solution paths rather than committing more strongly to those already learned by SFT.
\begin{figure}[H]
  \vspace{-0.4\baselineskip}
  \centering
  \includegraphics[width=0.98\linewidth]{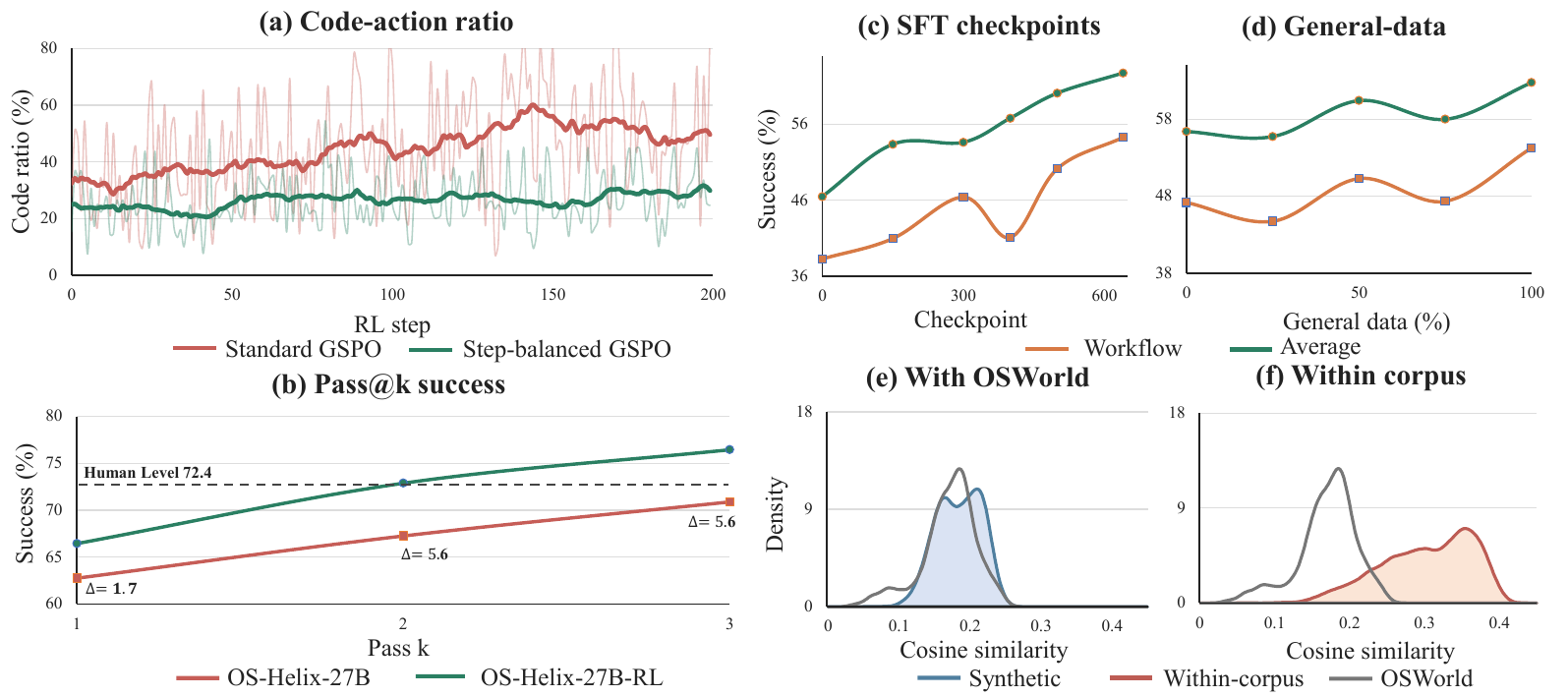}
  % \vspace{-0.6\baselineskip}
  \captionsetup{skip=5pt, position=bottom,font=footnotesize}
  % 中文：在线强化学习期间的代码动作占比。浅色曲线表示逐更新比例，粗色曲线表示 21 步中心移动平均。标准 GSPO 在训练中呈现明显的代码动作上升趋势，而 Step-balanced GSPO 在其记录区间内保持更低且更稳定的比例。
  % Original caption: Code-action ratios during online RL. Faint curves show per-update ratios and bold curves show 21-step centered moving averages. Standard GSPO develops a pronounced upward drift in code usage, whereas Step-balanced GSPO remains lower and more stable over its logged interval.
  % Original caption: Training behavior analysis. Panel (a) shows code-action ratios during online RL; panels (b)--(e) show scaling and instruction-similarity trends.
  \caption{Training dynamics, scaling, and coverage analysis.}
  % \label{fig:step_balance_code_ratio}
  \label{fig:training_behavior}
  \vspace{-1.0\baselineskip}
\end{figure}

\textbf{Why step balance matters.}
% Previous wording: Panel (a) of Figure~\ref{fig:training_behavior} uses the fraction of actions invoking code at each online RL update as a behavioral diagnostic. Standard GSPO progressively increases its reliance on code, whereas Step-balanced GSPO remains lower and more stable over the shared observed horizon. This trend agrees with the mechanism proposed in Section~\ref{subsec:training_paradigm}.
% Previous wording: \ybw{Add RL Pass@k analysis to show that step balance leads to more effective exploration and a broader solution space.}
% Previous wording: \ybw{Together with the controlled performance reversal in Table~\ref{tab:training_stage_ablation}, the behavioral evidence indicates that step balancing stabilizes routing and converts environment rewards into task performance more reliably.}
% 图~\ref{fig:training_behavior} 的面板 (a) 以每次在线 RL 更新中调用代码的动作占比作为行为诊断指标。标准 GSPO 呈现出逐步偏向代码密集型行为的漂移趋势，而 Step-balanced GSPO 在相同的观测窗口内始终保持较低且更为稳定的水平，与第~\ref{subsec:training_paradigm} 节中的动作步骤归一化设计相吻合。
Panel (a) of Figure~\ref{fig:training_behavior} uses the fraction of actions invoking code at each online RL update as a behavioral diagnostic. Under the same training setup, standard GSPO progressively drifts toward code-intensive behavior, whereas Step-balanced GSPO remains lower and more stable over the shared observed horizon.
% The theoretical result establishes only that step-balanced aggregation removes the deterministic token-count dependence of an action step's aggregate scalar coefficient; it does not in general equalize full parameter-gradient norms or variances across action lengths. The observed code-use contrast is therefore empirical evidence from this setting.
% Second, rule-based evaluators may underrepresent tasks whose success is only visually observable and could in principle bias the reward signal toward code actions, whereas it shows no such sustained drift under Step-balanced GSPO.
% 更重要的是，图（b）表明RL在重复采样下扩展了策略可达的解空间：与SFT策略相比，\ours{27B-RL}的Pass@1由62.75%提升至66.44%，Pass@3由70.88%提升至76.44%。Pass@3下更大的提升幅度表明，RL的作用不仅限于强化单条偏好轨迹，更在于提高了在多次独立尝试中发现不同成功轨迹的概率。
Panel (b) reports repeated-sampling coverage. Compared with the SFT policy, \ours{27B-RL} improves Pass@1 from 62.75\% to 66.44\% (+3.69 points) and Pass@3 from 70.88\% to 76.44\% (+5.56 points), with the latter supported by a paired-bootstrap confidence interval of 95\% [2.24,8.97]. 
The larger gain under Pass@3 indicates that RL does more than strengthen a single preferred trajectory and increases the probability of discovering distinct successful trajectories across independent attempts.
% 结合表~\ref{tab:training_stage_ablation} 中受控消融实验所呈现的性能反转现象，上述证据共同表明，Step-balanced RL 在稳定模态路由的同时，将环境反馈转化为更广泛、更有效的策略探索。
% Together with Table~\ref{tab:training_stage_ablation}, this evidence shows that step-balanced RL stabilizes modality routing while converting environmental feedback into broader, more effective exploration.

% 训练规模与数据分布分析。图~\ref{fig:training_behavior} 的子图 (b)--(e) 从性能扩展与指令分布两个角度进一步分析训练行为。子图 (b) 沿第一个 SFT epoch 的 checkpoint 报告 Workflow 域与整体平均成功率；子图 (c) 固定桌面数据，并将 40K 通用多模态数据池的使用比例从 0\% 提高至 100\%。两组曲线总体随监督规模增长，说明混合轨迹和通用数据分别支持桌面工作流学习与能力保持，但中间 checkpoint 的波动也表明收益并非简单的单调缩放。子图 (d)--(e) 仅使用连续空白经过归一化的指令文本：我们以 \texttt{text-embedding-3-large}~\citep{openai2024embeddingmodels} 编码指令并进行 L2 归一化，以向量点积计算余弦相似度。对每条合成指令，跨集合指标取其与所有 OSWorld 指令的平均相似度，内部指标取其与其余合成指令的平均相似度；OSWorld 内部分布以相同的留一平均方式计算，并作为参照。图中展示这些逐任务统计量的核密度估计。跨集合分布接近 OSWorld 内部参照，而合成语料内部呈现更高但仍较宽的相似度分布，说明数据共享可迁移的桌面任务结构，同时未收缩为单一的指令模板。
\textbf{Training scaling and data distributions.}
Panels (c)--(f) of Figure~\ref{fig:training_behavior} further examine the performance scaling and instruction distributions. Panel (c) reports success for the Work. category and the overall average across checkpoints from the first SFT epoch. Panel (d) holds the desktop data fixed and varies the fraction of the 40K general-purpose multimodal pool from 0\% to 100\%. 
Both sets of curves improve overall as the supervision grows, indicating complementary contributions from hybrid trajectories and general-purpose data. 
Panels (e)--(f) show kernel-density estimates of these per-task statistics. We encode each instruction with \texttt{text-embedding-3-large}~\citep{openai2024embeddingmodels}, L2-normalize the embedding, and compute cosine similarity by vector dot product. 
The cross-corpus similarity to OSWorld is comparable to OSWorld's internal-instruction similarity, indicating limited instruction-level overlap. 
Together with domain-removal ablation in Table~\ref{tab:data_composition_ablation}, this provides complementary evidence against direct instruction overlap or reliance on data from the evaluated domains.
Meanwhile, the synthesized corpus exhibits a higher but still broadly spread internal-similarity distribution, suggesting that it captures shared, transferable desktop-task structure rather than collapsing to a narrow instruction template family.

\section{Conclusion}
\label{sec:conclusion}
% 我们提出 \ours{}，一个通过可执行桌面合成、混合 SFT 与在线 RL 训练的原生混合计算机使用智能体，并提出 Step-balanced GSPO 以平衡异构 GUI/code 动作步骤。在 OSWorld 上，\ours{27B} 和 \ours{35B-A3B} 相较各自对应的 Qwen3.5 Hybrid baseline 分别提升 16.25 和 14.43 个绝对百分点，\ours{27B-RL} 达到 66.44\%；额外结果也为其向 WindowsAgentArena 的迁移提供了证据。
We present \ours{}, a native hybrid computer-using agent trained end-to-end through scalable executable desktop task synthesis, hybrid SFT, and Step-balanced GSPO online RL, designed to natively coordinate heterogeneous GUI and code actions within a single policy.
On OSWorld-Verified, \ours{27B} and \ours{35B-A3B} achieve relative improvements of 35.0\% and 33.7\% over their matched Qwen3.5 hybrid baselines, while \ours{27B-RL} further improves by 42.9\% to reach 66.44\% success. \ours{27B-RL} also transfers to WindowsAgentArena with 63.87\% success, surpassing GPT-5-based framework methods on both benchmarks.
% 消融实验与行为分析表明，SFT 贡献了主要性能增益；在固定数据规模下，混合监督优于仅 GUI 监督；相较于仅开放代码工具，学习到的模态路由同时改善了执行效率与任务覆盖率。在相同的 RL 设置下，Step-balanced GSPO 优于标准 GSPO，并缓解了其向代码密集型行为的偏移；同时，RL 将 Pass@3 从 70.88% 提升至 76.44%，表明模型在多次独立尝试中能够发现更广泛的成功轨迹。
Ablations and behavioral analyses show that SFT provides the main gain, hybrid supervision outperforms GUI-only supervision at a fixed scale, and learned routing improves both execution efficiency and task coverage beyond code access alone. 
Under matched RL settings, Step-balanced GSPO outperforms standard GSPO and mitigates its drift toward code-intensive behavior, while RL increases Pass@3 from 70.88\% to 76.44\%, indicating broader discovery of successful trajectories across independent attempts.
% 这些发现表明，原生 GUI/code 协同是一种可学习的混合范式，并为后续混合计算机使用智能体研究提供了可复现的基础。
These findings establish native GUI/code coordination as a learnable hybrid paradigm and provide a reproducible foundation for further research on hybrid computer-using agents.

\clearpage
\section*{AI Use Statement}
\label{sec:ai_use}

% 在本手稿的准备过程中，我们使用大语言模型辅助润色文字（包括语法和拼写修正）、起草部分章节内容，以及对附录~\ref{app:step_balanced_gspo} 中的公式进行初步推导。论文中的研究设计、实验、分析、论断与最终表述均由作者审核并承担责任。
In the preparation of this manuscript, we utilized LLMs to generate synthetic datasets, to refine the writing (including grammar and spelling corrections), to draft sections of the paper, and to assist with the preliminary derivation of mathematical formulas in Appendix~\ref{app:step_balanced_gspo}. The authors reviewed and take responsibility for the research design, experiments, analysis, claims, and final presentation.
\section*{Ethics Statement}
\label{sec:ethics}

% 本文不涉及人体受试者实验，也不收集真实用户的私有数据。我们的研究重点是通用计算机使用智能体的训练方法，所有轨迹采集、强化学习与评测均在受控且隔离的 Ubuntu 虚拟机中进行。虚拟机使用固定的应用版本和经过整理的本地文件库，并与宿主系统的个人文件及凭证隔离。这一设置降低了研究阶段的探索动作意外修改宿主状态或访问未授权资源的风险。
This work does not involve human-subject studies or the collection of private data from real users. We study training methods for general-purpose computer-using agents, and conduct all trajectory collection, reinforcement learning, and evaluation in controlled, isolated Ubuntu virtual machines. The VMs use fixed application versions and a curated local file library, and are separated from personal host files and credentials. This setup reduces the risk that exploratory actions during research inadvertently modify host state or access unauthorized resources.

% 尽管实验本身不存在显著的直接伦理风险，将具备 GUI 与代码混合执行能力的智能体部署到真实设备仍会带来重要的安全与隐私问题。此类智能体能够持续读取屏幕内容、修改本地文件并执行 Python 或 Bash，因此可能接触个人身份信息、机密文档与认证凭证，也可能受到界面内容或外部文件中提示注入的影响。代码动作进一步扩大了错误操作和恶意工作流的潜在影响范围。本文的隔离实验设置并不能解决这些真实部署风险。
Although the experiments themselves present no substantial direct ethical risk, deploying agents with combined GUI interaction and code execution on real devices raises important safety and privacy concerns. Such agents can continuously observe screen content, modify local files, and execute Python or Bash, and may therefore encounter personally identifiable information, confidential documents, or authentication credentials. They may also be exposed to prompt injection through interface content or external files. Code actions further increase the potential impact of erroneous behavior or malicious workflows. The isolated experimental setting used in this work does not resolve these deployment-time risks.

% 实际部署应遵循最小权限和安全内生设计原则，包括对应用、目录、网络与代码执行实施细粒度权限控制；在不可逆、涉及凭证或高影响的操作前要求人工确认；使用实时策略检查、沙箱执行、审计日志与执行后验证；并对截图、轨迹和工具输出进行数据最小化与敏感信息清理。用户应始终能够定义和撤销智能体的权限边界，银行、私人通信和凭证管理等敏感应用默认不应开放。由于通用自动化能力具有双重用途，未来的 CUA 研究应将能力提升与系统化红队测试、可解释的行动审计以及可靠的人工接管机制同步推进。
Real-world deployment should follow least-privilege and safety-by-design principles. These include granular permissions over applications, directories, networks, and code execution; explicit human confirmation before irreversible, credential-bearing, or otherwise high-impact operations; runtime policy checks, sandboxed execution, audit logs, and post-action verification; and data minimization and sanitization for screenshots, trajectories, and tool outputs. Users should retain the ability to define and revoke the agent's authority, with sensitive applications such as banking, private communication, and credential management inaccessible by default. 
Because general-purpose automation is inherently dual use, future CUA research should advance capabilities together with systematic red teaming, interpretable action auditing, and reliable human takeover mechanisms.

\bibliography{iclr2027_conference}
\bibliographystyle{iclr2027_conference}

\appendix
\ifoshelixopenversion
  % Public-version navigation page. Kept separate from the official ICLR style file.
\clearpage
\begingroup
\definecolor{appendixlinkblue}{RGB}{23,78,150}
\section*{Appendix Contents}
\vspace{0.5\baselineskip}
\noindent{\color{appendixlinkblue}\rule{\linewidth}{0.8pt}}\par
\vspace{1.6\baselineskip}

\newcommand{\appendixcontentsrow}[2]{%
  \noindent\hyperref[#1]{{\color{appendixlinkblue}\bfseries\ref*{#1}\hspace{1.2em}#2}}%
  \hfill\hyperref[#1]{{\color{appendixlinkblue}\bfseries\pageref*{#1}}}\par
  \vspace{1.6\baselineskip}%
}

\appendixcontentsrow{sec:limitations}{Limitations and Future Work}
\appendixcontentsrow{app:data_details}{Data Details}
\appendixcontentsrow{app:training_details}{Training Details}
\appendixcontentsrow{app:extra_experiments}{Extra Experiments}
\appendixcontentsrow{app:step_balanced_gspo}{Step-Balanced GSPO Details}
\appendixcontentsrow{app:case_study}{Case Study}
\appendixcontentsrow{app:prompts}{\ours{} Tool Schema and Prompts}
\endgroup

\fi
% Keep each major appendix self-contained and prevent its floats from crossing into the next one.
\clearpage

\ifoshelixopenversion\else
  
\fi

\section{Limitations and Future Work}
\label{sec:limitations}

% \ours{} 是迈向原生 GUI/代码混合计算机使用的初步探索，仍存在若干以计算资源与API资源为主的限制。
\ours{} is an initial step toward native GUI/code hybrid computer use, with remaining limitations centered primarily on computational and API resource constraints.

% 首先，桌面环境难以实现完备扩展：即使应用--产物桌面基础设施即使已经覆盖多样化应用，也无法涵盖真实用户设备、私有文档、应用版本和个性化配置的全部多样性。未来可进一步扩充应用与真实产物生态，并研究在保持可复现性的同时对系统状态和用户配置进行更广泛的随机化。
First, desktop environments are difficult to scale comprehensively: even an application--artifact desktop infrastructure spanning diverse applications cannot cover the full diversity of real user machines, private artifacts, application versions, and personalized configurations. Future work should further expand the application and artifact ecosystem and study broader randomization of system states and user configurations while preserving reproducibility.

% 第二，自动化验证仍不完善。规则评估器能够可靠覆盖许多以文件承载终态的任务，但需要经过谨慎设计，并可能依赖特定应用版本或导出格式。基于视觉语言模型的评估器能够覆盖视觉结果，却可能遗漏细粒度差异。未来工作应构建更强的评估器生成、修复与验证流水线，并系统研究多个评估器之间的一致性以及对奖励投机的抵抗能力。
% Second, automatic verification remains imperfect. Rule-based evaluators reliably cover many file-backed tasks, but require careful construction and may depend on particular application versions or export formats. VLM-based evaluators extend coverage to visual outcomes but may miss fine-grained discrepancies. Future work should develop stronger evaluator-generation, repair, and validation pipelines, and systematically study agreement across evaluators and resistance to reward hacking.
% 第二，自动化验证仍不完善。与代码任务不同，CUA自动化任务合成在有限资源条件下难以获取可靠的黄金终态，合成任务本身因此存在不可行风险，评估函数也随之具有内在脆弱性。具体而言，规则评估器能够可靠覆盖许多以文件承载终态的任务，但需要经过谨慎设计，且往往依赖特定应用版本或导出格式；更根本的问题在于，即便智能体的执行结果在语义上已大体正确，评估函数对文件属性的检查松紧不一，仍可能导致得分归零——部分工作（包括本文）已开始引入基于 rubric 的分阶段打分机制加以缓解，但尚不能从根本上解决这一问题。
% 基于视觉语言模型的评估器将覆盖范围延伸至文本输出与视觉结果，却可能遗漏细粒度差异；更重要的是，由于VLM评估器自身的脆弱性，现有RL工程实践通常将其排除在奖励信号之外，转而将构建重心集中于规则评估器——这在一定程度上限制了可纳入训练的任务类型与多样性。
% 未来工作应构建更强的评估器生成、修复与验证流水线，并系统研究多个评估器之间的一致性以及对奖励投机（reward hacking）的抵抗能力。
Second, automatic verification remains imperfect. Unlike software engineering tasks, where gold final states can be obtained deterministically, automated CUA task synthesis under limited resources cannot reliably capture ground-truth terminal states—tasks may therefore be subject to feasibility risks, and the resulting evaluators carry intrinsic brittleness. 
Concretely, rule-based evaluators reliably cover many file-backed tasks but require careful construction and often depend on specific application versions or export formats; more fundamentally, even when an agent's execution is semantically near-correct, inconsistent strictness in file-attribute checks can reduce the score to zero. Rubric-based partial scoring, adopted in several recent works including ours, partially alleviates this issue but does not resolve it entirely. 
VLM-based evaluators extend coverage to textual outputs and visual outcomes, yet may miss fine-grained discrepancies; more critically, due to their inherent brittleness, existing RL engineering practice typically excludes VLM-based scoring from the reward signal altogether, concentrating construction effort on rule-based evaluators—which in turn constrains the types and diversity of tasks that can be incorporated into training.
Future work should develop stronger evaluator-generation, repair, and validation pipelines, and systematically study cross-evaluator agreement as well as robustness to reward hacking.

% 第三，本文仅对稠密 27B 模型进行了 agentic RL。尽管 \ours{35B-A3B} 被纳入 SFT 实验，MoE 策略在环境交互式强化学习中会带来额外的服务、同步和优化困难。如何将 Step-balanced GSPO 稳定扩展到更大的 MoE 计算机使用智能体，是一个重要的后续方向。
% Third, we apply agentic RL only to the dense 27B model. Although \ours{35B-A3B} is included in the SFT experiments, MoE policies introduce additional serving, synchronization, and optimization challenges during environment-interactive RL. Stably scaling Step-balanced GSPO to larger MoE computer-using agents is an important direction for future work.
% 第三，受计算资源限制，本文的 Agentic RL 实验存在一定局限。我们仅对稠密 27B 模型进行了 RL 训练，在参数配置上较为保守，且受限于可用训练步数，尚无法在更大规模下系统探索 Agentic RL 的 scaling 行为。对于 \ours{35B-A3B}，尽管其已纳入 SFT 实验并取得良好效果，MoE 架构在 Agentic RL 中会引入额外的训推一致性与优化稳定性挑战，在现有资源条件下暂难可靠支撑这一训练流程。
% 我们认为，在更充裕的计算条件下深入探索 RL 的训练动态、scaling 规律，以及将 Step-balanced GSPO 稳定推广至更大规模 MoE 计算机使用智能体，是本工作最重要的后续方向之一。
Third, our agentic RL experiments are subject to certain limitations imposed by available computational resources. RL training was conducted only on the dense 27B model with relatively conservative hyperparameter settings; the total number of training steps was likewise constrained, precluding a systematic investigation of the scaling behavior of agentic RL. 
For \ours{35B-A3B}, despite its strong performance in SFT experiments, the Mixture of Experts~(MoE) architecture introduces non-trivial additional challenges in training-inference consistency and optimization stability during agentic RL, making it difficult to support this training pipeline reliably under current resource conditions. 
We regard a more thorough exploration of RL training dynamics and scaling laws, together with a stable extension of Step-balanced GSPO to larger MoE computer-using agents, as an important direction for future work.

% 此外，在CUA已具备一定能力的此刻，自进化(RSI)概念需要被抬到台面上来，如何利用现有CUA能力去自动化生成更复杂的任务，如何利用现有CUA能力去自动化生成更复杂的训练数据，并反哺给CUA继续训练提高能力，这些都是未来值得探索的方向。
% 除规模扩展之外，我们认为当前正是将自进化概念提上议程的关键节点。随着 CUA 模型在通用计算机使用能力上达到相当水平，模型自身即可充当自己的 teacher model：在现有任务上自主 rollout 交互轨迹，为下一轮训练提供数据来源，并通过这一自举循环持续提升能力。这一范式无需依赖人工示范或外部整理的训练数据，而是让智能体随着自身能力的增长不断生成质量更高的训练轨迹，并将其反哺至后续训练轮次。我们认为，这种由模型自身驱动的自进化训练范式，是推动 CUA 智能体长期演进最值得期待的方向之一。
Beyond scaling, we believe the community is now at an inflection point where recursive self-improvement~(RSI) deserves serious consideration. As CUA models attain a meaningful level of general computer-use capability, the model itself can serve as its own teacher: rolling out interaction trajectories on existing tasks, providing the training signal for the next iteration, and progressively improving through this self-bootstrapping loop. 
Rather than relying solely on human demonstration or externally curated data, such a paradigm allows the agent to continuously generate its own training trajectories with growing quality as capability improves and feed them back into subsequent training rounds. We view this self-evolving, teacher-free training paradigm as one of the most promising long-term directions for advancing CUA agents.

Finally, many existing desktop benchmarks were originally designed for GUI-only agents, with task design, reference solutions, and evaluation functions all implicitly centered on visual interaction paths. Although their tasks can often be completed with hybrid strategies, existing evaluators may not capture every valid GUI/code solution. Furthermore, using success rate as the sole metric is overly coarse: it neither distinguishes whether the agent selected the more efficient modality nor captures potential safety risks inherent to hybrid execution, such as unintended side effects of code execution or inadvertent system-level operations. 

% 值得关注的是，另一条颇具潜力的路径是将现有的通用智能体基准（如 GDPEval、Agent Last Exam 等）迁移至 GUI 环境中加以重构。这类基准本身面向真实文件操作与复杂推理任务而设计，天然涵盖了多样化的环境状态与真实工件，将其引入桌面 GUI 环境后，既可规避从零设计任务的高昂成本，又能在统一的评测框架下同时考察 GUI 操作、代码执行及两者的混合协同，为深入探索 GUI/Code 混合驱动范式提供更贴近真实使用场景的评测土壤。
A particularly promising complementary direction is the migration and adaptation of existing general-purpose agent benchmarks such as GDPval~\citep{patwardhan2026gdpval} and Agents' Last Exam~\citep{sun2026agents} into GUI environments. These benchmarks are inherently designed around authentic artifact manipulation and complex reasoning tasks, and thus naturally encompass diverse environment states and authentic artifacts. Porting them into desktop GUI settings not only amortizes the high cost of designing tasks from scratch, but also enables a unified evaluation framework that simultaneously exercises GUI interaction, code execution, and their hybrid coordination, providing a more ecologically valid testbed for deeply probing the GUI/code hybrid-driving paradigm.
Future benchmarks should explicitly support hybrid computer use from the task-design stage onward, providing multi-path reference solutions and jointly evaluating terminal-state correctness, execution efficiency, modality-selection appropriateness, and safety, thereby offering a more comprehensive and equitable evaluation framework for next-generation hybrid agents.

\section{Data Details}
\label{app:data_details}

\subsection{Task Synthesis Comparison}
% 表~\ref{tab:task_synthesis_comparison} 从任务扎根、应用扩展、真实文件集成、多应用任务支持和可执行评估五个维度，对近期 CUA 任务与环境合成流水线进行比较。该比较以合成覆盖维度为轴，而非衡量数据规模或成熟度。\ours{} 的核心区别在于其统一框架的覆盖广度：流水线在单一体系内同时处理 K/E/C 扎根、应用扩展、基于真实文件的初始状态构建，以及可执行的规则化评估。具体而言，应用—产物兼容图对应用配置与文件类型的有效配对进行组织，从而支持可复现初始状态的系统化实例化；评估器流水线验证持久化的终态结果而非约束动作序列，使其能够跨应用领域支持多样化的任务目标与评估标准。
Table~\ref{tab:task_synthesis_comparison} compares recent CUA task- and environment-synthesis pipelines across five dimensions: task grounding, application scaling, authentic artifact integration, cross-application task support, and executable evaluation. The comparison is organized along the axis of synthesis coverage rather than data scale or maturity. The broader contribution of \ours{} is to extend exploration-grounded task generation toward artifact-grounded executable desktop-state construction: the pipeline jointly addresses K/E/C grounding, application scaling, authentic artifact-backed state initialization, and executable rule-based evaluation within a single coherent framework. Concretely, an application--artifact compatibility graph organizes valid pairings of application configurations and supported artifact types or file formats, enabling reproducible initial states to be instantiated systematically. The evaluator pipeline verifies persistent terminal outcomes rather than constraining the action sequence, supporting diverse task objectives and evaluation criteria across application domains and providing the infrastructure for subsequent hybrid-policy training.

\begin{table}[t]
  \centering
  \scriptsize
  \setlength{\tabcolsep}{2.5pt}
  \renewcommand{\arraystretch}{1.10}
  \newcommand{\cmark}{\textcolor{gainGreen}{\textbf{$\checkmark$}}}
  \newcommand{\xmark}{\textcolor{lossRed}{\textbf{$\times$}}}
  \newcommand{\pmark}{%
    \textcolor{orange}{\textbf{$\checkmark$}}
  }
  \begin{tabularx}{\linewidth}{@{}l *{5}{>{\centering\arraybackslash}X}@{}}
    \toprule
    \textbf{Method} &
    \shortstack{\textbf{Synthesis}\\\textbf{Grounding}} &
    \shortstack{\textbf{Application}\\\textbf{Scaling}} &
    \shortstack{\textbf{Real Artifact}\\\textbf{Scaling}} &
    \shortstack{\textbf{Cross-App}\\\textbf{Tasks}} &
    \shortstack{\textbf{Rule-Based}\\\textbf{Evaluator}} \\
    % Hidden column: \shortstack{\textbf{Golden}\\\textbf{State}}
    \midrule
    OS-Genesis~\citeyearpar{OS-Genesis}              & E     & \xmark     & \xmark     & \xmark     & \xmark     \\ % Golden State: \xmark
    AUTO PLAY~\citeyearpar{ramrakhya2026scaling}     & K/E/C & \xmark     & \xmark     & \xmark     & \xmark     \\ % Golden State: \xmark
    TreeCUA~\citeyearpar{TreeCUA}                    & K/E/C & \xmark     & \xmark     & \xmark     & \xmark     \\ % Golden State: \xmark
    EvoCUA-1.5~\citeyearpar{huang2026evocua}         & K     & \xmark     & \cmark & \xmark     & \cmark \\ % Golden State: \xmark
    Gym-Anything~\citeyearpar{aggarwal2026gymanything} & K     & \cmark & \cmark & \xmark     & \xmark     \\ % Golden State: \xmark
    CUA-Gym~\citeyearpar{CUA-Gym}                    & K/C   & \cmark & \xmark     & \cmark & \cmark \\ % Golden State: \cmark
    OpenComputer~\citeyearpar{OpenComputer}          & K/C   & \cmark & \xmark     & \xmark     & \cmark \\ % Golden State: \xmark
    ScaleCUA~\citeyearpar{lv2026scalecua}            & K     & \xmark     & \xmark     & \xmark     & \cmark \\ % Golden State: \xmark
    Qwen-CUA~\citeyearpar{lu2026qwencua}             & K     & \cmark & \cmark & \cmark & \cmark \\ % Golden State: \xmark
    Qwen-UI-Agent~\citeyearpar{zhou2026qwenuiagent}  & K/C   & \xmark     & \xmark     & \cmark & \cmark \\ % Golden State: \xmark
    \midrule
    \ours{}                                           & K/E/C & \cmark & \cmark & \pmark & \cmark \\ % Golden State: \xmark
    \bottomrule
  \end{tabularx}
  \captionsetup{skip=3pt,position=bottom}
  % 中文翻译：近期 CUA 任务与环境合成流水线的比较。合成扎根区分外部知识或应用先验（K）、在最终任务指令提出前执行的任务特定环境或文件检查（E），以及跨实例覆盖或能力记忆（C）。仅用于验证的生成后执行不计为 E。应用扩展要求在继承的基准之外添加并配置应用环境。“真实文件”要求来源明确将文件支持的输入描述为真实世界或具有真实性；明确描述为合成的产物不计入。基于规则的评估器必须以程序方式检查终态；仅使用 VLM 的判断器不计入。绿色对勾表示明确支持，红色叉号表示按照上述定义不存在支持或文档未明确说明，黄色对勾表示已有初步原型但尚未形成大规模、系统化的支持。\(^{\dagger}\) 覆盖记忆目前用于我们的跨应用合成原型。我们的方法使用灰色底纹标出。
  % Original caption: Comparison of recent CUA task- and environment-synthesis pipelines. Synthesis grounding distinguishes external knowledge or application priors (K), task-specific environment or file inspection performed before the final task instruction is proposed (E), and cross-instance coverage or capability memory (C). Post-generation execution used only for validation does not qualify as E. Application scaling requires adding and configuring application environments beyond an inherited benchmark. The real-files criterion includes file-backed inputs explicitly presented as real-world or realistic by the source; artifacts explicitly described as synthetic do not qualify. Rule-based evaluators must programmatically inspect the terminal state; VLM-only judges do not qualify. $\checkmark$ denotes explicit support and $\times$ denotes that support is absent or not explicitly documented under these definitions. $^{\dagger}$ The coverage memory is currently used in our multi-application synthesis prototype. Our method is shaded.
  \caption{Comparison of recent CUA task- and environment-synthesis pipelines. Synthesis grounding distinguishes external knowledge or application priors (K), task-specific environment or file inspection before the final instruction is proposed (E), and cross-instance coverage or capability memory (C). Application scaling denotes support for adding and configuring applications beyond an inherited benchmark; real artifact scaling denotes authentic or realistic file-backed inputs; rule-based evaluators programmatically inspect terminal states. A green check denotes explicit support, a red cross denotes no explicit support, and an orange check denotes preliminary but not yet systematic support. 
  }
  \label{tab:task_synthesis_comparison}
  \vspace{-0.3cm}
\end{table}

% 跨应用任务合成目前仍处于雏形阶段，这一局限性在表中以半勾符号标注；将该组件扩展为规模更大、覆盖更系统的任务集合是后续工作的明确方向。Qwen-CUA 和 Qwen-UI-Agent 等工作已在大规模长程及跨应用工作流合成方面展示了更充分的覆盖，这些维度代表了 \ours{} 未来扩展的重要空间。
Cross-application task synthesis is currently realized as a limited prototype, reflected by the partial symbol in the table; extending this component into a larger and more systematically characterized collection remains an explicit direction for future work. Works such as Qwen-CUA and Qwen-UI-Agent have demonstrated more extensive coverage of long-horizon and cross-application workflows at scale, and those dimensions represent clear avenues for subsequent expansion of \ours{}.

\subsection{Applications and Artifacts Details}
\label{app:file_categories}

% 我们的完整桌面应用清单共包含 45 个应用，其中 44 个为覆盖多种功能类别的具名桌面应用，另有一个专用于系统级操作的 \texttt{os} 应用。表~\ref{tab:software_categories} 按功能类别逐一列出全部应用，表~\ref{tab:file_types} 列出流水线所覆盖的文件与产物类型，表~\ref{tab:file_library_sources} 则报告常用产物池的主要来源。
Our complete desktop application inventory comprises 45 applications, consisting of 44 named desktop applications spanning diverse functional categories and one dedicated \texttt{OS} application for system-level operations. Table~\ref{tab:software_categories} enumerates all applications organized by functional category, Table~\ref{tab:file_types} lists the artifact types and file formats covered by the pipeline, and Table~\ref{tab:file_library_sources} reports the primary sources of commonly used artifact pools.

% 表~\ref{tab:file_library_sources} 聚焦于应用—产物采样器所使用的主要产物池。表中未单独列出的长尾产物类型通过轻量化替代方案加以覆盖——包括小规模精选示例库、应用自带的演示资产、简单格式转换及人工构建的种子文件——而非专门开展大规模采集，这一处理方式反映了人力资源方面的实际约束。
Table~\ref{tab:file_library_sources} focuses on the major artifact pools consumed by the application--artifact sampler. Long-tail artifact types not individually itemized in the table are handled through lightweight alternatives including small curated example libraries, application-bundled demo assets, simple format conversions, and manually constructed seed files rather than dedicated large-scale collection efforts, reflecting practical constraints on human resource availability.

\begin{table}[t]
\centering
\scriptsize
\setlength{\tabcolsep}{4pt}
\renewcommand{\arraystretch}{1.16}
\rowcolors{2}{gray!4}{white}
\begin{tabular}{>{\raggedright\arraybackslash}p{0.25\linewidth}>{\centering\arraybackslash}p{0.07\linewidth}>{\raggedright\arraybackslash}p{0.61\linewidth}}
\toprule
\textbf{Category} & \textbf{Count} & \textbf{Applications} \\
\midrule
Web, communication, and knowledge work & 5 & \texttt{Chrome}, \texttt{ThunderBird}, \texttt{Notion}, \texttt{Zotero}, \texttt{Foliate} \\
Office, documents, and publishing & 9 & \texttt{LibreOffice Writer}, \texttt{LibreOffice Calc}, \texttt{LibreOffice Impress}, \texttt{WPS Presentation}, \texttt{WPS Writer}, \texttt{WPS Spreadsheets}, \texttt{PDF Arranger}, \texttt{TeXstudio}, \texttt{Text Editor} \\
Programming, databases, and version control & 5 & \texttt{PyCharm}, \texttt{VSCode}, \texttt{DBeaver}, \texttt{GitKraken}, \texttt{Meld} \\
Graphics, diagrams, CAD, and photography & 6 & \texttt{Gimp}, \texttt{Inkscape}, \texttt{Krita}, \texttt{Darktable}, \texttt{Draw.io}, \texttt{LibreCAD} \\
Audio, video, music, and 3D media & 9 & \texttt{Blender}, \texttt{Audacity}, \texttt{Handbrake}, \texttt{Mixxx}, \texttt{OBS Studio}, \texttt{MuseScore}, \texttt{Spotify}, \texttt{Shotcut}, \texttt{VLC} \\
Scientific, engineering, geospatial, and network tools & 8 & \texttt{Scilab}, \texttt{Wireshark}, \texttt{KiCad}, \texttt{Grass}, \texttt{ChimeraX}, \texttt{KAlgebra}, \texttt{Celestia}, \texttt{Google Earth} \\
Personal organization and finance & 2 & \texttt{Calendar}, \texttt{HomeBank} \\
Operating-system operations & 1 & \texttt{OS} \\
\midrule
% \rowcolor{gray!14}
\textbf{Total} & \textbf{45} & \textbf{45 applications} \\
\bottomrule
\end{tabular}
\rowcolors{2}{}{}
\captionsetup{skip=3pt,position=bottom}
\caption{Functional classification of all 45 applications in the Ubuntu desktop infrastructure.}
\label{tab:software_categories}
\end{table}

\begin{table}[t]
\centering
\scriptsize
\setlength{\tabcolsep}{4pt}
\renewcommand{\arraystretch}{1.16}
\rowcolors{2}{gray!4}{white}
\begin{tabular}{>{\raggedright\arraybackslash}p{0.35\linewidth}>{\raggedright\arraybackslash}p{0.60\linewidth}}
\toprule
% \rowcolor{gray!14}
\textbf{File group} & \textbf{Types} \\
\midrule
Documents, publishing, and office files & \texttt{.doc}, \texttt{.docx}, \texttt{.pptx}, \texttt{.xlsx}, \texttt{.pdf}, \texttt{.epub}, \texttt{.txt}, \texttt{.tex} \\
Code, projects, and structured data & \texttt{Project Folders}, \texttt{.py}, \texttt{.sql}, \texttt{.csv}, \texttt{.sci}, \texttt{.xhb} \\
Web, geospatial, and network artifacts & \texttt{url}, \texttt{.html}, \texttt{.kmz}, \texttt{.pcapng} \\
Images, vectors, diagrams, and design files & \texttt{.png}, \texttt{.jpeg}, \texttt{.svg}, \texttt{.drawio}, \texttt{.kra} \\
Audio, video, and music files & \texttt{.mp3}, \texttt{.wav}, \texttt{.mp4}, \texttt{.mscz} \\
CAD and 3D assets & \texttt{.blend}, \texttt{.dxf} \\
\bottomrule
\end{tabular}
\rowcolors{2}{}{}
\captionsetup{skip=3pt,position=bottom}
\caption{File and artifact types covered by the complete inventory of 45 applications.}
\label{tab:file_types}
\end{table}

\begin{table}[t]
\centering
\scriptsize
\setlength{\tabcolsep}{4pt}
\renewcommand{\arraystretch}{1.14}
\rowcolors{2}{gray!4}{white}
\begin{tabular}{@{}>{\raggedright\arraybackslash}p{0.15\linewidth}>{\raggedright\arraybackslash}p{0.15\linewidth}>{\raggedright\arraybackslash}p{0.62\linewidth}@{}}
\toprule
% \rowcolor{gray!14}
\textbf{Type} & \textbf{High-level group} & \textbf{Primary source} \\
\midrule
% \texttt{.docx} & Office documents & xlang-ai/OSWorld; superdoc-dev/docx-corpus \\
% \texttt{.pptx} & Office documents & xlang-ai/OSWorld; ForceLess/PPTAgent\_parsed\_data \\
% \texttt{.xlsx/.csv} & Office documents & xlang-ai/OSWorld; Multilingual-Multimodal-NLP/TableBench \\
% \texttt{.pdf} & Documents & xlang-ai/OSWorld; lmms-lab/DocVQA converted to PDF format \\
% \texttt{.png/.jpeg} & Images & xlang-ai/OSWorld; moonshotai/WorldVQA \\
% \texttt{.mp3} & Audio & xlang-ai/OSWorld; thang101020/musical\_audio\_cache\\
% \texttt{.mp4} & Video & xlang-ai/OSWorld; raushan-testing-hf/videos-test; 2e8konjak/youtube\_videos\_12\\
\texttt{.docx} & Office documents & superdoc-dev/docx-corpus \\
\texttt{.pptx} & Office documents & ForceLess/PPTAgent\_parsed\_data \\
\texttt{.xlsx/.csv} & Office documents & Multilingual-Multimodal-NLP/TableBench \\
\texttt{.pdf} & Documents & lmms-lab/DocVQA converted to PDF format \\
\texttt{.png/.jpeg} & Images & moonshotai/WorldVQA \\
\texttt{.mp3} & Audio & thang101020/musical\_audio\_cache\\
\texttt{.mp4} & Video & raushan-testing-hf/videos-test; 2e8konjak/youtube\_videos\_12\\
\texttt{.wav} & Audio & Sakshi113/MMAU \\
\texttt{.tex} & \LaTeX{} source & TIGER-Lab/arxiv-latex-5T \\
\texttt{.svg} & Vector graphics & lucide-icons/lucide \\
\texttt{.blend} & 3D assets & Poly Haven\textsuperscript{1}; DietCoke4671/BlenderBench \\
\texttt{.dxf} & CAD files & LibreCAD examples; tobygerlach/dxf\_dataset \\
\texttt{.py} & Code files & Common algorithmic problems collected from programming problem sets \\
\texttt{.txt} & Plain text & Plain-text files converted from other file types \\
\texttt{url} & Web artifacts & OpenGVLab/ScaleCUA Web URLs\\ 
\texttt{Project folder} & Project files & SWE-bench/SWE-bench \\
\bottomrule
\end{tabular}
\rowcolors{2}{}{}
\vspace{1pt}\\[-0.4ex]
\parbox{\linewidth}{\scriptsize \textsuperscript{1}\url{https://polyhaven.com/}}
\captionsetup{skip=3pt,position=bottom}
\caption{Primary sources of the file library by file type. Public upstream sources are written in repository-style form, including GitHub and dataset repositories where applicable. When an upstream source contains more files than needed for local deployment, we randomly sample a subset from that source. File types outside this table are covered through lightweight collection from small example libraries, application demo assets, simple conversions, or manually constructed seed files rather than large-scale harvesting.}
\label{tab:file_library_sources}
\end{table}

\subsection{Final State Materialization}
\label{app:terminal_state_materialization}

% 我们的 \emph{万物皆文件} 抽象将文件系统视为通用的检查接口，而非对任务状态来源的限制。终态条件可能初始驻留于已下载的网页、内存中的应用状态、未保存的布局或持久化的用户产物。只要能够保留评估所需的语义，我们便尝试将相关条件物化为虚拟机内的文件表示：网页状态可被下载或本地序列化，内存中的结果可被导出，答案可被写入结构化或纯文本文件。这一外化过程将状态的存储位置与其可验证表示相解耦，从而使异构的桌面任务目标共享统一的评估接口。
Our \emph{everything-is-a-file} abstraction treats the file system as a common inspection interface rather than as a restriction on where a task state must originate. A terminal condition may initially reside in a downloaded web page, an in-memory application state, an unsaved layout, or a persistent user artifact. We attempt to materialize the relevant condition into a VM-resident file representation whenever this can preserve the semantics needed for evaluation: a web state may be downloaded or serialized locally, an in-memory result may be exported, and an answer may be written to a structured or plain-text file. This externalization separates the location of a state from its verifiable representation and allows heterogeneous desktop objectives to share a common evaluation interface.

% 物化步骤并非在所有情况下均无损。部分条件——如精确的文档结构、元数据或序列化的应用状态——支持忠实的程序化检查；另一些条件则依赖于渲染后的布局、画布外观、瞬态 UI 状态，或序列化无法保留相关证据的内容。因此，评估器的选择反映了一种可观测性权衡：当确定性的文件或命令表示足以捕获所需语义时，采用基于规则的检查；当视觉或其他不可序列化的证据不可或缺时，采用基于 VLM 的检查；当二者能提供互补证据时，则采用混合评估器。最终得到的评估器对智能体是通过 GUI 操作、代码执行还是二者结合来达到终态条件保持不可知。
The materialization step is not uniformly lossless. Some conditions, such as exact document structure, metadata, or serialized application state, admit faithful programmatic inspection; others depend on rendered layout, canvas appearance, transient UI state, or content for which serialization does not preserve the relevant evidence. Evaluator selection therefore reflects an observability trade-off: rule-based checks are used when a deterministic file or command representation captures the required semantics, VLM-based checks are used when visual or otherwise non-serializable evidence is essential, and hybrid evaluators combine the two when they provide complementary evidence. 
The resulting evaluator is agnostic to whether the agent reached the terminal condition through GUI actions, code execution, or their combination.

\subsection{Instruction and Evaluator Synthesis Details}
\label{app:synthesis_pipeline}

% 主要任务合成流水线通过应用—产物兼容图实例化环境状态。流水线随后遵循单向信息流：应用与产物配置决定初始化上下文；\textbf{探索智能体}生成结构化的指令提案与文本形式的成功条件；独立的\textbf{评估器智能体}将冻结后的条件编译为基于规则或视觉的检查项；每对经独立验证的指令—评估器组合随即进入任务池。
The primary task-synthesis pipeline instantiates states through the application--artifact compatibility graph. The pipeline then follows a one-way information flow: application and artifact configurations determine the initialization context; the \textbf{Exploration Agent} produces a structured instruction proposal and textual success conditions; the independent \textbf{Evaluator Agent} compiles the frozen conditions into rule-based or visual checks; and each independently validated instruction-evaluator pair enters the task pool.

\paragraph{Application-centered initialization.}
% 我们为每个应用记录规范名称、版本、可处理的文件类型、启动命令和初始化模板，并据此构建应用--产物兼容图。每轮合成首先恢复受控虚拟机快照并采样一个主要应用，再根据应用初始化模板通过兼容图采样零个或若干兼容文件。对于每个被采样的已有产物，初始化程序会在任务执行前创建一份位于虚拟机内、仅供评估器访问的黄金副本，任务本身仍以原路径为操作目标；该副本与任务配置绑定，可在后续内容或结构比较中作为期望参考。系统随后将应用元数据、准确文件路径、启动配置、应用教程、全局应用记忆和初始截图提供给探索智能体。
For each application, we record a canonical name, version, supported artifact types and file formats, launch commands, and initialization templates, and use these fields to build the application--artifact compatibility graph. Each synthesis episode first restores a controlled VM snapshot and samples a primary application, then selects zero or more compatible artifacts through the graph according to the application's initialization templates.
% a small number of natural cross-application tasks additionally include a predefined suite companion or \texttt{os} operation. 
For every sampled pre-existing artifact, initialization creates a VM-resident, evaluator-only golden copy before task execution while leaving the original path as the active task target; this copy is bound to the task configuration and may serve as the expected reference in subsequent content or structural comparisons. The system then provides the Exploration Agent with application metadata, exact file paths, launch configuration, an application tutorial, global application memory, and an initial screenshot.

% 下面给出 PyCharm 的部分应用配置，用于说明应用元数据如何连接应用初始化与文件库采样。
The following partial PyCharm configuration illustrates how application metadata connects application initialization with artifact-library sampling.

% \texttt{version} 固定任务合成、轨迹采集与评估器实现所依据的应用版本；\texttt{type} 声明该应用能够直接接收的文件或项目类型。\texttt{commands} 给出两个可选初始化模板：第一个仅启动空白应用，第二个启动应用并打开一个采样产物。\texttt{random} 与这些模板按位置一一对应，并将其归一化为初始化采样概率；因此 \texttt{[10, 90]} 表示约有 10\% 的初始化采用空白 PyCharm，约有 90\% 采用带文件或项目的启动方式。选择第二个模板时，系统会从文件库中随机采样一个与 \texttt{type} 兼容的绝对路径，并在执行命令前用该路径替换 \texttt{PATH}。\texttt{related\_app} 与 \texttt{related\_type} 保存应用及文件类型之间的兼容关系，可用于选择常规关联应用以及后续的应用图原型。主要的应用--工件采样过程由当前应用的 \texttt{type}、\texttt{commands} 和 \texttt{random} 字段决定。
The \texttt{version} field fixes the application version assumed by task synthesis, trajectory collection, and evaluator implementation, while \texttt{type} declares the artifact or project types that the application can directly consume. The two entries in \texttt{commands} are alternative initialization templates: the first launches an empty application, and the second launches the application with a sampled artifact. The position-aligned values in \texttt{random} are normalized into initialization probabilities; hence \texttt{[10, 90]} assigns approximately 10\% probability to an empty PyCharm launch and 90\% to a file- or project-backed launch. When the second template is selected, the system randomly samples an absolute file path whose type is compatible with \texttt{type} from the artifact library and substitutes it for \texttt{PATH} before executing the command. The \texttt{related\_app} and \texttt{related\_type} fields retain compatibility relations among applications and artifact types for selecting conventional companion applications and for the graph-guided cross-application prototype. 
% The primary application--artifact sampler is driven by the current application's \texttt{type}, \texttt{commands}, and \texttt{random} fields.

% 这种声明式配置使应用程序的扩展具有可复现性，而非依赖一堆临时拼凑的启动脚本。添加一个新应用只需声明其版本、启动流程、所支持的制品或项目类型、初始化概率，以及与伴生应用的可选关联关系。兼容性图随后会在情节实例化到固定虚拟机之前，过滤出合法的应用-制品配对组合，从而使新应用与新制品类型能够持续扩充可执行状态空间，同时保留初始化与采样的语义一致性。
This declarative configuration is what makes application scaling reproducible rather than a collection of ad hoc launch scripts. Adding an application requires specifying its version, launch procedures, supported artifact types, initialization probabilities, and optional relations to companion applications. The compatibility graph then filters valid application--artifact pairings before an episode is instantiated in a fixed VM, allowing new applications and artifact types to expand the executable state space while preserving initialization and sampling semantics.

\begin{tcblisting}{title={Example Application Configuration: \texttt{PyCharm}}, colback=black!2, colframe=black!55, enhanced, boxrule=0.5pt, boxsep=1pt, left=2pt, right=2pt, top=2pt, bottom=2pt, listing only, listing options={basicstyle=\ttfamily\scriptsize, breaklines=true, breakatwhitespace=false, columns=fullflexible}}
  "pycharm": {
    "version": "PyCharm 2025.3.3",
    "type": [
      "py",
      "project_folder"
    ],
    "commands": [
      [
        [["pycharm-community"]]
      ],
      [
        [["pycharm-community", "PATH"]]
      ]
    ],
    "random": [10, 90],
    "related_app": [
      "vscode",
      "gitkraken"
    ],
    "related_type": [
      "py",
      "project_folder"
    ]
  }
  \end{tcblisting}
  
\paragraph{Grounded instruction proposal.}
% 为避免重复覆盖已熟悉功能，我们维护全局应用记忆，记录每个应用已被合成任务覆盖的功能，并在后续探索中提供给探索智能体。在需要额外界面信息时，探索智能体可以执行一段简短的非破坏性查看，其动作限制为 \texttt{open}、\texttt{click}、\texttt{scroll} 和 \texttt{done}。该阶段不提供键盘或自由文本输入，以避免破坏已知初始状态。探索智能体随后提出目标导向的指令、预估步数和评估器应检查的细粒度成功条件。由此产生的指令记录包含 \texttt{snapshot}、带版本号的 \texttt{related\_apps}、\texttt{instruction}、\texttt{config}、\texttt{complexity}、\texttt{category}、\texttt{success\_criteria} 与 \texttt{evaluation\_requirements\_text} 等字段。\texttt{instruction} 是面向用户的核心任务描述。\texttt{snapshot}、\texttt{related\_apps} 和 \texttt{config} 在初始化阶段确定，用于记录可复现的环境上下文。\texttt{category} 由探索智能体随指令一起生成，并分为三类。\texttt{app\_only} 指主要围绕应用或系统功能的任务，例如改变软件或用户配置。\texttt{file\_only} 指主要利用软件处理输入文件且最终产物为文件的任务。\texttt{mixed} 指二者兼有。\texttt{complexity} 从纯 GUI 操作角度估计完成任务所需步骤数。5--15 步标为 \texttt{easy}，15--40 步标为 \texttt{medium}，40 步以上标为 \texttt{complex}。
To avoid repeatedly covering familiar functionality, we maintain global application memory that records functions previously exercised for each application and expose it during subsequent exploration. When additional interface information is needed, the Exploration Agent may perform a short non-destructive inspection using only \texttt{open}, \texttt{click}, \texttt{scroll}, and \texttt{done}. Keyboard and free-form text input are unavailable to avoid mutating the known initial state. It then proposes a goal-oriented instruction and fine-grained success conditions for evaluator synthesis.
The resulting instruction record contains fields such as \texttt{snapshot}, versioned \texttt{related\_apps}, \texttt{instruction}, \texttt{config}, \texttt{complexity}, \texttt{category}, \texttt{success\_criteria}, and \texttt{evaluation\_requirements\_text}. The \texttt{instruction} field is the core user-facing task specification. The \texttt{snapshot}, \texttt{related\_apps}, and \texttt{config} fields are determined during initialization and record the reproducible environment context. 
The \texttt{category} field is generated by the Exploration Agent together with the instruction and is divided into three groups. \texttt{app\_only} denotes tasks centered on application or system functionality, such as changing software or user settings. \texttt{file\_only} denotes tasks centered on processing input files whose final deliverables are file artifacts. \texttt{mixed} denotes tasks that involve both.
The \texttt{complexity} field estimates the length of a pure-GUI solution. We label tasks with 5--15 steps as \texttt{easy}, tasks with 15--40 steps as \texttt{medium}, and tasks with more than 40 steps as \texttt{complex}.
% \texttt{success\_criteria} 以结果清单形式描述任务完成后必须成立的终态条件。\texttt{evaluation\_requirements\_text} 将这些终态条件转换为评估器应执行的检查要求，例如检查具体文件、页面、字段、格式或应用状态。
The \texttt{success\_criteria} field lists the terminal conditions that must hold after successful task completion. The \texttt{evaluation\_requirements\_text} field translates those conditions into concrete checks for evaluator synthesis, such as which file, slide, field, format, or application state should be inspected.
% 指令会写明源与目标位置、范围、顺序、数量、格式和文件名等非答案约束。准确观测值仅保留在隐藏的成功条件和评估要求字段中。
The instruction states non-answer constraints such as locations, scope, ordering, quantity, format, and filenames. Exact observed values are retained only in these hidden success and evaluation fields.

\paragraph{Evaluator synthesis.}
% 评估器合成仅在指令提议被接受后进行，并由独立的评估器智能体在另一上下文中完成。其输入为已冻结的指令字段、初始化配置、成功条件和文字化评估要求。评估器智能体只能生成验证规范，不能改写指令或提议元数据。规则评估器由一个或多个自包含的 Python 判断函数组成，每个函数在函数体内声明其所需导入，因此可以独立运行，并以 \texttt{result} 和 \texttt{expected} 作为输入。每个函数通常对应一个明确的子目标。两个输入均可通过 \texttt{vm\_file} 获取虚拟机内的文件，或通过 \texttt{vm\_command\_line} 获取命令可观测状态。若初始化阶段创建了本地黄金副本，\texttt{expected} 可将其作为内容或结构比较的期望参考；若目标可由规则直接编码，则 \texttt{expected} 可以为空。判断函数按照由细粒度完成点构成的 rubric 累计得分，并为各子项分配相应权重。函数返回 \([0,1]\) 内的分数，完全未满足时为零，全部满足时为一；若执行中发生异常，则返回异常前已累计的分数。评估器仅依赖虚拟机内的产物和确定性规则，不需要外部或云端黄金资产。它检查最终产物的内容、结构、格式、元数据或配置状态。
Evaluator synthesis begins only after an instruction proposal has been accepted and is performed by an independent Evaluator Agent in a separate context. Its inputs are the frozen instruction fields, initialization configuration, success criteria, and textual evaluation requirements. The Evaluator Agent may generate only the verification specification and cannot revise the instruction or proposal metadata. A rule-based evaluator comprises one or more self-contained Python judge functions. Each function declares its required imports locally, can run independently, accepts \texttt{result} and \texttt{expected} as inputs, and typically evaluates one well-defined subgoal. Either input may be supplied by \texttt{vm\_file} as a VM-resident file or by \texttt{vm\_command\_line} as command-observable state. When initialization creates a local golden copy, \texttt{expected} may refer to it for content or structural comparison. When the target can be encoded directly in the rule, \texttt{expected} may be empty. Each judge accumulates a rubric-based score over fine-grained completion items with assigned weights. It returns a value in \([0,1]\), with zero for no satisfied items and one for complete satisfaction; if execution raises an exception, it returns the score accumulated before the exception. The evaluator relies only on VM-resident artifacts and deterministic rules, without external or cloud-hosted golden assets, and checks the final artifact's content, structure, formatting, metadata, or configuration state.

% 当规则能够完整表达成功条件时，评估器设置 \texttt{need\_rule\_judge=true}，无需启用视觉判断。当视觉布局、画布状态、不可解析的应用内部状态或动态 Web 内容构成不可缺少的成功条件时，评估器可以设置 \texttt{need\_vlm\_judge=true}。此时，VLM-Judge 接收智能体的执行轨迹作为输入，\texttt{vlm\_desc} 明确规定应从轨迹中检查的视觉对象、属性和空间关系。
When deterministic rules fully express the success conditions, the evaluator sets \texttt{need\_rule\_judge=true} without enabling visual judgment. When visual layout, canvas state, unparsable application-internal state, or dynamic web content is indispensable for verification, it may set \texttt{need\_vlm\_judge=true}. In this case, the VLM judge receives the agent's execution trajectory as input, and \texttt{vlm\_desc} specifies the visual objects, attributes, and spatial relations to inspect in that trajectory.

% 规则与 VLM 可以联合用于任务验证，但两类得分分别记录。对于包含多个规则判断函数的任务，rule-based score 取所有规则函数得分的平均值；VLM-based score 则由视觉判断单独给出。SFT 阶段分别对两类得分进行质量过滤，要求 rule-based score 和 VLM-based score 均达到各自阈值。
Rule-based and VLM-based judgments may both be used for task verification, but their scores are recorded separately. 
For a task with multiple rule judges, the rule-based score is the mean of the individual rule-judge scores, whereas the VLM-based score is produced independently by visual judgment. During SFT filtering, the two scores are thresholded separately, and a task is retained only when both its rule-based score and VLM-based score meet their respective thresholds (both 0.8). 
% RL 阶段如正文所述，仅使用包含 rule-based 评估器的任务提供奖励信号。这一选择既避免了 VLM 判断的脆弱性和额外计算开销，也避免直接平均两类分数所带来的校准不匹配问题。尽管两类分数都位于 ([0,1])，规则分数基于细粒度、可加权的完成项，而 VLM 分数通常具有更粗粒度且主观的判断尺度，二者并不具有可交换的数值语义。视觉判断因此主要用于 SFT 阶段覆盖程序化接口的观测盲区。
As described in Section~\ref{subsec:evaluation_setup}, RL uses only tasks with rule-based evaluators to provide reward signals. 
This design avoids the brittleness and computational cost of VLM judgment and prevents the calibration mismatch that could arise from averaging the two score types directly. Although both scores lie in \([0,1]\), rule-based scores are derived from fine-grained, explicitly weighted completion items, whereas VLM scores generally reflect a coarser and more subjective judgment scale. The two scores therefore do not have interchangeable numerical semantics. Consequently, VLM judgment is used primarily to cover programmatic observability gaps during SFT-stage filtering.

\paragraph{Instruction and evaluator filtering.}
% 过滤首先检查 JSON 模式、getter 参数和 Python 代码能否被静态解析与编译。随后，系统按照初始化配置启动环境，并在任何动作发生前执行评估器；初态得分必须为零，否则说明评估器存在奖励泄漏。
Filtering first checks whether the JSON schema, getter arguments, and Python code can be parsed and compiled statically. The system then initializes the environment from the initialization configuration and executes the evaluator before any action. The initial-state score must be zero, as a nonzero score indicates reward leakage. 
% 独立审查模型再分别评估指令质量与评估器质量：前者关注指令是否明确、可达且现实，后者关注评估器是否覆盖全部成功条件、足够严格并能够抵抗奖励投机。指令质量低于阈值时直接删除对应提议。若指令有效但评估器未通过自动检查或质量审查，系统会把上一版评估器及其失败信息作为修订上下文，重新生成评估器并完整重复上述检查与评分；经过多次修订仍不合格时，才删除对应提议。
An independent review model next scores instruction quality and evaluator quality separately. The former assesses whether the instruction is unambiguous, reachable, and realistic, while the latter assesses whether the evaluator covers all success conditions, is sufficiently strict, and resists reward hacking. An instruction below the quality threshold is removed directly. If the instruction is valid but its evaluator fails an automated check or quality review, the system uses the previous evaluator and its failure information as revision context, regenerates the evaluator, and repeats the full validation and scoring procedure. The associated proposal is removed only after multiple repair attempts remain unsuccessful.

% 完成正确性过滤后，我们使用嵌入相似度移除跨批次产生的近重复指令，得到用于主要任务池。
After correctness filtering, we use embedding similarity to remove near-duplicate instructions generated across batches, producing the task pool used for training stages. 
% 而RL的任务池是由训练后的SFT模型在任务池里生成轨迹，并根据得分挑选难度适中的任务组成的。
The RL task pool is then built by rolling out trajectories with the SFT-trained model on this pool and keeping tasks of moderate difficulty.

\paragraph{Prototype cross-application extension.}
\label{app:prototype_multiapp}
% 除主要任务池中常规的应用—系统操作组合外，我们还实现并初步采集了更复杂的跨应用任务。该原型利用环境构建阶段软件配置中的 texttt{related\_app} 字段，将应用表示为节点，并依据应用之间共享的文件类型或格式、可转换的中间工件以及互补的工作流关系建立图边，从而将应用环境组织为一张可用于组合采样的应用图。生成任务时，系统首先指定目标应用数量，例如要求任务涉及三个应用，然后随机采样一个主应用，并从其邻接节点中采样后续应用；每次采样后，已选节点的邻接节点都会加入候选集合，以支持沿图结构继续扩展。为避免图结构过度限制组合空间，每一步采样均保留较小概率的跳跃机制，此时可忽略现有边并从任意应用中随机采样。工件采样采用相应的组合思路，可根据当前应用关系采样零个或多个相关工件。与单应用任务相比，这一设置要求探索智能体在更大的状态空间中进行探索，并额外提出应用之间的执行顺序、工件或中间结果传递等依赖关系，进而生成具有明确跨应用依赖的长时程指令。
Beyond the application--system compositions in the primary task pool, we implemented and began collecting a more structured form of cross-application task synthesis. The prototype uses the \texttt{related\_app} field in the application configurations to construct an application graph. Applications serve as nodes, while edges encode compositional relations induced by shared artifact types or file formats, convertible intermediate artifacts, or complementary workflows. This graph provides a structured prior for composing applications rather than treating all application combinations as equally plausible. For each task, the system first specifies a target number of applications, such as three, randomly samples a primary application, and then samples subsequent applications from its neighboring nodes. After each selection, the neighbors of the selected nodes are added to the candidate frontier, allowing the application set to expand along graph structure. To prevent the graph from overly constraining the composition space, each sampling step retains a small-probability jump mechanism that ignores graph connectivity and samples an application uniformly from the broader inventory. Artifact selection follows a corresponding compositional procedure and may introduce zero or more related artifacts. This setting expands the exploration space and requires the Exploration Agent to additionally infer cross-application dependencies, including the execution order and the transfer of artifacts or intermediate results, before proposing a long-horizon instruction.

% 目前，我们已经生成少量此类原型任务并将其纳入任务池，但其规模仍不足以支持系统性的跨应用能力评估；后续工作将扩大采集规模，并进一步研究其对跨应用长时程执行的影响。
We have already generated a small number of such prototype tasks and included them in the task pool. Their current scale is insufficient for a systematic evaluation of cross-application capabilities, so future work will expand the collection and study its effect on long-horizon execution across applications.

\subsection{Instruction and Action Distributions}
\label{app:data_stats}

\begin{figure}[!t]
\centering
\vspace{0.6em}
\includegraphics[width=0.98\linewidth]{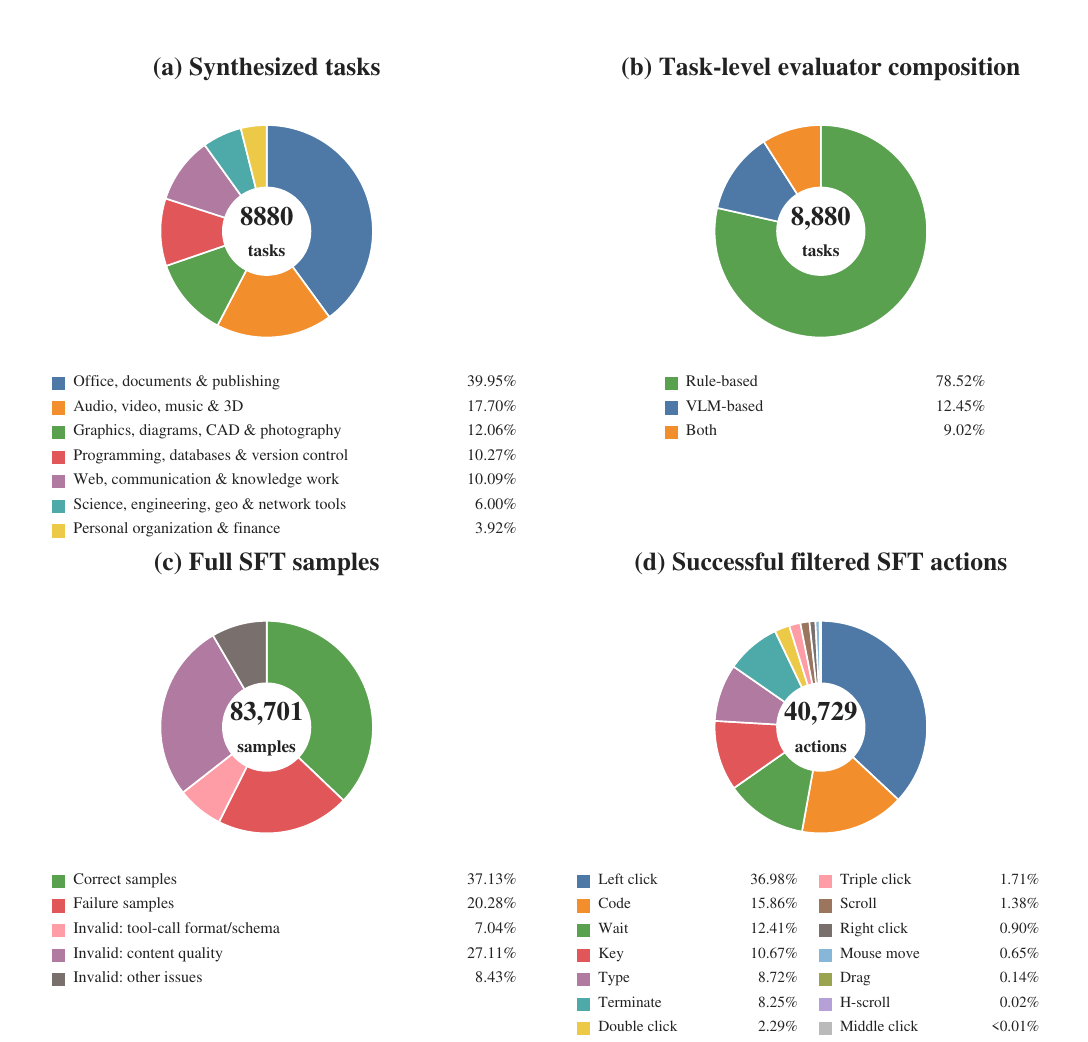}
\vspace{0.6em}

% 中文：四个子图分别统计全部合成指令的应用功能领域、过滤后成功 SFT 样本中的工具动作、全部合成任务的评估器组成，以及全量桌面 SFT 样本的后处理结果。百分比由对应统计对象的原始计数计算，并在图例中保留两位小数。
\captionsetup{skip=4pt, position=bottom}
\caption{Distributions of synthesized instructions, tool actions in successful filtered SFT samples, task-level evaluator composition, and outcomes of the full desktop SFT sample pool. Percentages are computed from the raw counts for each panel and displayed to two decimal places.}
\label{fig:data_distributions}
\end{figure}

% 图~\ref{fig:data_distributions} 从四个互补角度概括合成任务与桌面 SFT 数据。子图 (a) 按应用功能领域统计过滤前生成的全部 8,880 个任务，全量 SFT 轨迹实际覆盖其中 5,416 个任务。子图 (b) 统计过滤后成功 SFT 样本中的 40,729 个工具动作，因此其分母是成功样本中的动作总数，而不是全部桌面 SFT 样本。子图 (c) 在全部 8,880 个生成任务上统计任务级评估器组成，分别区分仅规则、仅 VLM 和规则与 VLM 混合三类。子图 (d) 统计全量 83,701 个桌面 SFT 样本的后处理结果，包括成功样本、失败样本，以及工具调用格式或模式、内容质量和其他问题三类无效样本。
Figure~\ref{fig:data_distributions} summarizes the synthesized tasks and desktop SFT data from four complementary perspectives. 
Panel (a) groups the 8,880 tasks generated before filtering by application domain. 
Panel (b) reports the task-level evaluator composition over all 8,880 generated tasks, distinguishing rule-only, VLM-only, and hybrid rule--VLM evaluation.
Panel (c) partitions the full pool of 83,701 desktop SFT samples, collected from trajectories for 5,416 tasks after filtering, including successful samples, failed samples, and three invalid-sample categories covering tool-call format or schema errors, content-quality issues, and other issues. 
Panel (d) shows the distribution of 40,729 tool actions in the successful SFT samples. 

\subsection{General Data Mixture}
\label{app:sft_data_mixture}

% SFT 数据混合遵循普遍约1:1的混合比例，由约 31K 个过滤后的蒸馏桌面样本和 40K 个通用样本组成。通用数据包含 GUI、代码、数学、OCR、图表和对话数据，用于稳定模型训练并降低模型对桌面轨迹的过度专化。
The SFT data mixture follows a commonly used roughly 1:1 mixing ratio, consisting of approximately 31K filtered distilled desktop samples and 40K general-purpose samples. The general-purpose data covers GUI understanding, coding, mathematics, OCR, chart interpretation, and dialogue, helping stabilize model training and reduce over-specialization to desktop trajectories.

% Let the next appendix's prose use the space below this deferred figure.

\section{Training Details}
\label{app:training_details}

\subsection{SFT Configuration}
\label{app:sft_config}

% 我们使用 Megatron-SWIFT 进行全参数 SFT。训练采用张量并行大小 4、流水线并行大小 1 和 bfloat16 精度；全局批大小为 16、微批大小为 1，训练 1 个 epoch，最大序列长度为 32,768，并启用样本拼接。学习率设为 \(1\times10^{-5}\)，最小学习率为 \(1\times10^{-6}\)，预热比例为 0.05。除表~\ref{tab:sft_training_parameters} 所列参数外，我们还启用完全重计算、FlashAttention、FlashLinearAttention 和序列并行，以高效、稳定地进行长上下文全参数训练。
We perform full-parameter SFT with Megatron-SWIFT~\citep{zhao2024swiftascalablelightweightinfrastructure}. Training uses tensor parallelism of 4, pipeline parallelism of 1, and bfloat16 precision. We set the global and micro batch sizes to 16 and 1, respectively, train for one epoch, use a maximum sequence length of 32,768, and enable sample packing. The learning rate is \(1\times10^{-5}\), with a minimum learning rate of \(1\times10^{-6}\) and a warmup fraction of 0.05. In addition to the settings in Table~\ref{tab:sft_training_parameters}, we enable full recomputation, FlashAttention~\citep{dao2022flashattention}, FlashLinearAttention~\citep{yang2024fla}, and sequence parallelism for efficient and stable long-context full-parameter training.

\begin{table}[H]
\vspace{-0.4\baselineskip}
\centering
\footnotesize
\setlength{\tabcolsep}{3.5pt} 
\renewcommand{\arraystretch}{1.0}
\begin{tabular*}{1.0\linewidth}{p{0.48\linewidth}p{0.48\linewidth}}
\toprule
\textbf{Parameter} & \textbf{Value} \\
\midrule
Training machine & \(8 \times\) Nvidia H200 \\
Training framework & Megatron-SWIFT \\
Training type & Full-parameter fine-tuning \\
Tensor parallel size & 4 \\
Pipeline parallel size & 1 \\
Precision & bfloat16 \\
Global batch size & 16 \\
Micro batch size & 1 \\
Packing & Enabled \\
Training epochs & 1 \\
Maximum sequence length & 32K \\
Optimizer & Adam \\
Adam \(\beta_1 / \beta_2\) & 0.9 / 0.95 \\
Learning rate & \(1\times10^{-5}\) \\
Minimum learning rate & \(1\times10^{-6}\) \\
Warmup fraction & 0.05 \\
Enable Thinking & False \\
% Attention backend & FlashAttention \\
\bottomrule
\end{tabular*}
\captionsetup{skip=2pt, position=bottom,font=footnotesize}
\caption{\ours{} SFT training parameters.}
\label{tab:sft_training_parameters}
\vspace{-0.8\baselineskip}
\end{table}

\subsection{GSPO RL Configuration}
\label{app:gspo_config}

% 强化学习系统以异步框架 Slime 为基础，并在其上实现了在线桌面交互所需的自定义训练逻辑。每次 rollout 采集 16 个 prompt，并为每个 prompt 采样 8 条轨迹，即单次采集 128 条样本。采样温度设为 0.5，最大响应长度为 10,240，并根据任务成功率的指数移动平均进行动态任务采样。Rollout 后端使用 SGLang，策略训练后端使用 Megatron。策略采用 Adam 优化，学习率为 \(1\times10^{-6}\)，权重衰减为 0.1，使用恒定学习率调度，并冻结视觉参数。RL 目标为 Step-balanced GSPO，GUI 和代码步骤的权重均设为 1.0。完整配置见表~\ref{tab:rl_training_parameters}。
Our RL system is built on the asynchronous Slime framework~\citep{slime_github}, augmented with custom training logic for online desktop interaction. Each rollout collection contains 8 prompts and samples 8 trajectories per prompt, yielding 64 rollout samples per collection. We use a maximum of 50 GUI interaction steps, a rollout temperature of 0.5, and a maximum response length of 10,240. Tasks are dynamically sampled according to an exponential moving average of their success rates. SGLang~\citep{zheng2024sglang} serves as the rollout backend, while Megatron~\citep{megatron-lm} is used for policy training. The RL objective is Step-balanced GSPO, with disabled KL divergence and equal weights of 1.0 for GUI and code steps. The full configuration is listed in Table~\ref{tab:rl_training_parameters}.

\subsection{Online RL Sampling, Segmentation, and Rewards}
\label{app:rl_sampling_segmentation}

% 均匀任务采样不适合在线 CUA 训练，因为大量轨迹采集会被耗费在模型已经解决的任务或当前仍无法完成的任务上。遵循以往可扩展 CUA 强化学习工作中的能力边界采样思想~\citep{lv2026scalecua}，我们为每个任务维护成功率估计，并使采样偏向当前模型能力边界附近的任务。
Uniform task sampling is inefficient for online CUA training because many rollouts are spent on tasks that are either already solved or still out of reach. Following the frontier-sampling intuition in prior ScaleCUA~\citep{lv2026scalecua}, we maintain a success-rate estimate for each task and bias sampling toward tasks near the current capability frontier.

% 对于每个任务 \(i\)，我们将 \(\hat{p}_i\) 初始化为 0.5。每个更新窗口结束后，若任务 \(i\) 新获得了 \(n_i\) 条轨迹，则按照下式更新其成功率估计，其中 \(\alpha\) 为 EMA 的历史保留系数，与表~\ref{tab:rl_training_parameters} 中的任务采样 EMA 系数一致。随后，我们使用高斯核计算任务采样权重，其中 \(\mu\) 为目标成功率，\(\sigma\) 控制核的带宽。因此，估计成功率很高或很低的任务会获得较低权重，而成功率接近目标值的任务会被更频繁地采样。
For each task \(i\), we initialize \(\hat{p}_i=0.5\). After each update window, if \(n_i\) new rollouts are available for task \(i\), we update its success-rate estimate by
\[
    \hat{p}_i
    \leftarrow
    \alpha\hat{p}_i
    +
    (1-\alpha)\cdot
    \frac{1}{n_i}\sum_{j=1}^{n_i}\mathbf{1}[r_{i,j}>0],
\]
where \(\alpha\) is the EMA retention coefficient, consistent with the task-sampling EMA parameter in Table~\ref{tab:rl_training_parameters}. The task sampling weight is then computed with a Gaussian kernel,
\[
    w_i
    =
    \exp\left(
    -\frac{(\hat{p}_i-\mu)^2}{2\sigma^2}
    \right),
\]
We set the target success rate to \(\mu=0.5\) and the bandwidth to \(\sigma=0.25\). Thus tasks with very high or very low estimated success rates receive lower weights, while tasks near the target success rate are sampled more often.

% 为控制视觉上下文长度，轨迹保留有界的图像历史，其最大长度为 10，最小长度为 5。按完整轨迹训练会在上下文中保留过多历史截图，而逐步骤拆分会为每个动作步骤生成一个训练样本，使样本数量随轨迹长度膨胀。因此，我们每隔若干交互步骤对轨迹的视觉上下文进行切分。每个训练片段保留最近的视觉上下文，并可包含多个损失掩码为 \(1\) 的 assistant 响应；更早的上下文 token 仅作为条件信息保留，其损失掩码为 \(0\)。与纯逐步骤拆分相比，该设计在提高训练效率的同时，使训练设置更加接近推理过程。
% 为控制视觉上下文长度，我们采用有界滑动窗口策略。每条rollout维护一段包含5至10张截图的历史记录。当历史截图数量超过10张时，我们将当前流保存为一个训练片段，丢弃最旧的5张截图，保留最新的5张截图，并继续收集观测。仅当保留的截图历史处于5至10张范围内时，助手生成的响应才参与损失计算；更早的上下文token仅作为条件信息保留，其损失掩码设为0。
To control visual-context length, we use a bounded sliding-window strategy. Each rollout maintains a history of 5--10 screenshots. When the history exceeds 10 screenshots, we save the current stream as a training segment, discard the five oldest screenshots, retain the five most recent screenshots, and resume collecting observations. Only assistant responses generated while the retained screenshot history lies in this 5--10 range contribute to the loss. Earlier context tokens are retained only as conditioning information with loss mask \(0\). 
% This keeps training closer to inference while improving training efficiency over pure step-wise decomposition.

% 奖励仅来自与任务配套的环境评估器 \texttt{env.evaluate}。我们不使用辅助格式奖励或人工设计的步骤级奖励。对于有效轨迹和片段，评估器分数是位于 \([0,1]\) 内的实数。如果某个 assistant 响应包含无法解析的工具调用、超出允许模式的动作或其他无效动作格式错误，则该响应会被提取为单独的无效样本，并被赋予奖励 \(0\)，即使轨迹其余部分原本可能获得正的任务级得分。这样可以防止格式错误的动作因后续恢复行为而意外获得正向信用。
Rewards come only from the paired environment evaluator, \texttt{env.evaluate}. We do not use auxiliary format rewards or handcrafted step-level rewards. For valid trajectories and segments, evaluator scores are real values in \([0,1]\). If an assistant response contains an unparsable tool call, an action outside the allowed schema, or another invalid action-format failure, that response is extracted as a separate invalid sample and assigned reward \(0\), even if the remaining trajectory could otherwise obtain a positive task-level score. This prevents malformed actions from receiving accidental positive credit through later recovery.

\begin{table}[t]
\vspace{-0.4\baselineskip}
\centering
\footnotesize
\setlength{\tabcolsep}{3.5pt}
\renewcommand{\arraystretch}{1.0}
\begin{tabular*}{1.0\linewidth}{p{0.48\linewidth}p{0.48\linewidth}}
\toprule
\textbf{Parameter} & \textbf{Value} \\
\midrule
\multicolumn{2}{l}{\textbf{Basic Configuration}} \\
Base reinforcement learning framework & Slime \\
Total GPUs & \(32\times\) Nvidia H200 \\
Training GPUs & \(16\times\) Nvidia H200 \\
Rollout GPUs & \(16\times\) Nvidia H200 \\ 
Training steps & 1,000 \\
\midrule
\multicolumn{2}{l}{\textbf{Data}} \\
Task sampling & Success-rate EMA \\
Task-sampling EMA \(\alpha\) & 0.9 \\
Task-sampling min weight / \(\beta_{\mathrm{sample}}\) & 0.1 / 1.0 \\
\midrule
\multicolumn{2}{l}{\textbf{Rollout}} \\
Rollout backend & SGLang \\
Rollout GPUs per engine & 1 \\
Rollout batch size & 8 \\
Samples per prompt & 8 \\
Steps per rollout update & 2 \\
Rollout temperature & 0.5 \\
Maximum GUI steps & 50 \\
Maximum response length & 10,240 \\
Enable Thinking & False \\
Image history length & 5--10 screenshots; retain newest 5 on overflow \\
\midrule
\multicolumn{2}{l}{\textbf{Training}} \\
Training Backend & Megatron \\
Actor GPUs per engine & 4 (tp=4) \\
Frozen parameters & Vision parameters \\
Dynamic batch size & Enabled, max 30,000 tokens/GPU \\
Optimizer & Adam \\
Adam \(\beta_1 / \beta_2\) & 0.9 / 0.95 \\
Learning rate & \(1\times10^{-6}\) \\
Learning-rate schedule & Constant \\
Weight decay & 0.1 \\
\midrule
\multicolumn{2}{l}{\textbf{GSPO Objective}} \\
RL objective & Step-balanced GSPO \\
GUI/code step weights & 1.0 / 1.0 \\
Lower policy clip \(\epsilon_{\mathrm{low}}\) & 0.2 \\
Upper policy clip \(\epsilon_{\mathrm{high}}\) & 0.28 \\
Dual-clip coefficient \(c\) & 3.0 \\
KL loss & Disabled \\
\bottomrule
\end{tabular*}
\captionsetup{skip=2pt, position=bottom,font=footnotesize}
\caption{RL training parameters.}
\label{tab:rl_training_parameters}
\vspace{-0.8\baselineskip}
\end{table}

% \subsection{OSWorld Evaluation Harness}
% \label{app:osworld_eval}

% % 在 OSWorld-Verified 评测中，所有报告的基线和 \ours{} 变体均使用同一套由我们复现的执行框架。观察历史固定最多保留八张截图，解码温度设为 \(0.3\)。除非某条复现基线明确遵循另一套已有日志设置，否则最大交互预算统一设为 50 步。启用 GUI+Code 设置时，我们使用相同的 GUI/代码混合动作实现，并按照评测协议排除 Google Drive 任务。
% For OSWorld-Verified evaluation, we use the same reproduced harness across all reported baselines and \ours{} variants. The observation history keeps a fixed maximum of eight screenshots, and decoding uses temperature \(0.3\). The maximum interaction budget is set to 50 steps unless a reproduced baseline row explicitly follows a different logged setting. We evaluate with the same hybrid GUI/code action implementation whenever the GUI+Code setting is enabled, and we exclude Google Drive tasks following our protocol.

% % 中文：主结果表中的所有基线数值均来自这套执行框架下的复现，而非直接复制外部官方报告。这样可以控制图像历史长度、温度、工具模式、最大步数预算以及后端执行细节，因为这些因素都可能显著改变任务成功率。
% All main-table baseline numbers are reproduced in this harness rather than copied from external official reports. This controls for image-history length, temperature, tool schema, maximum-step budget, and backend execution details, which can otherwise change success rates substantially.

\section{Extra Experiments}
\label{app:extra_experiments}

\begin{table}[t]
  \centering
  \small
  \setlength{\tabcolsep}{2.6pt}
  \renewcommand{\arraystretch}{1.05}
  \resizebox{\textwidth}{!}{ %
  \begin{tabular}{l c c c c c c c c c} % (commented out last c for Avg. w/o Inf)
    \toprule
    \multirow{2}{*}{\textbf{Method}} & \multirow{2}{*}{\textbf{Setting}} & \multirow{2}{*}{\textbf{Max Steps}}
    & \multicolumn{7}{c}{\textbf{Success Rate (\%)}} \\ % was {8}
    \cmidrule(lr){4-10} % was {4-11}
    & & & \textbf{Office} & \textbf{Web} & \textbf{Sys.} & \textbf{Code}
      & \textbf{Media} & \textbf{Util.} & \textbf{Avg.}% & \textbf{Avg. w/o Inf}
      \\
    \midrule
    \noalign{\vskip -0.5ex}
    \rowcolor{lightgray}
    \multicolumn{10}{c}{\textcolor{darkgray}{\textbf{Agentic Frameworks}}} \\[-0.5ex] % was {11}
    \midrule
    Agent S3 w/ GPT-5~\citeyearpar{gonzalez2025unreasonable}
      & Hybrid & 50 & -- & -- & -- & -- & -- & -- & 54.10 % & --
      \\
    Agent S3 w/ GPT-5~\citeyearpar{gonzalez2025unreasonable}
      & Hybrid & 100 & -- & -- & -- & -- & -- & -- & 56.60 % & --
      \\
    OS-Symphony w/ GPT-5-Mini~\citeyearpar{OS-Symphony}
      & Hybrid & 50 & 42.86 & 73.00 & 79.17 & 68.42
      & 48.66 & 66.67 & 62.15 % & 60.64
      \\
    OS-Symphony w/ GPT-5~\citeyearpar{OS-Symphony}
      & Hybrid & 50 & 54.76 & 73.00 & 75.00 & 42.11
      & 70.09 & 75.00 & 63.45 % & 63.44
      \\
    \midrule
    \noalign{\vskip -0.5ex}
    \rowcolor{lightgray}
    \multicolumn{10}{c}{\textcolor{darkgray}{\textbf{General Models \& Specialist Models}}} \\[-0.5ex] % was {11}
    \midrule
    Qwen3-VL-32B-Thinking~\citeyearpar{bai2025qwen3}
      & GUI & 50 & 17.07 & 56.33 & 68.18 & 64.29 & 29.33 & 63.64 & 45.95% & --
      \\
    UI-TARS-1.5-7B~\citeyearpar{qin2025uitars}
      & GUI & 50 & -- & -- & -- & -- & -- & -- & 42.10% & --
      \\
    UI-TARS-2~\citeyearpar{wang2025uiTARS2}
      & GUI & 50 & -- & -- & -- & -- & -- & -- & 50.60% & --
      \\
    UI-Mate-27B~\citeyearpar{ding2026ui}
      & GUI & -- & -- & -- & -- & -- & -- & -- & 66.20% & --
      \\
    EvoCUA-32B~\citeyearpar{EvoCUA}
      & GUI & -- & -- & -- & -- & -- & -- & -- & 56.48% & --
      \\
    EvoCUA-1.5-32B~\citeyearpar{huang2026evocua}
      & GUI & -- & -- & -- & -- & -- & -- & -- & 62.10% & --
      \\
    \midrule
    \noalign{\vskip -0.5ex}
    \rowcolor{lightgray}
    \multicolumn{10}{c}{\textcolor{darkgray}{\textbf{Backbone Baselines}}} \\[-0.5ex] % was {11}
    \midrule
    Qwen3.5-27B~\citeyearpar{qwen2026qwen35}$^{\clubsuit}$
      & GUI & 50 & 30.23 & 36.33 & 62.50 & 50.00
      & 28.41 & 50.00 & 40.82% & 43.88
      \\
    Qwen3.5-27B~\citeyearpar{qwen2026qwen35}$^{\clubsuit}$
      & Hybrid & 50 & 41.86 & 63.00 & 83.33 & 54.17
      & 42.70 & 66.67 & 56.41% & 60.90
      \\
    \midrule
    \ours{27B}
      & Hybrid & 50 & 39.53 & 66.33 & 91.67 & 54.17
      & 39.75 & 75.00 & 57.95% & 63.30
      \\
    \ours{27B-RL}
      & Hybrid & 50 & 39.53 & 76.33 & 75.00 & 73.91
      & 65.81 & 75.00 & 63.87 % & 64.80
      \\
    \bottomrule
  \end{tabular}%
  }
  \captionsetup{skip=3pt,position=bottom}
  \caption{Additional results on WindowsAgentArena. Office includes Writer and Calc; Web includes Edge and Chrome; Sys. denotes Windows system tasks; Code includes VS Code; Media includes VLC; Util. includes Windows utilities. $^{\clubsuit}$ marks baselines reproduced by us.}
  \label{tab:waa_results}
\end{table}

\subsection{Transfer to WindowsAgentArena}
\label{app:waa_generalization}

% 表~\ref{tab:waa_results} 报告 27B 主干及其对应 \ours{} 模型在 WindowsAgentArena 上的额外结果，用于检验从 Ubuntu OSWorld 训练与评测设置到 Windows 桌面环境的跨平台泛化性。我们沿用主实验中的 GUI 与 Hybrid baseline，以保持动作空间和训练贡献之间的比较关系一致。
Table~\ref{tab:waa_results} reports additional results for the 27B backbone and its corresponding \ours{} variants on WindowsAgentArena, providing evidence of transfer from the Ubuntu-based training setting to this Windows benchmark. We retain the GUI and Hybrid baselines from the main experiments to preserve the same comparison between action-space access and learned hybrid behavior.
% 特别注意的是，由于操作系统环境是Windows，我们将Code动作强制为使用 Python 语言，而排除了 Bash 语言的使用，若使用 Cmd/PowerShell 语言，则直接在 Python 中调用系统命令，这一点通过修订提示词实现。
Of particular note, since the operating system environment is Windows, we restrict Code actions to Python only, excluding the use of Bash. When Cmd/PowerShell commands would otherwise be required, system calls are instead issued directly within Python; this adaptation is implemented via prompt revision.

Despite the absence of Windows-specific training data, \ours{} improves on this benchmark. The SFT-only \ours{27B} raises the matched GUI backbone from 40.82\% to 57.95\%, a gain of 17.13 absolute points, and the RL variant reaches 63.87\%, exceeding the GUI baseline by 23.05 points and the Hybrid baseline by 7.46 points. The unadapted Hybrid baseline also improves substantially over GUI (56.41\% versus 40.82\%), while \ours{27B} adds a further 1.54 points. These comparisons provide benchmark-specific evidence that learned GUI/code routing transfers to WindowsAgentArena under the reported Python-only prompt adaptation.

The remaining gap is also diagnostic. Manual inspection of failed WindowsAgentArena trajectories shows that \ours{27B} often identifies the correct high-level strategy but can fail in the concrete system-command layer, mixing Bash-like patterns into Windows execution or producing brittle PowerShell-style logic despite the Python-only prompt adaptation. 
This pattern is compatible with the absence of Windows-specific command data. 
In contextual comparisons reported on the same benchmark, \ours{27B-RL} slightly exceeds OS-Symphony w/ GPT-5 and Agent S3 w/ GPT-5 under the shared 50-step budget. These results motivate scaling the synthesis pipeline along the operating-system dimension and evaluating on broader Windows and macOS task distributions.

We further analyze infeasible tasks as a complementary limitation of the current data pipeline. Both OSWorld and WindowsAgentArena contain a non-negligible fraction of infeasible tasks, roughly 10\% of their task sets. These tasks mix at least two phenomena: genuinely impossible instructions whose target condition cannot be satisfied, and GUI-infeasible tasks where the visual route may be blocked while a programmatic route could still modify the underlying state. Our synthesis pipeline is designed around executable goal states and therefore does not naturally generate infeasible training tasks. As a result, the model can under-learn when to stop and declare infeasibility, often continuing until the maximum interaction budget is exhausted. Conversely, for tasks that are infeasible only under GUI interaction, a hybrid agent may legitimately solve them through code, which can still be judged as incorrect if the benchmark expects infeasibility recognition. As a diagnostic, after removing infeasible tasks from the evaluation set, the average success rate on OSWorld and WindowsAgentArena increases by up to \textbf{3--5 absolute points}. This suggests that future task design and data synthesis should explicitly model infeasibility, distinguishing true task impossibility from modality-specific infeasibility so that agents can learn both effective execution and calibrated refusal.

\subsection{Additional Analysis}
\label{app:additional_analysis}

% 本章额外分析中的 grounding 实验使用 Qwen3 系列，而非上文实验使用的 Qwen3.5 系列；训练数据保持完全一致。
The diagnostic experiments in this subsection are conducted with Qwen3-series models~\citep{bai2025qwen3}, which differ from the Qwen3.5-series base models used in the experiments above. The training data and data recipe, however, are kept exactly the same, so the comparison isolates how the same \ours{} training signal behaves on a different backbone family.

All experiments in this subsection use supervised fine-tuning (SFT) only and do not include reinforcement learning (RL). Accordingly, the analyses below study the transferability of the learned GUI/code behavior and the information content of the SFT data, rather than the contribution of online policy optimization.

\begin{table}[t]
  \centering
  \small
  \setlength{\tabcolsep}{5.5pt}
  \renewcommand{\arraystretch}{1.05}
  \begin{tabular*}{\textwidth}{@{\extracolsep{\fill}}lcc}
    \toprule
    \textbf{Model} & \textbf{OSWorld-G} & \textbf{ScreenSpot-V2} \\
    \midrule
    Qwen3-VL-32B-Instruct & 65.10 & 95.80 \\
    Qwen3-VL-32B-Instruct$^{\clubsuit}$ & 61.35 & 95.05 \\
    \midrule
    \ours{32B} & 57.21 & 93.47 \\
    \bottomrule
  \end{tabular*}
  \captionsetup{skip=3pt,position=bottom}
  \caption{Grounding benchmark results on OSWorld-G and ScreenSpot-V2. $^{\clubsuit}$ represents the result reproduced by us.}
  \label{tab:grounding_results}
\end{table}

\subsubsection{Grounding Ability Check}
\label{app:grounding_sanity}

% 表~\ref{tab:grounding_results} 报告了 Qwen3-VL-32B-Instruct 及以 Qwen3-VL-32B 为基础、采用与我们 Qwen3.5 系列模型相同数据方案训练的 \ours{32B} 变体的仅定位（grounding-only）合理性检验结果。复现的基础模型在 OSWorld-G 上取得 61.35 分，在 SS-V2 上取得 95.05 分；而 \ours{32B} 在上述两项指标上分别取得 57.21 与 93.47 分。
Table~\ref{tab:grounding_results} reports grounding benchmark performance on Qwen3-VL-32B-Instruct and an \ours{32B} variant with the same data recipe as our Qwen3.5-series models. The reproduced base model obtains 61.35 on OSWorld-G~\citep{xie2026scaling} and 95.05 on ScreenSpot-V2~\citep{wu2025atlas}, while \ours{32B} obtains 57.21 and 93.47, respectively.

% 这一下降是符合预期的，与我们的主实验结果并不矛盾。训练目标并非仅优化像素级定位能力，还同时教导模型格式化工具调用、对文件进行推理、在程序化执行更为可靠时调用代码，以及从长时序环境反馈中恢复。从这一角度来看，模型将部分能力从孤立的定位基准测试转移至混合桌面执行任务。关键的观察在于：即便纯定位能力略有下降，具备代码能力的动作选择也能在端到端任务中弥补这一损失——通过规避脆弱的像素级交互，转而借助操作系统来完成精确的文件或数据操作。
This decrease is expected rather than contradictory to our main results. The training objective does not only optimize pixel grounding; it also teaches the model to format tool calls, reason over files, invoke code when programmatic execution is more reliable, and recover from multi-step environment feedback. In this sense, the model reallocates some capacity away from isolated grounding benchmarks toward hybrid desktop execution. The key observation is that even when pure grounding slightly decreases, code-capable action selection can compensate for this loss in end-to-end tasks by avoiding brittle pixel-level interactions and solving precise file or data operations through the operating system.

\subsubsection{Backbone Replacement Analysis}
\label{app:training_model_analysis}

% 中文注释：本实验应在相同 Setting 下比较。OS-Helix 的 Hybrid 结果分别与 Qwen3-VL-32B-Instruct Hybrid 和 Qwen3-VL-32B-Thinking Hybrid 对比；表中我们模型右侧的箭头和数值表示相对同 Setting baseline 的绝对百分点变化。GUI 与 Hybrid baseline 的差异仅用于说明直接开放 Code 工具本身并不能稳定带来收益。
To test whether the training signal transfers beyond the original backbone, we repeat the SFT experiment with Qwen3-VL-32B-Instruct and Qwen3-VL-32B-Thinking. Table~\ref{tab:training_model_analysis} reports the complete OSWorld breakdown; arrows in the \ours{} rows denote absolute-point changes relative to the Qwen3-VL baseline under the same Setting. With the Instruct backbone, adding Code without training a hybrid policy slightly decreases the baseline from 34.17\% (GUI) to 32.68\% (Hybrid). After SFT, however, \ours{32B} reaches 50.56\%, a +17.88-point improvement over the matched Hybrid baseline. With the Thinking backbone, the unadapted Hybrid baseline is somewhat better than GUI (42.88\% versus 39.08\%), but \ours{32B-Thinking} still improves it by +8.79 points to 51.67\%. Thus, the result isolates two effects: Code availability is not itself sufficient for reliable hybrid execution, while the SFT training signal transfers across both backbones and converts the available action space into a substantially stronger policy.

The domain-level changes clarify what is learned. For Instruct, \ours{32B} improves all five domains over the matched Hybrid baseline. The corresponding gains in OS, Office, Daily, Prof., and Work. are 16.67, 19.66, 16.83, 12.24, and 19.76 points, respectively. The gains are therefore broad rather than being explained by a single application family, with the largest improvements in Office and multi-application tasks in the Work. category. This pattern is consistent with learning cross-step state management and modality routing, rather than simply improving visual grounding on isolated screens. For Thinking, the gains remain broad in OS (+8.33), Office (+11.94), Daily (+11.40), and Work. (+9.40), but Prof. decreases by 2.83 points. The single negative domain is a useful qualification: backbone replacement does not erase model-specific priors, and SFT can redistribute capability unevenly across application families. Nevertheless, the positive average and the gains across the OS, Office, Daily, and Work. categories show that the central hybrid behavior is not tied to one backbone's pretrained tool-use bias.

These results also strengthen the distinction between action-space access and policy learning in the main claim. The Instruct baseline provides a controlled counterexample in which the same model becomes slightly worse when Code is merely exposed at inference time. In contrast, the trained model gains most strongly on domains where successful execution requires coordinating GUI observations, executable operations, and persistent task state. The improvement should therefore be interpreted as evidence for learned coordination and routing, not as an artifact of giving the model an additional tool.

\subsubsection{Training Recipe and Context Analysis}
\label{app:training_recipe_analysis}

We next isolate how the structure of demonstrations and test-time history affects performance, using Qwen3-VL-32B-Instruct with OSWorld. The normal recipe reaches 50.56\%. Removing CoT from training and retaining actions only reduces performance to 37.74\% (-12.82 points), while keeping the training data unchanged but removing CoT from the inference context yields 37.82\% (-12.74 points). The near-identical degradation is important: the reasoning trace is not merely an auxiliary training target. It also functions as an execution-time state representation that helps the model maintain intent, resolve intermediate observations, and select between GUI and code actions across multi-step execution.

Removing only historical CoT is less damaging, but still lowers the average to 43.22\% (-7.34 points). This partial recovery indicates that the current-step decision can retain useful local information, yet removing the accumulated reasoning history weakens temporal credit assignment and state tracking. The effect is especially visible in the Work. category, where success falls from 38.54\% to 28.87\% without training CoT, 31.43\% without inference CoT, and 31.30\% without historical CoT. Multi-application tasks therefore expose the central role of structured context most clearly: their difficulty lies not only in individual grounding decisions, but in preserving a coherent plan while the active application and tool modality change.

Finally, mixing in trajectories rejected during filtering reduces the score to 47.06\% (-3.50 points). This smaller but consistent decline suggests that data quality is not interchangeable with data quantity. Rejected failures may preserve superficial action patterns, but they also introduce inconsistent or non-executable supervision that blurs the association between observations, tool calls, and successful state transitions. Taken together, the ablations support a precise interpretation of the SFT recipe: CoT and history provide the structured latent state needed for hybrid execution, while rejection filtering protects the reliability of the behavior-to-outcome mapping. The full comparison is given in Table~\ref{tab:training_recipe_analysis}.

\begin{table}[t]
  \centering
  \footnotesize
  \setlength{\tabcolsep}{1.8pt}
  \resizebox{\linewidth}{!}{%
  \begin{tabular}{l c c c c c c c c}
    \toprule
    \textbf{Model} & \textbf{Setting} & \textbf{Step} & \textbf{OS} & \textbf{Office} & \textbf{Daily} & \textbf{Prof.} & \textbf{Work.} & \textbf{Avg.} \\
    \midrule
    Qwen3-VL-32B-Instruct$^{\clubsuit}$ & GUI & 50 & 41.67 & 31.54 & 51.23 & 44.90 & 15.57 & 34.17 \\
    Qwen3-VL-32B-Instruct$^{\clubsuit}$ & Hybrid & 50 & 50.00 & 33.25 & 39.31 & 38.78 & 18.78 & 32.68 \\
    Qwen3-VL-32B-Thinking$^{\clubsuit}$ & GUI & 50 & 58.33 & 35.84 & 54.44 & 55.10 & 18.71 & 39.08 \\
    Qwen3-VL-32B-Thinking$^{\clubsuit}$ & Hybrid & 50 & 62.50 & 40.97 & 49.81 & 53.85 & 27.95 & 42.88 \\
    \midrule
    \ours{32B} & Hybrid & 50 & \gaincell{66.67}{16.7} & \gaincell{52.91}{19.7} & \gaincell{56.14}{16.8} & \gaincell{51.02}{12.2} & \gaincell{38.54}{19.8} & \gaincell{50.56}{17.9} \\
    \ours{32B-Thinking} & Hybrid & 50 & \gaincell{70.83}{8.3} & \gaincell{52.91}{11.9} & \gaincell{61.21}{11.4} & \losscell{51.02}{2.8} & \gaincell{37.35}{9.4} & \gaincell{51.67}{8.8} \\
    \bottomrule
  \end{tabular}%
  }
  \captionsetup{skip=3pt,position=bottom}
  \caption{Backbone replacement analysis on OSWorld. $^{\clubsuit}$ marks baselines reproduced by us.}
  \label{tab:training_model_analysis}
\end{table}

\begin{table}[t]
  \centering
  \small
  \setlength{\tabcolsep}{3.2pt}
  \resizebox{\linewidth}{!}{%
  \begin{tabular}{l l c c c c c c}
    \toprule
    \textbf{Training Recipe} & \textbf{Inference Context} & \textbf{OS} & \textbf{Office} & \textbf{Daily} & \textbf{Prof.} & \textbf{Work.} & \textbf{Avg.} \\
    \midrule
    Normal & Normal & 66.67 & 52.91 & 56.14 & 51.02 & 38.54 & 50.56 \\
    \midrule
    Normal & $-$ CoT & \losscell{54.17}{12.5} & \losscell{28.98}{23.9} & \losscell{53.58}{2.6} & \losscell{38.77}{12.3} & \losscell{31.43}{7.1} & \losscell{37.82}{12.7} \\
    $-$ CoT & Normal & \losscell{50.00}{16.7} & \losscell{35.55}{17.4} & \losscell{46.99}{9.2} & \losscell{38.77}{12.3} & \losscell{28.87}{9.7} & \losscell{37.74}{12.8} \\
    $-$ History CoT & Normal & \losscell{60.87}{5.8} & \losscell{43.51}{9.4} & \losscell{54.19}{2.0} & \losscell{39.80}{11.2} & \losscell{31.30}{7.2} & \losscell{43.22}{7.3} \\
    $+$ Failure Data & Normal & \losscell{54.17}{12.5} & \losscell{48.67}{4.2} & \losscell{54.80}{1.3} & \losscell{50.28}{0.7} & \losscell{35.16}{3.4} & \losscell{47.06}{3.5} \\
    \bottomrule
  \end{tabular}%
  }
  \captionsetup{skip=3pt,position=bottom}
  \caption{Training-recipe ablations on OSWorld with Qwen3-VL-32B-Instruct.
  ``$-$ CoT'' denotes the removal of all CoT text while retaining action text only;
  ``$-$ History CoT'' denotes the removal of historical CoT while retaining the current-step CoT and action text;
  ``$+$ Failure Data'' denotes augmenting the SFT samples with partial rejection-sampled failure samples.}
  \label{tab:training_recipe_analysis}
\end{table}

\begin{table}[htbp]
\vspace{-0.4\baselineskip}
\centering
\footnotesize
\setlength{\tabcolsep}{3pt}
\renewcommand{\arraystretch}{0.9}
\begin{tabular*}{\textwidth}{@{\extracolsep{\fill}}llll@{}}
\toprule
\textbf{Metric} & \textbf{Comparison} & \textbf{$\Delta$ (pp)} & \textbf{95\% CI} \\
\midrule
Primary score (Pass@1) & SFT $-$ Hybrid baseline & +16.21 & [+10.75, +21.66] \\
Primary score (Pass@1) & RL $-$ Hybrid baseline & +19.89 & [+14.24, +25.59] \\
Primary score (Pass@1) & RL $-$ SFT & +3.68 & [+0.93, +8.33] \\
Pass@3 & RL $-$ SFT & +5.56 & [+2.24, +8.97] \\
\bottomrule
\end{tabular*}
\captionsetup{skip=2pt, position=bottom,font=footnotesize}
\caption{Paired-bootstrap confidence intervals on OSWorld. We resample the 361 tasks with replacement for 100,000 replicates and recompute the paired mean-score difference.}
\label{tab:paired_bootstrap_ci}
\vspace{-0.8\baselineskip}
\end{table}

\FloatBarrier

\section{Step-Balanced GSPO Details}
\label{app:step_balanced_gspo}

% 在强化学习阶段，一个训练样本可以包含多个连续动作步骤对应的 assistant 回复，而不是只包含一个动作步骤。
During reinforcement learning, a training sample may contain multiple assistant responses that contribute to the policy loss, with each response corresponding to one action step.
% 因此，训练样本的优化单位与其中的环境动作步骤并不天然一致。
The optimization unit of a training sample therefore need not coincide with an individual environment action step.
% 具体来说，code action step 对应的完整 assistant span 在 token 空间里通常比 GUI step 长得多。
In particular, the assistant span associated with a code action step is often substantially longer than that associated with a GUI step.
% 标准 GSPO 目标通过对训练样本中所有有效 assistant token 的 token-level log-importance weight 求平均，计算一个样本级 importance ratio。
The standard GSPO objective computes one sample-level importance ratio by aggregating token-level log-importance weights over all valid assistant tokens in the training sample~\citep{zheng2025gspo}.
% 虽然这能避免训练样本仅因为 token 更多而获得更大的 sample-level loss，但它无法平衡同一样本内不同 action step 的贡献。
Although this prevents a sample from receiving a larger sample-level weight solely because it contains more tokens, it does not equalize the contributions of different action steps within the same sample.
% 因此，即使一个长 code 动作只是训练样本中多个动作之一，它也可能主导样本级比率和裁剪决策。
As a result, a long code action can dominate both the sample-level ratio and the clipping decision even when it is only one of several actions in the training sample.

% 本节形式化刻画这一影响，并推导 GSPO ratio 的 step-balanced 聚合方式。
This section formalizes the effect and derives a step-balanced aggregation for the GSPO ratio.
% 我们的目标是保留 GSPO 的 token-level importance weight 与裁剪结构，同时改变样本级聚合方式，以消除动作步骤的聚合梯度系数对 token 长度的确定性依赖。
Our goal is to preserve GSPO's token-level importance weights and clipping structure while changing their sample-level aggregation to remove the deterministic dependence of an action step's aggregate gradient coefficient on its token count.
% 已有工作也研究了 LLM policy optimization 中 importance ratio 的粒度。
Recent work has also studied the granularity of importance ratios for LLM policy optimization, including sequence-level GSPO~\citep{zheng2025gspo}, subsentence-level ratios~\citep{yang2025sspo}, cumulative token ratios~\citep{zhang2026ctpo}, and hybrid token/sequence ratios~\citep{min2026dhpo}.
% 中文：据我们所知，这些方法没有处理一个训练样本内由异构 GUI/code 动作导致的 action-step 不平衡。
These methods differ from our setting at the objective level. SSPO assigns separate ratios to text fragments, but its published objective aggregates fragments with token-length normalization and does not use environment-action boundaries or action-type weights. CTPO and DHPO use token-level (or token/response-hybrid) objectives, while GSPO and GSPO-token use one response-level ratio. None of these objectives replaces token-count weights by normalized weights over labeled GUI/code action steps within a multi-action training sample.

\paragraph{Notation.}
% 中文：令 i 表示一个训练样本。
Let \(i\) denote a training sample.
% 中文：训练样本 i 包含 K_i 个动作步骤。
Training sample \(i\) contains \(K_i\) action steps.
% 中文：我们用 a_{i,k} 表示第 k 个 action step 对应的全部参与 loss 的 assistant token 位置集合，用 |a_{i,k}| 表示其 token 数。
We denote by \(a_{i,k}\) the set of assistant-token positions that contribute to the loss for action step \(k\), and by \(|a_{i,k}|\) its cardinality. Thus, \(a_{i,k}\) may include reasoning tokens, tool-format wrapper tokens, and action-content tokens.
% 中文：令 m_{i,t} 表示 token 位置 t 的 loss mask，并定义 token-level importance weight 及其对数形式。
Let \(m_{i,t}\in\{0,1\}\) indicate whether token position \(t\) contributes to the loss. For each loss-participating assistant token, define the token-level importance weight and its logarithm as
\[
    w_{i,t}(\theta)
    =
    \frac{\pi_\theta(y_{i,t}\mid h_{i,t})}
    {\pi_{\mathrm{old}}(y_{i,t}\mid h_{i,t})},
    \qquad
    \delta_{i,t}
    =
    \log w_{i,t}(\theta)
    =
    \log \pi_\theta(y_{i,t}\mid h_{i,t})
    -
    \log \pi_{\mathrm{old}}(y_{i,t}\mid h_{i,t}).
\]
% 中文：其中，y_{i,t} 是采样出的 token，h_{i,t} 是其因果上下文。
Here, \(y_{i,t}\) is the sampled token and \(h_{i,t}\) is its preceding causal context.
% 中文：样本 i 中有效 assistant token 的总数为 N_i。
The total number of valid assistant tokens in sample \(i\) is
\[
    N_i
    =
    \sum_t m_{i,t}
    =
    \sum_{k=1}^{K_i}|a_{i,k}|.
\]
% 中文：对每个 action step，定义 step-level mean log-importance weight。
For each action step, we define the mean token-level log-importance weight as
\[
    \bar{\delta}_{i,k}
    =
    \frac{1}{|a_{i,k}|}
    \sum_{t\in a_{i,k}}\delta_{i,t}.
\]
% 中文：令 \tau_{i,k} 表示第 k 个 step 的动作类型，例如 GUI 或 code。
Let \(\tau_{i,k}\) denote the action type of step \(k\), e.g., \(\tau_{i,k}\in\{\mathrm{GUI},\mathrm{code}\}\).
% 中文：我们为每一种动作类型分配一个非负 balancing coefficient \alpha_{\tau_{i,k}}。
We assign each action type a nonnegative balancing coefficient \(\alpha_{\tau_{i,k}}\). At least one action type in each sample must have a positive coefficient so that the normalizer below is nonzero.

\subsection{GSPO}
\label{app:standard_gspo_ratio}

% 中文：样本 i 的标准 GSPO sample-level importance ratio 如下。
The standard GSPO sample-level importance ratio for sample \(i\) is
\[
    w_i^{\mathrm{GSPO}}(\theta)
    =
    \exp\left(
    \frac{1}{N_i}
    \sum_t m_{i,t}\delta_{i,t}
    \right).
\]
% 中文：按 action step 对有效 token 分组后，可以得到下面的等价形式。
Grouping the loss-participating tokens by action step gives
\[
    \log w_i^{\mathrm{GSPO}}(\theta)
    =
    \sum_{k=1}^{K_i}
    \frac{|a_{i,k}|}{N_i}
    \bar{\delta}_{i,k}.
\]
% 中文：因此，在标准 GSPO 中，第 k 个 action step 在 sample-level log-ratio 中的隐式权重为：
Therefore, in standard GSPO, the implicit weight of action step \(k\) in the sample-level log-ratio is
\[
    \omega_{i,k}^{\mathrm{GSPO}}
    =
    \frac{|a_{i,k}|}{\sum_j |a_{i,j}|}.
\]
% 中文：这个权重由 token 数决定，而不是由动作语义决定。
This weight is determined by token count rather than by action semantics.
% 中文：如果一个样本包含两个短 GUI 动作和一个长 code 动作，那么 code 动作可能仅因为有效 assistant token 更多就主导 ratio。
If a sample contains two short GUI actions and one long code action, the code action can dominate \(\log w_i^{\mathrm{GSPO}}(\theta)\) simply because it has more valid assistant tokens.
% 中文：同一个 ratio 随后会用于样本内所有 token，因此长动作也会对整个样本是否被 clip 产生不成比例的影响。
The same ratio is then used for all tokens in the sample, so the long action also has disproportionate influence over whether the whole sample is clipped.

\subsection{Step-Balanced GSPO}
\label{app:step_balanced_ratio}

% 中文：我们将基于 token 数的 action 权重替换为显式的 action-level 权重。
We replace token-count-based action weighting with explicit action-level weighting. Define
\[
    Z_i
    =
    \sum_{k=1}^{K_i}\alpha_{\tau_{i,k}},
    \qquad
    \omega_{i,k}
    =
    \frac{\alpha_{\tau_{i,k}}}{Z_i}.
\]
% 中文：Step-balanced GSPO sample-level importance ratio 定义如下。
The step-balanced GSPO sample-level importance ratio is
\[
    \boxed{
    w_i^{\mathrm{step}}(\theta)
    =
    \exp\left(
    \sum_{k=1}^{K_i}
    \omega_{i,k}\bar{\delta}_{i,k}
    \right)
    }.
\]
% 中文：等价地，可以把它写成显式的 importance ratio 形式。
Equivalently, the ratio can be written directly in terms of policy log-probabilities as
\[
    \boxed{
    w_i^{\mathrm{step}}(\theta)
    =
    \exp\left[
    \frac{
    \sum_{k=1}^{K_i}
    \alpha_{\tau_{i,k}}
    \left(
    \frac{1}{|a_{i,k}|}
    \sum_{t\in a_{i,k}}
    \log
    \frac{\pi_\theta(y_{i,t}\mid h_{i,t})}
         {\pi_{\mathrm{old}}(y_{i,t}\mid h_{i,t})}
    \right)
    }{
    \sum_{k=1}^{K_i}\alpha_{\tau_{i,k}}
    }
    \right]
    }.
\]
This formulation differs from standard GSPO only in the aggregation weights used to form the sample-level log-ratio. Standard GSPO weights each action step in proportion to its number of valid assistant tokens, whereas Step-balanced GSPO uses explicit action-level weights. When these weights are set proportional to token counts, the formulation recovers the original GSPO ratio as a special case.

% 该 policy objective 保持原始 GSPO 的结构。
The policy objective retains the standard GSPO structure. Given the sample-level advantage \(A_i\), the unclipped objective is \(J_i(\theta)=A_i w_i^{\mathrm{step}}(\theta)\), and the clipped objective is
\[
    J_i^{\mathrm{clip}}(\theta)
    =
    \min\left(
    A_i w_i^{\mathrm{step}}(\theta),
    A_i\operatorname{clip}\left(
    w_i^{\mathrm{step}}(\theta),
    1-\epsilon_{\mathrm{low}},
    1+\epsilon_{\mathrm{high}}
    \right)
    \right).
\]
Any additional dual-clipping rule for negative advantages is applied in the same way as in standard GSPO, after replacing its sample-level ratio with \(w_i^{\mathrm{step}}(\theta)\). In the reported experiments, the KL loss is disabled (Table~\ref{tab:rl_training_parameters}), so the objective contains no additional token-level KL term. If a KL penalty is enabled in future variants, applying corresponding step-level weighting would avoid reintroducing token-count-dependent regularization. The per-sample training loss is \(L_i(\theta)=-J_i^{\mathrm{clip}}(\theta)\), and the batch loss over \(B\) samples is
\[
    L(\theta)=\frac{1}{B}\sum_{i=1}^{B}L_i(\theta).
\]

\subsection{Gradient Contribution}
\label{app:step_balanced_gradient}

% 下面证明该 ratio 可以移除 action step 聚合标量梯度系数与 token 长度之间的确定性线性依赖。
We now show that the proposed ratio removes the deterministic linear dependence between an action step's aggregate scalar gradient coefficient and its token count.
% 考虑未进入 clipped 分支的情况，此时 loss 为：
Consider the unclipped branch, where \(L_i(\theta)=-A_i w_i^{\mathrm{step}}(\theta)\).
% 对于任意 token t\in a_{i,k}，只有 \bar{\delta}_{i,k} 依赖于该 token 的当前 policy log-probability。
For any token \(t\in a_{i,k}\), only \(\bar{\delta}_{i,k}\) depends on the corresponding current-policy log-probability. Therefore,
\[
    \frac{
    \partial \log w_i^{\mathrm{step}}(\theta)
    }{
    \partial \log \pi_\theta(y_{i,t}\mid h_{i,t})
    }
    =
    \omega_{i,k}
    \frac{1}{|a_{i,k}|}.
\]
% 由于 w_i^{step} 对其 log 值的导数为自身，可以得到每个 token 的 loss 对当前 policy log-probability 的导数。
Since \(\partial w_i^{\mathrm{step}}/\partial \log w_i^{\mathrm{step}}=w_i^{\mathrm{step}}\), we have
\[
    \frac{
    \partial L_i
    }{
    \partial \log \pi_\theta(y_{i,t}\mid h_{i,t})
    }
    =
    -
    A_i w_i^{\mathrm{step}}(\theta)
    \frac{\omega_{i,k}}{|a_{i,k}|}.
\]
% 中文：对第 k 个 action step 内的所有 token 求和，得到：
Summing over all tokens in action step \(k\) gives
\[
    \sum_{t\in a_{i,k}}
    \frac{
    \partial L_i
    }{
    \partial \log \pi_\theta(y_{i,t}\mid h_{i,t})
    }
    =
    -
    A_i w_i^{\mathrm{step}}(\theta)
    \omega_{i,k}.
\]

% 因此，第kk个动作步骤的聚合标量系数与∣ai,k∣无关。这正是步骤均衡GSPO改变优化信号的意义所在：它保持token级重要性权重wi,t​不变，但改变了这些权重的对数值聚合为样本级重要性比率的方式。在标准GSPO中，一个较长的序列化动作在样本级对数比率中占据更大的份额，因此一个成功的代码步骤仅因包含更多助手token便可获得更多的聚合标量信用。这并不意味着选择代码工具的路由token会被后续代码token的数量直接放大；而是指在相同的样本级优势下，较长的序列化动作会贡献更多的token级梯度项，进而可能对更新方向产生不成比例的影响。步骤均衡GSPO从动作步骤聚合系数中消除了这一基于长度的确定性乘数，以步骤为单位对成功动作进行强化，而非与其序列化长度成比例。完整的参数梯度仍然依赖于各token的梯度方向、裁剪激活状态以及奖励相关性，因此本分析解释了token长度驱动的优化偏差向代码动作漂移的一个来源，而非对正确模态路由的形式化保证。
Thus, the aggregate scalar coefficient of action step \(k\) is independent of \(|a_{i,k}|\). This is the sense in which Step-balanced GSPO changes the optimization signal: it leaves the token-level importance weights \(w_{i,t}\) intact, but changes how their log values are aggregated into the sample-level importance ratio. Under standard GSPO, a long serialized action occupies a larger share of the sample-level log-ratio, so a successful code step can receive more aggregate scalar credit simply because it contains more assistant tokens. 
This does not mean that the routing token selecting the code tool is directly multiplied by the number of subsequent code tokens. Rather, the longer serialized action contributes more token-level gradient terms under the same sample-level advantage and can therefore exert disproportionate influence on the update direction. Step-balanced GSPO removes this deterministic length-based multiplier from the action-step aggregate coefficient, reinforcing a successful action as a step rather than in proportion to its serialized length. The full parameter gradient still depends on token-specific gradient directions, clipping activation, and reward correlations, so this analysis explains a source of token-length-driven drift toward code actions rather than providing a formal guarantee of correct modality routing.

\section{Case Study}
\label{app:case_study}

% 本节展示两个经过筛选的合成任务，分别覆盖办公/代码工作流和媒体工作流。第一个任务基于 TableBench 表格，第二个任务则在录屏工具与视频编辑器之间构建并迁移持久化应用配置。二者共同展示了应用--产物组合、基于环境的任务规范，以及可执行任务记录中的组件级终态验证。这两个任务都使用经过人工检查的规则评估器，在多个可独立验证的任务组件上分配部分终态分数。我们复现任务元数据、完整指令和部分规则代码，以具体说明这些成功条件。
This section presents two filtered synthesis tasks covering an office/code workflow and a media workflow. The first is grounded in a spreadsheet, whereas the second constructs and transfers persistent application configuration between a recording tool and a video editor. Together, they illustrate application--artifact composition, environment-grounded task specification, and component-wise terminal verification in executable task records. Both use reviewed rule-based evaluators that assign partial terminal credit across independently verifiable task components. We reproduce the task metadata, complete instructions, and selected rule code to make these success conditions concrete.

\subsection{Synthesis Task Case Study 1: Office and Code Workflow}
\label{app:case_spreadsheet_code}

% 第一个任务跨越 LibreOffice Calc 和 VSCode。任务从包含车辆记录的 TableBench 表格开始，要求在电子表格中进行逐行分类，并进一步要求针对保存后的产物编写可运行的 Python 审计脚本。该任务标记为复杂，预计需要 30 个交互步骤。
The first task spans LibreOffice Calc and VSCode. It starts from a spreadsheet containing vehicle records, requires row-level categorization in the spreadsheet, and then asks for a runnable Python audit script over the saved artifact. The task is labeled complex and estimated to require 30 interaction steps.

\begin{tcolorbox}[title={Reviewed Synthesized Task 1}, colback=black!2, colframe=black!55, boxrule=0.5pt, left=4pt, right=4pt, top=3pt, bottom=3pt]
\small
\textbf{Task ID:} \texttt{86350ffd-56e5-4fd9-90da-416688b5b182} \\
\textbf{Applications:} LibreOffice Calc 7.3.7.2; VSCode 1.85.1 \\
\textbf{Asset source:} Multilingual-Multimodal-NLP/TableBench spreadsheet library \\
\textbf{Category:} \texttt{mixed}\quad
\textbf{Complexity:} complex\quad
\textbf{Estimated steps:} 30 \\
\textbf{Initialization:} Open \path{/home/user/Desktop/test_files/xlsx/tablebench_0a53ffc65168d29f6a0784ec6741ffb0.xlsx} in LibreOffice Calc. \\
\textbf{Instruction:} The spreadsheet contains car data with columns \texttt{Model}, \texttt{Fuel Type}, \texttt{US gallons (mpg)}, \texttt{L/100km}, and \texttt{MZ Rating}. Add a new column F with the header \texttt{Efficiency Category}. For each car, assign \texttt{High} when \texttt{L/100km} is below 5.0, \texttt{Medium} when it is at least 5.0 and below 6.0, and \texttt{Low} otherwise. Save the spreadsheet. Then create \path{~/Desktop/categorize_check.py}, which reads the saved spreadsheet and prints the category counts as \texttt{High: <N>}, \texttt{Medium: <N>}, and \texttt{Low: <N>}, one line per category. \\
\textbf{Rule inputs:} modified source spreadsheet; \texttt{categorize\_check.py}; stdout from executing the script.
\end{tcolorbox}

% 评估器对三个终态检查取平均。第一个检查表头、逐行标签正确性以及是否不存在非法标签，并给予加权分数。第二个检查脚本是否读取 XLSX 文件，并是否基于数据计算计数，而不是只硬编码答案。第三个执行脚本，并在检查严格输出顺序和格式之前，对每个类别计数独立打分。
The evaluator averages three terminal checks. The first gives weighted credit for the header, correct per-row labels, and the absence of invalid labels. The second checks that the script reads an XLSX file and implements data-derived counting rather than only hard-coding an answer. The third runs the script and scores each category count independently before checking the strict output order and format.

% 该案例展示了一个跨越办公应用和代码编辑器的工作流如何进行组件级终态评分。奖励只在 episode 结束后分配，但它区分了表格正确性、数据驱动程序的构建，以及程序执行输出，而不是将这些结果压缩成单一的通过/失败判断。
This case illustrates component-wise terminal scoring for a workflow that crosses an office application and a code editor. The reward is assigned only after the episode, but it distinguishes spreadsheet correctness, the construction of a data-derived program, and the program's executed output rather than collapsing those outcomes into a single pass--fail check.

\begin{tcblisting}{
  enhanced,
  breakable,
  title={Rule-Based Evaluator for Task 1},
  title after break={Rule-Based Evaluator for Task 1 (continued)},
  colback=black!2,
  colframe=black!55,
  boxrule=0.5pt,
  listing only,
  listing options={language=Python,basicstyle=\ttfamily\tiny,breaklines=true,columns=fullflexible,showstringspaces=false},
  left=3pt,right=3pt,top=2pt,bottom=2pt
}
def call_rule_judge_1(result, expected, **options) -> float:
    """Check the spreadsheet header and per-row efficiency labels."""
    try:
        import openpyxl
        wb = openpyxl.load_workbook(result, data_only=True)
        ws = wb['Sheet1'] if 'Sheet1' in wb.sheetnames else wb.worksheets[0]

        f1_val = ws.cell(row=1, column=6).value
        header_score = 1.0 if f1_val is not None and str(f1_val).strip() == 'Efficiency Category' else 0.0

        data_rows = []
        for row in range(2, ws.max_row + 1):
            d_val = ws.cell(row=row, column=4).value
            f_val = ws.cell(row=row, column=6).value
            if d_val is not None or f_val is not None:
                data_rows.append((d_val, f_val))
        if not data_rows:
            return 0.0

        valid_values = {'High', 'Medium', 'Low'}
        correct = 0
        invalid = 0
        for d_val, f_val in data_rows:
            try:
                l100 = float(d_val)
                value = str(f_val).strip()
            except Exception:
                invalid += 1
                continue
            if value not in valid_values:
                invalid += 1
                continue
            expected_value = 'High' if l100 < 5.0 else ('Medium' if l100 < 6.0 else 'Low')
            if value == expected_value:
                correct += 1
            else:
                invalid += 1

        total = len(data_rows)
        correctness_score = correct / total
        no_invalid_score = max(0.0, 1.0 - invalid / total)
        return round(0.2 * header_score + 0.5 * correctness_score + 0.3 * no_invalid_score, 4)
    except Exception:
        return 0.0


def call_rule_judge_2(result, expected, **options) -> float:
    """Check that the audit script derives category counts from the spreadsheet."""
    try:
        import os
        import re
        if not os.path.isfile(result):
            return 0.0
        with open(result, 'r', encoding='utf-8', errors='replace') as stream:
            content = stream.read()
        if not content.strip():
            return 0.0

        lower = content.lower()
        score = 0.0
        reads_xlsx = any(package in lower for package in ('openpyxl', 'pandas', 'xlrd'))
        references_file = ('tablebench_0a53ffc65168d29f6a0784ec6741ffb0' in content or
                           ('.xlsx' in content and any(token in lower for token in ('read_excel', 'load_workbook', 'open'))))
        has_conditionals = bool(re.search(r'\bif\b.*\belse\b|\belif\b', content))
        has_threshold = bool(re.search(r'5\.0|5,0|[<>=]=?\s*5|[<>=]=?\s*6', content))
        has_labels = all(re.search(label, content, re.IGNORECASE) for label in ('high', 'medium', 'low'))
        has_loop = bool(re.search(r'\bfor\b|\bwhile\b|iterrows|itertuples', content))
        has_count_logic = bool(re.search(r'count|sum|len|value_counts|groupby', lower))

        if reads_xlsx:
            score += 0.25
        if references_file:
            score += 0.25
        if has_conditionals or has_threshold:
            score += 0.15
        if has_labels:
            score += 0.20
        if 'print' in content:
            score += 0.15
        if not reads_xlsx or (not has_loop and not has_count_logic):
            score *= 0.5
        return round(min(score, 1.0), 4)
    except Exception:
        return 0.0


def call_rule_judge_3(result, expected, **options) -> float:
    """Run the script output against counts derived from the saved spreadsheet."""
    try:
        import re
        import openpyxl
        lines = [line.strip() for line in (result or '').splitlines() if line.strip()]
        pattern = re.compile(r'^(High|Medium|Low):\s*(\d+)$')
        parsed = {}
        for line in lines:
            match = pattern.match(line)
            if match:
                parsed[match.group(1)] = int(match.group(2))
        if not all(label in parsed for label in ('High', 'Medium', 'Low')):
            return 0.0

        wb = openpyxl.load_workbook(expected, data_only=True)
        ws = wb['Sheet1'] if 'Sheet1' in wb.sheetnames else wb.worksheets[0]
        counts = {'High': 0, 'Medium': 0, 'Low': 0}
        for row in range(2, ws.max_row + 1):
            value = ws.cell(row=row, column=4).value
            if value is None:
                continue
            l100 = float(value)
            counts['High' if l100 < 5.0 else ('Medium' if l100 < 6.0 else 'Low')] += 1

        count_score = sum(parsed[label] == counts[label] for label in counts) / 3.0
        order_ok = len(lines) == 3 and [pattern.match(line).group(1) for line in lines] == ['High', 'Medium', 'Low']
        return round(0.7 * count_score + 0.3 * float(order_ok), 4)
    except Exception:
        return 0.0
\end{tcblisting}

\subsection{Synthesis Task Case Study 2: Media Workflow}
\label{app:case_media_workflow}

% 第二个任务连接 OBS Studio 和 Shotcut。它先在 OBS 中配置录制流水线，然后将分辨率和帧率要求迁移到兼容的 Shotcut 编辑项目中。该任务标记为中等难度，预计需要 22 个交互步骤。
The second task connects OBS Studio and Shotcut. It configures a recording pipeline in OBS and then transfers the resulting resolution and frame-rate requirements to a compatible editing project in Shotcut. The task is labeled medium and estimated to require 22 interaction steps.

\begin{tcolorbox}[title={Reviewed Synthesized Task 2}, colback=black!2, colframe=black!55, boxrule=0.5pt, left=4pt, right=4pt, top=3pt, bottom=3pt]
\small
\textbf{Task ID:} \texttt{df728e7b-240a-49f8-972e-27171051f3ab} \\
\textbf{Applications:} OBS Studio 27.2.3+dfsg1-1; Shotcut 22.01.30 \\
\textbf{Category:} \texttt{app\_only}\quad
\textbf{Complexity:} medium\quad
\textbf{Estimated steps:} 22 \\
\textbf{Initialization:} Launch OBS Studio. \\
\textbf{Instruction:} Configure OBS Studio with recording path \path{/home/user/Videos}, recording format \texttt{mkv}, output resolution \texttt{1280x720}, and common frame rate \texttt{30}. Then create a Shotcut project whose video mode is \texttt{HD 720p 30 fps} so that it matches the OBS output configuration. Save the project as \path{~/Videos/obs_match_project.mlt}. Finally, create \path{~/Desktop/video_config.txt} with exactly two lines: \texttt{Resolution: 1280x720} and \texttt{FPS: 30}. \\
\textbf{Rule inputs:} persisted OBS profile configuration; saved Shotcut \texttt{.mlt} project; final text record.
\end{tcolorbox}

% 评估器对两个应用的配置检查和最终文本记录取平均。OBS 根据录制路径、格式、分辨率和帧率字段的正确数量给予部分分数。Shotcut 在项目配置匹配 720p/30 时获得满分，并在三个核心字段中有两个匹配时获得部分分数。文本记录则按行独立评分。
The evaluator averages configuration checks for the two applications and the final text record. OBS receives partial credit according to the number of correct recording-path, format, resolution, and frame-rate fields. Shotcut receives full credit for a matching 720p/30 project profile and partial credit when two of its three core fields match. The text record is scored independently by line.

% 该任务具有明确的跨应用交接过程：OBS 建立录制规格，Shotcut 必须实例化一个与之兼容的项目。评估器检查持久化应用状态和项目元数据，而不是回放预设动作序列；部分终态分数则区分了不完整配置和完全兼容的生产设置。
This task has an explicit cross-application handoff: OBS establishes the recording specification, and Shotcut must instantiate a project that is compatible with it. The evaluator checks persistent application state and project metadata rather than replaying a prescribed action sequence, while the partial terminal score distinguishes incomplete configuration from a fully compatible production setup.

\begin{tcblisting}{
  enhanced,
  breakable,
  title={Rule-Based Evaluator for Task 2},
  title after break={Rule-Based Evaluator for Task 2 (continued)},
  colback=black!2,
  colframe=black!55,
  boxrule=0.5pt,
  listing only,
  listing options={language=Python,basicstyle=\ttfamily\tiny,breaklines=true,columns=fullflexible,showstringspaces=false},
  left=3pt,right=3pt,top=2pt,bottom=2pt
}
def call_rule_judge_1(result, expected, **options) -> float:
    """Check persistent OBS recording and video settings."""
    try:
        import configparser
        parser = configparser.RawConfigParser()
        parser.read_string(result or '')

        def get_key(key):
            for section in parser.sections():
                for name, value in parser.items(section):
                    if name.lower() == key:
                        return value.strip()
            return None

        recording_path = get_key('recfilepath') or get_key('filepath')
        recording_format = get_key('recformat2') or get_key('recformat')
        output_x = get_key('outputcx') or get_key('basecx')
        output_y = get_key('outputcy') or get_key('basecy')
        fps = get_key('fpscommon')
        checks = [
            recording_path == '/home/user/Videos',
            recording_format is not None and recording_format.lower() == 'mkv',
            output_x == '1280',
            output_y == '720',
            fps == '30',
        ]
        passed = sum(checks)
        return 1.0 if passed == 5 else (round(passed / 5.0, 4) if passed >= 3 else 0.0)
    except Exception:
        return 0.0


def call_rule_judge_2(result, expected, **options) -> float:
    """Check that the Shotcut project matches the OBS 720p/30 configuration."""
    try:
        import xml.etree.ElementTree as ET
        root = ET.fromstring(result)
        profile = root if root.tag == 'profile' else next((node for node in root.iter('profile')), None)
        if profile is None:
            return 0.0

        width_ok = profile.get('width', '').strip() == '1280'
        height_ok = profile.get('height', '').strip() == '720'
        try:
            fps = float(profile.get('frame_rate_num', '')) / float(profile.get('frame_rate_den', '1'))
            fps_ok = abs(fps - 30.0) < 0.05
        except Exception:
            fps_ok = False
        passed = sum((width_ok, height_ok, fps_ok))
        return 1.0 if passed == 3 else (0.5 if passed == 2 else 0.0)
    except Exception:
        return 0.0


def call_rule_judge_3(result, expected, **options) -> float:
    """Check the two-line handoff record."""
    try:
        lines = (result or '').splitlines()
        while lines and not lines[-1].strip():
            lines.pop()
        if len(lines) != 2:
            return 0.0
        resolution_ok = lines[0] == 'Resolution: 1280x720'
        fps_ok = lines[1] == 'FPS: 30'
        return 1.0 if resolution_ok and fps_ok else (0.5 if resolution_ok or fps_ok else 0.0)
    except Exception:
        return 0.0
\end{tcblisting}

\subsection{Agent Behavior Case Study: GUI--Code Coordination}
\label{app:case_behavior_coordination}

% 图~\ref{fig:case_study_coordination} 展示了两条成功轨迹，其中 \ours{} 将 GUI 感知与代码执行进行协调。这些案例旨在使学习到的混合行为更加具体：当任务依赖当前应用状态或用户可见界面时使用 GUI，而当同一目标包含结构化抽取、精确检查或重复操作时使用代码。
Figure~\ref{fig:case_study_coordination} shows two successful trajectories in which \ours{} coordinates GUI perception with code execution. These cases are intended to make the learned hybrid behavior concrete: the GUI is used where the task depends on the current application state or a user-facing interface, while code is used where the same goal contains structured extraction, precise inspection, or repetitive operations.

% 在第一个案例中，智能体从浏览器/PDF 界面出发，利用已打开的课程材料推断讲义文件的 URL 模式。当这一模式通过 GUI 观察得到 grounding 后，智能体切换到代码，并通过循环下载其余周次的 PDF。关键行为并不只是代码更适合批量下载文件，而是代码动作依赖于先前从实时 GUI 状态中获得的信息。这展示了一种互补路径：GUI 交互解析任务特定上下文，代码随后执行工作流中可扩展的部分。
In the first case, the agent starts from the browser/PDF interface and uses the opened course material to infer the lecture-file URL pattern. Once this pattern is grounded by GUI observation, the agent switches to code and downloads the remaining weekly PDFs in a loop. The important behavior is not simply that code is faster for downloading files, but that the code action is conditioned on information first obtained from the live GUI state. This illustrates a complementary route in which GUI interaction resolves task-specific context and code then executes the scalable part of the workflow.

% 在第二个案例中，智能体用代码检查缴费表格，以识别尚未缴纳学费的学生，然后回到 GUI，在邮件客户端中填写收件人。这里，代码负责对表格记录进行可靠的结构化筛选，而 GUI 交互完成面向用户的通信步骤。这两个例子共同表明，统一动作空间允许策略跨模态分解任务，而不是在整条轨迹中固定使用单一界面。
In the second case, the agent inspects a payment spreadsheet with code to identify students whose tuition has not been paid, then returns to the GUI to fill the e-mail recipients in the mail client. Here, code provides reliable structured filtering over tabular records, while GUI interaction completes the user-facing communication step. Together, the two examples show that the unified action space allows the policy to decompose a task across modalities instead of committing to a single interface throughout the trajectory.

\ifoshelixopenversion
  \begin{figure}[!t]
\else
  \begin{figure}[H]
\fi
\centering
\includegraphics[width=\linewidth]{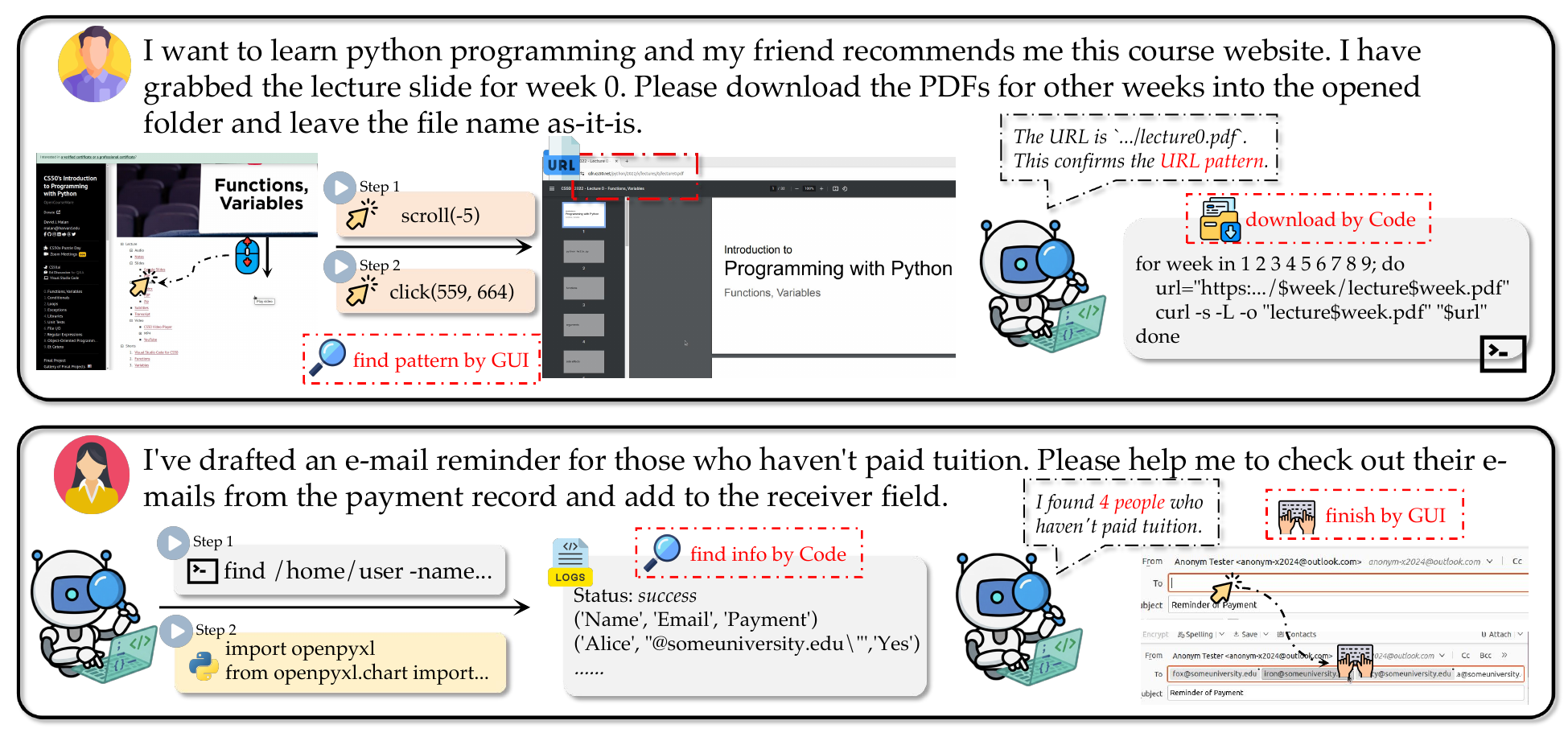}
\captionsetup{skip=3pt,position=bottom}
\caption{Successful GUI--code coordination cases. The top case first uses GUI observation to infer the URL pattern from an opened lecture slide, then uses code to download the remaining weekly PDFs. The bottom case uses code to extract payment records from a spreadsheet and then returns to the GUI to fill the e-mail recipients.}
\label{fig:case_study_coordination}
\end{figure}

\subsection{Agent Behavior Case Study: Cross-Modality Verification}
\label{app:case_behavior_cross_modality}

% 图~\ref{fig:case_study_cross_modality} 展示了两条成功轨迹，其中一种模态完成任务，另一种模态验证或修复结果。这种行为不同于在任务开始时简单选择 GUI 或代码：智能体可以将第二种模态作为中间反馈通道，发现预期产物与实际产物之间的不一致，并继续通过纠正动作推进轨迹。
Figure~\ref{fig:case_study_cross_modality} presents two successful trajectories where one modality solves the task and the other verifies or repairs the result. This behavior is qualitatively different from merely choosing between GUI and code at the beginning of a task: the agent can use the second modality as an intermediate feedback channel, detect a mismatch between the intended and realized artifact, and then continue the trajectory with a corrective action.

% 在电子表格图表案例中，智能体首先用代码创建请求的簇状柱形图。随后 GUI 观察发现生成的图表只包含 Sales 系列而缺少 COGS；如果仅沿代码路径执行且不显式检查渲染产物，这一问题很难被发现。智能体随后通过代码修复图表，生成所需的双系列可视化。该轨迹表明，GUI 反馈可以暴露代码生成产物中的视觉或应用层错误，并将其转化为可执行的修正。
In the spreadsheet-chart case, the agent first creates the requested clustered column chart with code. The subsequent GUI observation reveals that the produced chart only contains the Sales series and misses COGS, which would be difficult to notice from the code path alone without explicitly checking the rendered artifact. The agent then repairs the chart through code, producing the required two-series visualization. This trace shows how GUI feedback can expose visual or application-level errors in code-generated artifacts and turn them into actionable corrections.

% 在幻灯片编辑案例中，智能体通过 GUI 执行可见编辑：在第二张幻灯片上添加备注，并将标题加粗。随后它切换到基于代码的已保存演示文件检查，以验证持久化产物是否符合指令，包括备注文本和标题格式。这一反向路径，即 GUI 用于编辑、代码用于验证，说明混合策略的收益是双向的。附录~\ref{app:case_behavior_coordination} 和附录~\ref{app:case_behavior_cross_modality} 中的四条成功轨迹共同表明，学习到的策略将 GUI 和代码作为相互检查、相互支撑的工具，支持统一 GUI/代码动作空间带来互补求解能力，而不是简单地用一种模态替代另一种模态。
In the slide-editing case, the agent performs the visible edits through GUI actions: it adds the note on the second slide and bolds the title. It then switches to code-based inspection of the saved presentation file to verify that the persistent artifact matches the instruction, including both the note text and the title formatting. This reverse direction, GUI for editing and code for validation, highlights that the benefit of the hybrid policy is bidirectional. Across the four successful traces in Appendix~\ref{app:case_behavior_coordination} and Appendix~\ref{app:case_behavior_cross_modality}, the learned policy uses GUI and code as mutually checking and mutually enabling tools, supporting the claim that a unified GUI/code action space yields complementary problem solving rather than replacing one modality with the other.

\ifoshelixopenversion\else
  \ifoshelixopenversion
  \begin{figure}[!t]
\else
  \begin{figure}[H]
\fi
\centering
\includegraphics[width=\linewidth]{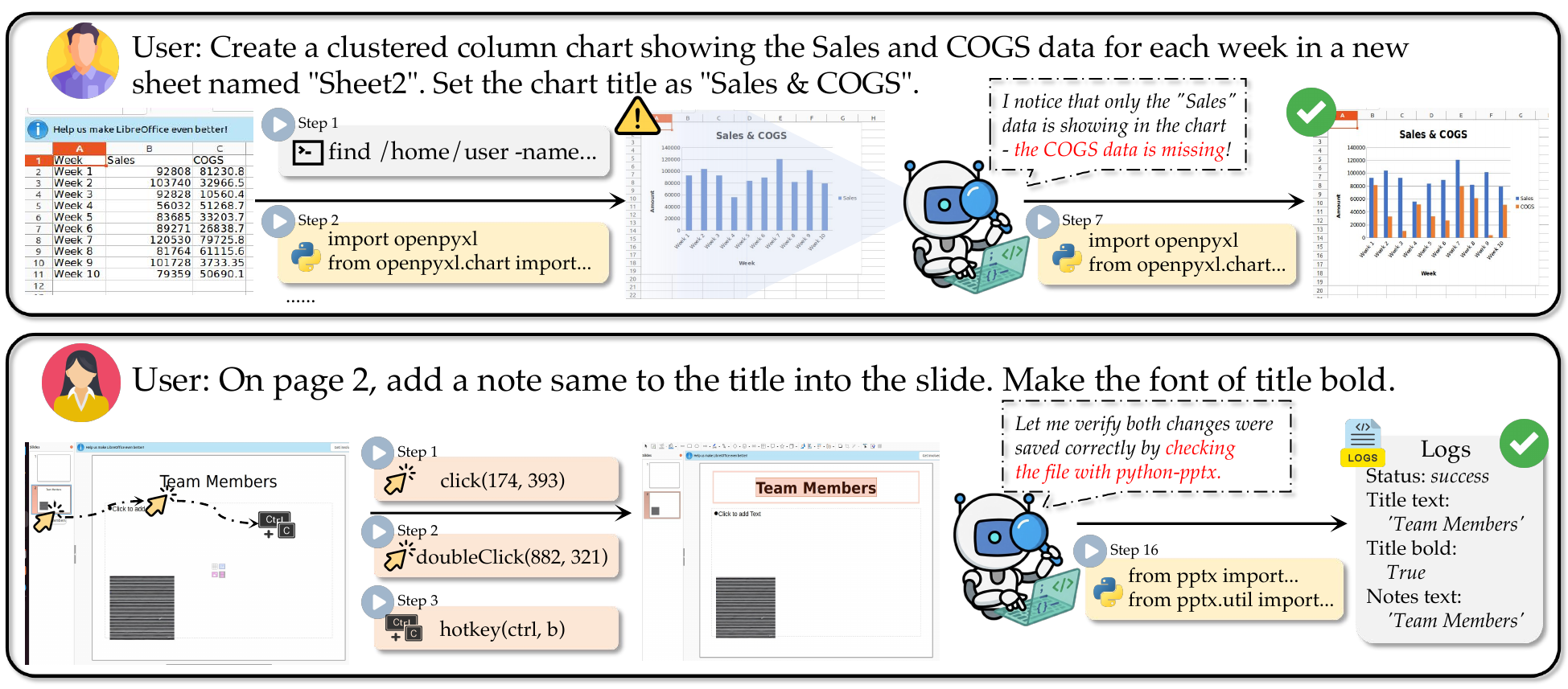}
\captionsetup{skip=3pt,position=bottom}
\caption{Successful cross-modality solve-and-verify cases. The top case creates a spreadsheet chart with code and then detects from the GUI that one data series is missing before repairing it. The bottom case completes slide editing through GUI actions and then verifies the saved file with code.}
\label{fig:case_study_cross_modality}
\end{figure}

\fi

\subsection{Failure Analysis: Evaluation Coverage and Cross-Modality Handoffs}
\label{app:case_behavior_bad_case}

% 图~\ref{fig:case_study_bad_case} 展示两类具有代表性的失败，分别揭示了任务规范与评估覆盖之间的偏差，以及 GUI 与代码切换时的状态交接问题。它们并非单一模态能力不足，而是混合策略在解释视觉意图和维护跨模态持久化状态时仍面临的系统性困难。
Figure~\ref{fig:case_study_bad_case} presents two representative failures that expose a gap between task specification and evaluation coverage, as well as a state-handoff problem at the GUI--code boundary. Rather than indicating a deficiency of either modality in isolation, these traces identify systematic challenges for a hybrid policy in interpreting visual intent and maintaining persistent state across modalities.

% 上例反映了规范--评估覆盖缺口。``保持原始格式'' 是一个由行间底纹、数字显示、边框和其他渲染属性共同构成的视觉等价条件；若指令与验证信号未完整刻画这些属性，或未对不同求解路径提供等价覆盖，就会形成偏向某一模态的欠规范代理目标。智能体在代码转换后通过 GUI 发现不匹配，却将其错误归因为缺少货币符号，从而添加了美元符号而未恢复真正缺失的交替行底纹。此任务的自然 GUI 解法是选中 Calc 表格并复制粘贴到 Writer，以保留其渲染外观；经由代码序列化则需要显式重建该视觉格式。因此，评测应使用户意图、渲染语义与验证覆盖保持一致，并覆盖不预设操作模态的完整有效解集。
The top trace illustrates a specification--evaluation coverage gap. ``Preserve the original formatting'' defines a visual equivalence condition involving row banding, numeric display, borders, and other rendered attributes. When the instruction and validation signal do not fully characterize these attributes, or do not cover admissible solution paths equivalently, they create an underspecified proxy objective that can favor one modality. After code-based conversion, the agent detects a mismatch through the GUI but attributes it to missing currency symbols, adding dollar signs rather than restoring the missing alternating row shading. A natural GUI solution would select the Calc table and paste it into Writer, preserving its rendered appearance; serialization through code instead requires the visual formatting to be reconstructed explicitly. This case motivates aligning user intent, rendering semantics, and verification coverage so that the benchmark recognizes the full set of valid solutions without presupposing an action modality.

% 下例体现了跨模态状态交接失配。智能体首先通过 GUI 正确完成第一个子目标；随后代码侧修改并保存第二个子目标时，仅将其局部磁盘状态写入文件，未合并仍驻留于 GUI 内存态的前序修改。之后 GUI 将代码保存误解为全局提交并选择 ``Don't Save''，从而丢弃了第一个子目标的正确更新。该错误并非两个子目标各自的规划失败，而是持久化状态同步与提交语义在模态切换时丢失：可靠的混合执行需要在切换前协调 GUI 内存态与代码侧工件状态，并维护对保存范围和提交状态的一致表征。
The bottom trace exposes a cross-modality state-handoff failure. The agent first completes the first subgoal correctly through the GUI. When code subsequently edits and saves the second subgoal, it writes only a local disk-side state and does not incorporate the preceding modification that remains in the GUI-resident state. The GUI then mistakes the code-side save for a global commit and selects ``Don't Save'', discarding the correct update for the first subgoal. This is not a planning failure on either subgoal in isolation; it is a loss of persistence and commit semantics at the modality transition. Reliable hybrid execution requires reconciling GUI-resident and code-side artifact states before handoff, while maintaining a consistent representation of save scope and commit status.

\ifoshelixopenversion
  
\fi
\begin{figure}[H]
    \centering
    \includegraphics[width=\linewidth]{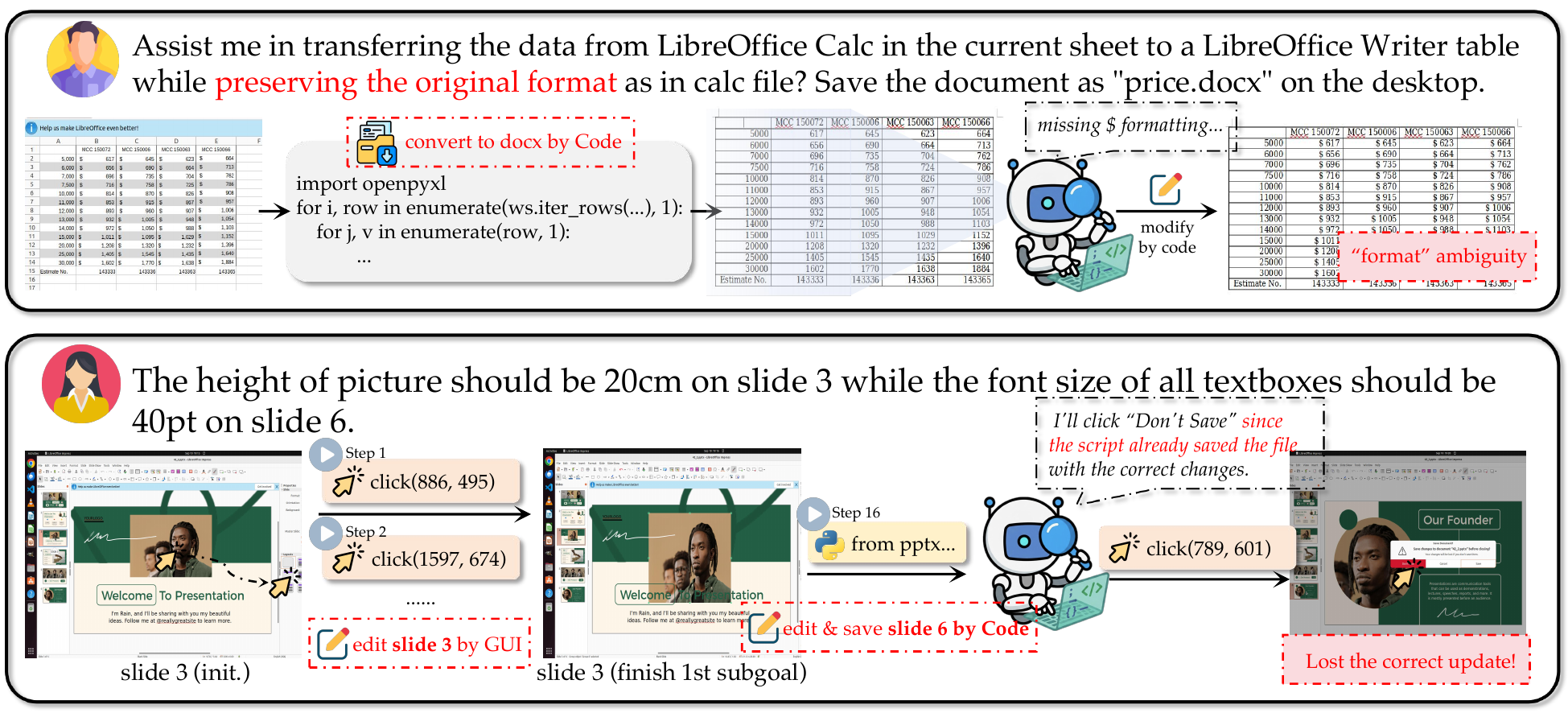}
    \captionsetup{skip=3pt,position=bottom}
    \caption{Representative failure cases. \textbf{Top:} In a Calc-to-Writer transfer task requiring preservation of the original formatting, code-based conversion fails to retain the alternating row shading. The subsequent GUI observation identifies a mismatch but misattributes it to missing currency symbols, so the agent adds dollar signs instead of restoring the table style. \textbf{Bottom:} After correctly completing the first PowerPoint subgoal through GUI actions, the agent edits and saves the second with code. The code-side save does not include the unsaved GUI state, and selecting ``Don't Save'' subsequently discards the earlier update.}
    \label{fig:case_study_bad_case}
\end{figure}

\subsection{Trajectory Audit}
\label{app:multiapp_audit}

% 我们从多步骤、跨应用的 Workflow 子集中人工标注了 30 条具有代表性的轨迹，以考察模态切换是否反映了以任务为导向的协同机制。对于每条轨迹，我们记录了最终结果、参与协同的关键 GUI 与代码阶段的压缩流程，以及任何状态或信息交接的目的；仅显示应用程序界面的截图不计入 GUI 交互。如表~\ref{tab:multiapp_audit} 所示，30 条轨迹中有 27 条包含有根基的 GUI/代码交接，其中 22 条轨迹中一种模态对另一种模态产生的状态进行了验证或修复。被审计集合包含 21 条成功、4 条部分成功和 5 条失败的轨迹。在 5 条失败轨迹中的 4 条里，跨模态状态传递本身是正确的，失败发生在更晚的阶段，源于任务未完成或应用程序/工具错误。剩余 1 条轨迹暴露了一次交接失败：代码写入电子表格时，将 GUI 中观察到的 PDF 来源替换为了另一个未经验证的 URL。
We manually annotated 30 representative trajectories from the multi-step Work. category to examine whether modality transitions reflect task-grounded coordination. 
For each trajectory, we recorded the final outcome, a compressed flow of the key GUI and code stages involved in coordination, and the purpose of any state or information handoff. 
The reported flow retains the principal stages that produce, consume, verify, or repair the reported handoff, while omitting repeated within-modality actions and transitions unrelated to that handoff. 
As summarized in Table~\ref{tab:multiapp_audit}, 27/30 trajectories contain a grounded GUI/code handoff, including 22/30 in which one modality verifies or repairs state produced by the other. 
The audited set contains 21 successful, 4 partially successful, and 5 failed trajectories. In four of the five failed trajectories, cross-modal state is transferred correctly and failure arises later from task incompletion or application/tool errors. The remaining trajectory exposes a failed handoff: the PDF source observed in the GUI is replaced by a different, unverified URL when code writes the spreadsheet. 
% 由于审计仅覆盖一个具有代表性的任务子集，其中观察到的行为比例可能无法反映整个基准上的总体趋势。此外，这些轨迹本身也不能证明所采用的路由选择是最优的。即便存在这些限制，这些案例仍直接展示了 \ours{} 能够执行以任务为导向的信息交接、跨模态验证与修复，即将 GUI 与代码作为相互赋能和相互校验的机制加以使用。这种过程层面的证据补充了仅反映最终结果的聚合任务得分，并进一步支持了本工作关于模型能够学习实质性跨模态协同的核心主张。
As the audit covers only a representative task subset, the observed behavioral frequencies may not reflect the overall trends across the full benchmark. Moreover, the audited trajectories do not establish that the observed routing decisions are optimal. 
Even with these limitations, the examples directly show that \ours{} performs task-grounded information handoffs and cross-modal verification and repair, using GUI and code as mutually enabling and checking mechanisms. This process-level evidence complements aggregate task scores, which capture outcomes rather than coordination behavior, and further supports our central claim that the model learns substantive cross-modal coordination.

\begin{table}[H]
\centering
\scriptsize
\setlength{\tabcolsep}{2.6pt}
\renewcommand{\arraystretch}{0.84}
\begin{tabularx}{\textwidth}{@{}l l l X c@{}}
\toprule
\textbf{Task ID} & \textbf{Outcome} & \textbf{Key Modality Flow} & \textbf{Handoff Purpose} & \textbf{Label} \\
\midrule
\texttt{2c9fc0de} & Success & Code only & Command-line completion; no handoff & A \\
\texttt{e135df7c} & Success & Code only & Programmatic conversion; no handoff & A \\
\texttt{a503b07f} & Success & Code only & Programmatic artifact creation; no handoff & A \\
\texttt{26660ad1} & Success & GUI $\rightarrow$ Code & Transfer measured network results to a text record & B \\
\texttt{4c26e3f3} & Success & GUI $\rightarrow$ Code & Identify the visual target, then extract and edit its asset & B \\
\texttt{f5c13cdd} & Success & Code $\rightarrow$ GUI & Transfer filtered recipients into an e-mail draft & B \\
\texttt{415ef462} & Success & GUI $\rightarrow$ Code & Transfer e-mail and attachment context for file extraction & B \\
\texttt{c2751594} & Success & GUI $\rightarrow$ Code & Transfer a saved attachment for programmatic extraction and use & B \\
\texttt{c867c42d} & Success & GUI $\rightarrow$ Code & Verify and convert a GUI-exported address book & C \\
\texttt{48d05431} & Success & Code $\rightarrow$ GUI $\rightarrow$ Code & Validate and repair an environment configuration & C \\
\texttt{8e116af7} & Success & Code $\rightarrow$ GUI & Inspect a spreadsheet edit and reload stale application state & C \\
\texttt{3a93cae4} & Success & Code $\rightarrow$ GUI & Visually confirm a programmatic spreadsheet edit & C \\
\texttt{26150609} & Success & Code $\rightarrow$ GUI & Inspect the repaired program state before termination & C \\
\texttt{9219480b} & Success & Code $\rightarrow$ GUI & Test repaired game logic in the running application & C \\
\texttt{da922383} & Success & GUI $\rightarrow$ Code & Validate and repair GUI-saved web articles & C \\
\texttt{81c425f5} & Success & Code $\rightarrow$ GUI $\rightarrow$ Code & Check GUI edits against the persisted file state & C \\
\texttt{02ce9a50} & Success & GUI $\rightarrow$ Code & Validate and capture a GUI terminal state & C \\
\texttt{7ff48d5b} & Success & GUI $\rightarrow$ Code $\rightarrow$ GUI & Generate a document from web facts and visually repair it & C \\
\texttt{ce2b64a2} & Success & Code $\rightarrow$ GUI $\rightarrow$ Code & Identify images visually, rename them, and check the result & C \\
\texttt{0e5303d4} & Success & GUI $\rightarrow$ Code $\rightarrow$ GUI & Infer a URL pattern, batch-download files, and inspect them & C \\
\texttt{aceb0368} & Success & Code $\rightarrow$ GUI & Visually verify an automatically graded spreadsheet & C \\
\texttt{2c1ebcd7} & Partial & Code $\rightarrow$ GUI $\rightarrow$ Code & Cross-check programmatic edits and rendered formatting & C \\
\texttt{aad10cd7} & Partial & GUI $\rightarrow$ Code $\rightarrow$ GUI & Clean and repair content extracted through the browser & C \\
\texttt{778efd0a} & Partial & Code $\rightarrow$ GUI $\rightarrow$ Code & Verify and repair audio and presentation state & C \\
\texttt{47f7c0ce} & Partial & Code $\rightarrow$ GUI & Inspect a generated slide asset and repair its placement & C \\
\texttt{69acbb55} & Failure & Code $\rightarrow$ GUI $\rightarrow$ Code & Fall back to browser download and monitor file state & C \\
\texttt{dd60633f} & Failure & Code $\rightarrow$ GUI $\rightarrow$ Code & Replace a failed fetch with GUI download and parse its output & C \\
\texttt{3c8f201a} & Failure & Code $\rightarrow$ GUI & Transfer a prepared asset into GIMP and inspect execution & C \\
\texttt{e1fc0df3} & Failure & GUI $\rightarrow$ Code $\rightarrow$ GUI & Transfer an asset URL and verify alternative download paths & C \\
\texttt{5990457f} & Failure & GUI $\rightarrow$ Code $\rightarrow$ GUI & Transfer Scholar evidence to a spreadsheet; PDF link becomes unverified & D \\
\bottomrule
\end{tabularx}
\captionsetup{skip=2pt,position=bottom,font=footnotesize}
\caption{Manual audit of 30 representative Work. trajectories. \textbf{Task IDs} are shown by unique eight-character prefixes; \textbf{Key Modality Flow} is a compressed summary of the principal stages relevant to the reported handoff; \textbf{Handoff Purpose} states what information, artifact, or application state is transferred or checked; \textbf{Labels} follow the definitions: \textit{A} denotes single-modality execution without a substantive handoff, \textit{B} denotes a task-grounded cross-modal handoff, \textit{C} denotes cross-modal verification or repair, and \textit{D} denotes a failed handoff.}
\label{tab:multiapp_audit}
\vspace{-0.6\baselineskip}
\end{table}

\section{\ours{} Tool Schema and Prompts}
\label{app:prompts}

\tcbset{
  osPromptBase/.style={
    enhanced,
    breakable,
    listing only,
    boxrule=0.45pt,
    arc=2pt,
    outer arc=2pt,
    boxsep=2pt,
    left=4pt,
    right=4pt,
    top=4pt,
    bottom=4pt,
    before skip=0.65em,
    after skip=0.9em,
    fonttitle=\bfseries\small,
    coltitle=white,
    listing options={
      basicstyle=\ttfamily\scriptsize,
      breaklines=true,
      breakatwhitespace=false,
      columns=fullflexible,
      keepspaces=true,
      showstringspaces=false
    }
  },
  osPromptSchema/.style={osPromptBase, colback=black!2, colframe=black!55, colbacktitle=black!55},
  osPromptPolicy/.style={osPromptBase, colback=blue!2, colframe=blue!50!black, colbacktitle=blue!55!black},
  osPromptData/.style={osPromptBase, colback=green!2, colframe=green!45!black, colbacktitle=green!45!black}
}

% 本附录报告 \ours{} 在 SFT、强化学习和测试时推理中使用的提示词与工具模式，并额外记录用于数据基础设施的任务生成和评估器生成提示词。训练提示词表示 SFT 阶段使用的系统提示词；强化学习与测试时推理均使用推理提示词。带代码动作的工具模式在 SFT、强化学习和测试之间共享，动作空间遵循同一模式。用于从不同教师模型蒸馏轨迹的提示词随教师而变化，因此不在此逐一列出。
We report the prompts and tool schema used by \ours{} for SFT, RL, and test-time inference, together with shortened generation and quality-review prompts used by the data infrastructure. The train prompt denotes the system prompt used during SFT, while RL and test-time inference use the inference prompt. The tool schema with code is shared across SFT, RL, and testing, and the action space follows the same schema. Prompts used to distill trajectories from different teacher models vary by teacher and are not enumerated here. 
% The task-generation subsection explicitly distinguishes the primary application-centered process from the later multi-application prototype.

\clearpage
\subsection{\ours{} Tool Schema}
\label{app:tool_schema}

\begin{tcblisting}{osPromptSchema, title={OS-Helix Tool Schema}}
[
  {
    "type": "function",
    "function": {
      "name": "custom_computer_use",
      "description": "Control a desktop GUI and execute system-level code. Use it to move the mouse, click, type, scroll, wait, terminate tasks, and run raw Python or Bash code on the operating system.",
      "parameters": {
        "type": "object",
        "required": ["action"],
        "properties": {
          "action": {
            "type": "string",
            "enum": [
              "key", "type", "mouse_move", "left_click", "left_click_drag",
              "right_click", "middle_click", "double_click", "triple_click",
              "scroll", "hscroll", "wait", "terminate", "code"
            ],
            "description": "The operation to perform."
          },
          "keys": {"type": "array", "description": "Required only by action=key."},
          "text": {"type": "string", "description": "Required only by action=type."},
          "coordinate": {"type": "array", "description": "The x,y coordinates for mouse actions."},
          "pixels": {"type": "number", "description": "The amount of scrolling."},
          "time": {"type": "number", "description": "The seconds to wait."},
          "status": {"type": "string", "enum": ["success", "failure"]},
          "execute_code": {"type": "string", "description": "Required only when action=code."},
          "language": {"type": "string", "enum": ["python", "bash"]}
        }
      }
    }
  }
]
\end{tcblisting}

\clearpage
\subsection{\ours{} System Prompt on OSWorld}
\label{app:prompt_train_osworld}

\begin{tcblisting}{osPromptPolicy, title={\ours{} System Prompt on OSWorld}}
# Role & Goal
You are a powerful OS Agent capable of both GUI interaction and direct system-level programming and are using an Ubuntu virtual machine with x86_64 architecture with internet access.
Your goal is to complete tasks with MAXIMUM efficiency and MINIMUM steps.

# Environment & Screen
- The user's home directory is "/home/user".
- The user's sudo password is "password".
- The screen's resolution is represented on a 1000x1000 relative coordinate grid.
\end{tcblisting}

\subsection{\ours{} System Prompt on WindowsAgentArena}
\label{app:prompt_train_waa}

\begin{tcblisting}{osPromptPolicy, title={\ours{} System Prompt on WindowsAgentArena}}
# Role & Goal
You are a powerful OS Agent capable of both GUI interaction and direct system-level programming, and you are utilizing a Windows 11 virtual machine using x86_64 architecture with internet access.
Your goal is to complete tasks with MAXIMUM efficiency and MINIMUM steps.

# Environment & Screen
- **User**: Your username is "Docker".
- **Home**: Your home path is "C:\\Users\\Docker".
- The screen's resolution is represented on a 1000x1000 relative coordinate grid.

# Strict Code Action Requirement
- When using a `code` action, you MUST output and execute Python code only, the Windows environment does not support Bash. Treat Bash as unavailable.
- If command-line behavior is required, write Python code and explicitly call Windows `cmd.exe` or PowerShell from Python. For example, use `subprocess.run(["cmd", "/c", "..."])` for CMD commands, or `subprocess.run(["powershell", "-Command", "..."])` for PowerShell commands.
\end{tcblisting}

\clearpage
\subsection{Instruction Generation Prompts}
\label{app:prompt_instruction_generation}

\begin{tcblisting}{osPromptData, title={Instruction Generation Prompt (Simplified Version)}}
# Role and Objective
You are a safe multi-turn Exploration Agent that generates grounded Ubuntu GUI
tasks. Choose one non-destructive observation action, or finish with done.

# Environment
- The GUI user is "user"; the home directory is "/home/user".
- Sampled files normally live under "/home/user/Desktop". Prefer user-facing
  paths under /home/user; avoid sudo and package installation.

# Safe Exploration
- Context provides the sampled primary app, compatible artifacts, exact paths,
  launch configuration, tutorial, coverage memory, and initial screenshot.
- Prefer inspecting a sampled artifact with distinctive content when available.
- `open.app` must be the sampled app or an allowed companion; `open.path` must be a
  sampled path or omitted. Use `click` and `scroll` only for observation.
- Do not mutate files or settings, run scripts, submit forms, send messages, or
  access unstable network resources. Perform one atomic GUI action per turn.

# Proposal Grounding and Quality
- Generate exactly `requested_proposal_count` proposals.
- Center each proposal on the sampled app and its observed state or compatible
  artifacts, using only genuinely required resources.
- Add an allowed companion app or OS file operation only for a natural,
  verifiable dependency.
- `related_apps` and `used_files` list only required initialized resources; include
  exact user-visible paths in file-related instructions.
- Write concise, realistic goals, not tutorials. Do not prescribe GUI or code.
- Hide values the agent must derive behind instruction placeholders and record
  exact expected values only in the hidden evaluation fields.
- State all evaluator-critical non-answer constraints, including source, target,
  scope, order, quantity, format, filename, and final artifact.
- Avoid trivial, subjective, unstable, destructive, vague, or benchmark-like tasks.

# Coverage and Metadata
- Use `app_memory_summary` to prefer grounded under-covered features while avoiding
  recent task patterns and filename-only variants.
- Keep initialization context consistent across `config`, `related_apps`, and `used_files`.
- `category` is `app_only` for application/system functionality, `file_only` for
  processing input files into file artifacts, and `mixed` when both are involved.
- `complexity` estimates a pure-GUI solution: `easy` for 5--15 steps, `medium` for
  15--40 steps, and `complex` for more than 40 steps.
- `target_features` is keyed by entries in `related_apps` and names app-specific capabilities.

# Verifiability
- `success_criteria` states concrete terminal conditions; `evaluation_requirements_text`
  gives semantically equivalent deterministic checks. Neither may omit a condition.
- Use stable rule-based anchors whenever VM files or command output capture the
  condition; never rely on file existence alone. Specify exact VLM evidence only
  for indispensable visual or otherwise unparsable conditions.
- For a natural composition, `dependency_chain` records ordered source, operation,
  target, and deterministic verification anchor; otherwise use [].
- Do not write evaluator code.

# Proposal Output
Return valid JSON only. Before finishing:
{"actions": [{"tool": "open|click|scroll|done", "arguments": {}}]}

When using `done`, `arguments` contains `proposals` with:
- `instruction`, `config`, `related_apps`, `used_files`
- `category`: `file_only` | `app_only` | `mixed`
- `complexity`: `easy` | `medium` | `complex`
- `target_features`, `success_criteria`, `evaluation_requirements_text`, `dependency_chain`
\end{tcblisting}

\clearpage
\subsection{Evaluator Generation Prompt}
\label{app:prompt_evaluator_generation}

% 中文：评估器生成提示词为已经接受且内容固定的指令提议合成验证逻辑。评估器智能体在与指令生成相互独立的上下文中运行，不能改写指令字段，并优先生成确定性的规则验证；只有必要的成功条件无法通过文件或命令可靠观测时，才补充或改用 VLM 判断。这里将提示词缩减为核心验证契约、getter 约束、规则函数要求和视觉判断条件。
% The Evaluator Agent synthesizes verification logic for an accepted and frozen instruction proposal. It runs in a context independent of instruction generation, cannot rewrite instruction fields, and prefers deterministic rule-based verification; VLM judgment is added or substituted only when an essential success condition cannot be observed reliably through files or commands. We shorten the prompt to its core verification contract, getter constraints, rule-function requirements, and visual-judgment conditions.

\begin{tcblisting}{osPromptData, title={Evaluator Synthesis Prompt (Simplified Version)}}
# Role and Objective
You synthesize only the verification spec for one accepted GUI instruction proposal.
Do not repeat or rewrite proposal fields such as `instruction`, `config`,
`related_apps`, `used_files`, `category`, `complexity`, or `target_features`.

# Environment
- The GUI user is "user"; the home directory is "/home/user".
- User-facing "~" maps to "/home/user".
- Avoid sudo, installation, destructive commands, GUI launches, network access,
  and any evaluator behavior that modifies the VM.

# Verification Contract
- Treat `proposal.evaluation_requirements_text` as the deterministic contract for
  `proposal.success_criteria`.
- Use `proposal.dependency_chain` to check that final artifacts are grounded in
  required source information, not merely plausible final outputs.
- Prefer file or command checks over visual checks. Do not merely check that a
  file exists; check content, structure, formatting, metadata, or state.
- Use rule-based verification whenever files or commands can fully express the
  success conditions. Add VLM judgment only for indispensable visual or
  otherwise unparsable state; VLM-only verification is allowed only when no
  reliable programmatic check exists.

# Getter Rules
Allowed getter types are `vm_file`, `vm_command_line`, and `empty`.
- `vm_file.path` must be an absolute VM path.
- `vm_command_line.command` must be a list of arguments, not a shell string.
- Use `empty` when the expected answer can be hard-coded inside the rule function.

# Rule Function Rules
- Each rule item contains a complete Python function named call_rule_judge_N.
- Signature: def call_rule_judge_N(result, expected, **options) -> float:
- Return a float in [0.0, 1.0]; catch exceptions and return 0.0.
- Do not directly open VM paths in code. Use getter-provided local result or
  expected paths, or parse stdout from `vm_command_line`.
- Decompose scoring into meaningful subgoals when possible, so partial progress
  receives partial credit but only dependency-faithful completion receives 1.0.

# Output
Return JSON:
{
  "verification": {
    "need_rule_judge": true,
    "need_vlm_judge": false,
    "vlm_desc": "When enabled, describe the exact visual evidence to inspect.",
    "rule_items": [
      {
        "result_getter": {},
        "expected_getter": {"type": "empty"},
        "code": "def call_rule_judge_1(result, expected, **options):\n    return 0.0"
      }
    ]
  }
}
\end{tcblisting}

\clearpage
\subsection{Instruction Quality Review Prompt}
\label{app:prompt_instruction_critic}

% 中文：指令提议生成后，独立审查模型对任务的可达性、无歧义性、真实性、扎根性、答案泄漏和验证条件充分性进行评分，并在目标数量约束下筛选候选项。复杂度仅作为后续难度筛选信号，而不单独决定接受与否。下面给出该审查提示词的缩略版本。
% After proposal generation, an independent reviewer scores task reachability, unambiguity, realism, grounding, answer leakage, and verification sufficiency, and selects candidates subject to the requested count. Complexity is retained as a downstream difficulty signal rather than an acceptance criterion by itself. The shortened review prompt is given below.

\begin{tcblisting}{osPromptData, title={Instruction Quality Review Prompt (Simplified Version)}}
# Role and Objective
Critique every GUI task proposal and accept at most `target_count`. Use
`rationality_score` as the primary quality signal; require strong verification
and answer-isolation scores. Treat `complexity_score` as descriptive only.

# Review Dimensions
- Grounding and feasibility: the task is reachable from `config`, uses only
  initialized `related_apps` and `used_files`, and makes no hidden assumptions.
- Clarity and realism: `instruction` is a concise user goal rather than a tutorial;
  it identifies exact paths, objects, destinations, output shape, and constraints.
- Verification sufficiency: `success_criteria` and
  `evaluation_requirements_text` are complete, mutually consistent, and precise
  enough to implement without access to exploration screenshots.
- Answer isolation: `instruction` does not reveal values that must be obtained by
  reading, counting, comparing, transforming, or inspecting source material.
- Rule grounding: require stable deterministic checks whenever VM files or command
  output capture the condition; file existence alone is insufficient. Visual-only
  conditions require an explicit, task-specific VLM rubric.
- Safety and stability: reject destructive or irreversible state changes, package
  installation, external side effects such as messaging, subjective-only goals,
  and current web/date dependencies.
- Data and config fit: sampled artifacts are used when relevant, target paths and
  application state match `config`, and existing files are edited in place unless
  the goal explicitly requests a new output.
- Diversity: avoid recent task patterns and filename-only variants recorded in
  application memory.
- If multiple applications are genuinely required, `dependency_chain` must state
  a verifiable source-to-operation-to-target dependency; do not reward unrelated
  parallel actions merely for involving more applications.

# Scoring and Selection
Use integer scores from 1 (unusable) to 5 (excellent).
- `rationality_score`: realism, reachability, grounding, safety, and unambiguity.
- `complexity_score`: planning horizon, transformations, precision, and state.
- `success_criteria_score`: completeness and consistency of the verification contract.
- `instruction_leakage_score`: withholding derived answers while remaining clear.
Reject low-quality proposals directly. For each rejection, provide a short
machine-readable `reason` and actionable `suggested_repair`. Rank accepted items
by `rationality_score`, using verification and leakage scores as quality gates.

# Output
Return valid JSON only:
{
  "proposal_scores": [{
    "proposal_id": "p01",
    "critic_scores": {
      "rationality_score": 1, "complexity_score": 1,
      "success_criteria_score": 1, "instruction_leakage_score": 1
    },
    "rationale": "brief score rationale"
  }],
  "accepted": ["p01"],
  "rejected": [{
    "proposal_id": "p02", "reason": "short_reason",
    "suggested_repair": "concise actionable feedback"
  }],
  "coverage_summary": {}
}
\end{tcblisting}

\clearpage
\subsection{Evaluator Quality Review Prompt}
\label{app:prompt_evaluator_critic}

% 中文：指令通过审查后，评估器审查模型独立检查生成的验证规范是否完整覆盖冻结后的成功条件、是否正确使用 getter、是否检查任务相关的内部结构，并能否抵抗初态、错误路径和部分完成等假阳性。模式验证、静态编译与初态测试由自动流程执行，其结果作为审查时的硬信号。下面给出该审查提示词的缩略版本。
% After instruction acceptance, a separate evaluator reviewer checks whether the generated verification specification covers the frozen success conditions, uses getters correctly, validates task-relevant structure, and resists false positives from the initial state, wrong targets, or partial completion. Schema validation, static compilation, and initial-state preflight are performed automatically and supplied as hard review signals. The shortened prompt is given below.

\begin{tcblisting}{osPromptData, title={Evaluator Quality Review Prompt (Simplified Version)}}
# Role and Objective
Critique and score the existing `verification` for one accepted task. Do not
rewrite the instruction, proposal metadata, or evaluator code. Decide whether
repair is required and give concise feedback for evaluator regeneration.

# Review Dimensions
- Coverage: verify every critical condition in `instruction`, `success_criteria`,
  and `evaluation_requirements_text`, including target identity, values, counts,
  order, formatting, metadata, negative conditions, and any genuine dependency.
- Getter correctness: require concrete absolute VM paths consistent with the task.
  A `vm_file` result is a local cached file path that rule code must open or parse;
  a `vm_command_line` result is stdout text.
- Structural validity: inspect the internal content or application-specific
  structure that proves completion, rather than file existence or size alone.
- False-positive resistance: initial, empty, partial, stale, wrong-path, wrong-file,
  or superficially similar states should receive low reward.
- Evidence matching: use deterministic rule checks when files or commands capture
  the condition; use VLM evidence only for indispensable visual or unparsable state.
- Safety and determinism: avoid network access, package installation, destructive
  commands, current external facts, and hidden assumptions.

# Scoring and Repair
Use integer scores from 1 (unusable) to 5 (excellent). Treat
`evaluator_quality_score` as primary and use `coverage_score`,
`getter_correctness_score`, `structure_validation_score`, and
`false_positive_resistance_score` to explain it. Set `repair_required` to true
whenever schema/static validation or initial-state preflight failed, or whenever
the review finds a material quality defect. Name the missing check, incorrect
getter semantics, unsafe behavior, or false-positive path in `suggested_repair`.

# Output
Return valid JSON only:
{
  "evaluator_scores": {
    "evaluator_quality_score": 1,
    "coverage_score": 1,
    "getter_correctness_score": 1,
    "structure_validation_score": 1,
    "false_positive_resistance_score": 1
  },
  "rationale": "brief score rationale",
  "repair_required": true,
  "rejected": {
    "reason": "short_reason",
    "suggested_repair": "concise actionable feedback"
  }
}
\end{tcblisting}

\end{document}